\pdfoutput=1
\documentclass[11pt]{article}
\usepackage[dvipsnames]{xcolor}
\usepackage{tcolorbox}
\usepackage{pdfpages}
\usepackage{svg}
\usepackage{float}
\usepackage{tabularx}
\usepackage{makecell}
\usepackage{booktabs}
\usepackage{ragged2e}
\usepackage{array}
\usepackage{subcaption}
\usepackage{graphicx}
\usepackage{adjustbox}
\usepackage{tabularx}
\usepackage{comment}
\usepackage{arydshln}
\usepackage{mdframed,lipsum}
\usepackage{soul}
\usepackage{booktabs}

\newtcolorbox{promptbox}{
  colback=white!20,
  colframe=gray!50!black,
  boxsep=10pt,
  left=10pt,
  right=10pt,
  top=10pt,
  bottom=10pt
}
\newtcolorbox{mybox}[1]{colback=white!20,colframe=gray!50!black,fonttitle=\bfseries,title=#1}
\usepackage{soul}

\definecolor{delectricblue}{RGB}{93, 117, 131}
\colorlet{lightdelectricblue}{delectricblue!30}

\newcommand{\hldb}[1]{%
    {%
    \sethlcolor{lightdelectricblue}%
    \hl{#1}%
    }%
}
\definecolor{lightblue}{rgb}{.90,.95,1}
\sethlcolor{lightblue}

\usepackage[final]{acl}
\usepackage{multirow}

\usepackage{times}
\usepackage{latexsym}

\usepackage[T1]{fontenc}
\usepackage[utf8]{inputenc}

\usepackage{microtype}

\usepackage{inconsolata}
\usepackage{comment}
\title{\emph{Decision-Making with Deliberation:}\\ Meta-reviewing as a Document-grounded Dialogue}

 \author{Sukannya Purkayastha$^1$, Nils Dycke$^1$, Anne Lauscher$^2$, Iryna Gurevych$^1$ \\
         $^1$ Ubiquitous Knowledge Processing Lab (UKP Lab),\\ Department of Computer Science, Technical University of Darmstadt and \\ National Research Center for Applied Cybersecurity ATHENE, Germany \\
         \url{www.ukp.tu-darmstadt.de} \\
         $^2$ Data Science Group, University of Hamburg
         }% 
\begin{document}
\maketitle
\begin{abstract}

%\todo{TBA}
%\vspace{8cm}

Meta-reviewing is a pivotal stage in the peer-review process, serving as the final step in determining whether a paper is recommended for acceptance. Prior research on meta-reviewing has treated this as a summarization problem over review reports. However, complementary to this perspective, meta-reviewing is a decision-making process that requires weighing reviewer arguments and placing them within a broader context. Prior research has demonstrated that decision-makers can be effectively assisted in such scenarios via dialogue agents. In line with this framing, %introduced in \citet{balke5},
we explore the practical challenges for realizing dialog agents that can effectively assist meta-reviewers. Concretely, we first address the issue of data scarcity for training dialogue agents by generating synthetic data using Large Language Models (LLMs) based on a self-refinement strategy to improve the relevance of these dialogues to expert domains. Our experiments demonstrate that this method produces higher-quality synthetic data and can serve as a valuable resource towards training meta-reviewing assistants. Subsequently, we utilize this data to train dialogue agents tailored for meta-reviewing and find that these agents outperform \emph{off-the-shelf} LLM-based assistants for this task. Finally, we apply our agents in real-world meta-reviewing scenarios and confirm their effectiveness in enhancing the efficiency of meta-reviewing.\footnote{Code available at: \url{https://github.com/UKPLab/eacl2026-meta-review-as-dialog}}
\end{abstract}

\section{Introduction}

Peer review is the cornerstone of academic quality control across all scientific disciplines \cite{ware2015stm}. In a typical peer review process, a group of experts (the \textit{reviewers}) assesseses the paper at hand, summarizes their opinion in written reports (the \textit{reviews}), and then a meta-reviewer weighs the presented arguments to decide to \emph{accept} or \emph{reject} the paper for publication resulting in a \textit{meta-review} report.
This task requires significant effort and expertise by the meta-reviewer, as they need to match arguments, weigh different opinions, and contextualize the reviews within the broader scientific field. The growing load of research output~\cite{CHEN2022100001, Krenn_2023} puts an additional strain on meta-reviewers requiring to arrive at reliable decisions \textit{quickly}. While the assistance of reviewers through means of natural language processing (NLP) has received considerable attention~\cite[e.g.,][]{dycke-etal-2023-nlpeer, kuznetsov2024natural}, so far, the support for the decision-making process of meta-reviewers is under-investigated despite the apparent practical needs.

\begin{figure}[!t]
    \centering
    \includegraphics[width=0.4\textwidth]{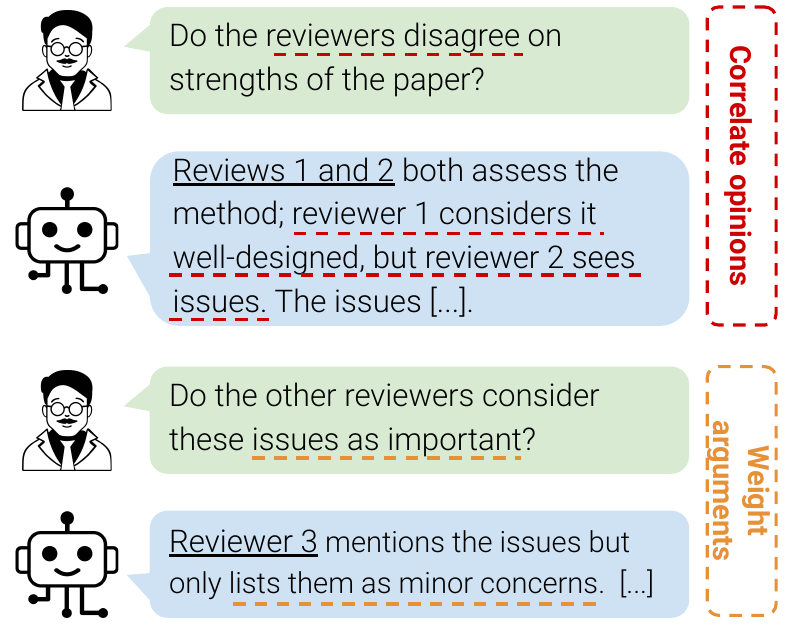}
    \caption{Illustration of the process of \textit{meta-reviewing as a dialogue}. Dialogues include requests to summarize opinions, weight arguments, and  contextualize them.} 
    \label{fig:meta-review}
\end{figure}

Meta-reviewing has traditionally been framed as a multi-document summarization task \cite{shen-etal-2022-mred, li-etal-2023-summarizing, zeng2023scientific}, aimed at generating meta-review reports by condensing the underlying reviews. While these summarization methods can aid meta-reviewers by simplifying the writing process, they often fall short of directly supporting decision-making. %Additionally, such approaches can sometimes introduce biases and errors~\cite{donker2023dangers, schintler2023critical}, potentially compromising the scientific autonomy of meta-reviewers. 
A more application-driven and human-centered approach to decision-making remains an important gap.

This paper advocates a new angle on assisting meta-reviewers that focuses on supporting the underlying \textit{decision-making process} to address these needs. For human decision-making and information assimilation over source materials, dialogue assistance has proven effective in many domains \cite{lever2011discussing, golinkoff2019language}. Following up on a recent proposal by \citet{balke5}, we explore the practical implications of developing a dialogue agent that assists the meta-reviewer during their step-wise decision-making process by collaboratively reflecting on the underlying peer reviews. Figure~\ref{fig:meta-review} illustrates an example of such a dialogue.
Ultimately, we investigate the following research questions. (RQ1) \textbf{What requirements does the meta-reviewing domain pose on a dialogue agent?} (RQ2) \textbf{How can we develop such a dialogue agent to account for these requirements?} (RQ3) \textbf{Is a dialogue agent practically useful during meta-reviewing?}

Developing a meta-reviewing decision-making dialogue assistant presents unique challenges. Unlike prior work in other domains on general topics~\cite{wang-etal-2019-persuasion, farag-etal-2022-opening}, a meta-reviewing dialogue agent must engage with highly technical topics and ensure its responses are strictly grounded in the underlying reviews. As shown in our experiments in Section~\ref{sec:synthetic_data}, off-the-shelf Large Language Models (LLMs) are inadequate for this task, as they fail to meet both these requirements. To bridge this gap, dedicated fine-tuning is crucial to tailor the models for technical grounding and decision-making tasks. Additionally, nuanced evaluation methods are required to ensure their reliability. However, fine-tuning and evaluating such agents is costly, as it demands significant effort to generate high-quality human dialogue data.

To address these issues we develop a dedicated fine-tuning and evaluation suite for meta-reviewing dialogue agents. First, we tackle the lack of training data by leveraging LLMs for \textbf{synthetic data generation}, which has shown promise across various fields \cite[e.g.,][]{møller2023prompt,chen-etal-2023-places}. Specifically, we simulate meta-review decision-making dialogues with LLMs such as ChatGPT~\cite{ChatGPT}
, LLaMa \cite{touvron2023llama}, Mistral \cite{jiang2023mistral}, and Mixtral \cite{jiang2024mixtral}. Through human and automated evaluations, we find that out-of-the-box LLMs generate dialogues that are unspecific and lack diversity. Therefore, we devise a new self-refinement strategy that addresses these challenges in the meta-reviewing domain but also show its generality for other decision-making domains. 
Afterwards, in Section~\ref{sec:agents} we utilize the synthetic dataset to \textbf{fine-tune dialogue agents} for meta-reviewing assistance and propose an evaluation strategy. Both automatic and human evaluations show that smaller, fine-tuned open-source models outperform larger, closed zero-shot LLMs like ChatGPT in generating relevant and high-quality responses.
Finally, in Section \ref{sec:time_analysis}, we \textbf{deploy the best-performing dialogue agent} within a human user study demonstrating the effectiveness of such a dialogue agent in time-efficient meta-reviewing. Our results show that dialogue agents can effectively \textbf{reduce the meta-reviewing time by almost up to 50\%}, while enabling the creation of more comprehensive and detailed meta-review reports, compared to when no agent is used. This demonstrates the potential of dialogue agents to significantly enhance both the quality and efficiency of the meta-reviewing process.\footnote{Overall workflow of this work is in Appendix~\ref{sec:end2end} } 

In summary, our contributions are three-fold: (\textbf{1}) We present the first study on meta-reviewing as a dialogue to-date and propose extensive measures towards developing dialogue agents for this scenario~(RQ1).
(\textbf{2}) We address the data scarcity issue for training dialogue agents via synthetic data generation using LLMs. We further propose a self-refinement strategy to enhance the generated dialogues using dialogue quality metrics as feedback~(RQ2).
(\textbf{3}) We utilize the synthetically generated data for training dialogue agents tailored to the meta-reviewing scenario and evaluate the utility of these agents in real-world meta-reviewing~(RQ3).\looseness=-1

\section{Meta-reviewing as a  Dialogue} \label{sec:perspective}
Existing NLP literature treats meta-reviewing exclusively as a summarization problem \cite{shen-etal-2022-mred, li-etal-2023-summarizing, zeng2023scientific}. While meta-reviewing involves summarization steps, it is primarily a decision-making process where the human meta-reviewer has to arrive at an overall recommendation on the paper based on the review reports. This decision-making process requires complex reasoning, including weighing, contextualizing, and meta-reasoning on the reviews. For instance, meta-reviewers need to adjust for reviewer uncertainty, biases, and review quality or need to consider the related work and state of the field. 
Summaries can assist in organizing information but may offer limited direct support for decision-making in meta-reviewing. This is because humans can be influenced by readily presented recommendations, a phenomenon known as \textit{automation bias}~\cite{goddard2014automation, schemmer2022influence}. In combination with the fact that LLMs sporadically introduce biases and errors in the summaries, this may impact human agency over the process~\cite{donker2023dangers, schintler2023critical}. Moreover, current ACL policies discourage the use of LLMs as ghostwriters during meta-reviewing~\cite{ACL23ReviewPolicies}.

In response to these issues, we propose a new \emph{human-centered} perspective on meta-reviewing assistance by framing meta-reviewing as a decision-making process that benefits from assistance through a dialogue agent. This concept has shown promise in related domains ~\cite{wang-etal-2019-persuasion, farag-etal-2022-opening} and in a piloting study by \citet{balke5}, who simulate this scenario between a human meta-reviewer and a human dialogue assistant.
More formally, we define meta-reviewing as a document-grounded decision-making dialogue where a dialogue agent provides information on the underlying reviews, and the meta-reviewer gains insights to move closer to their decision with each turn. In line with the ACL guidelines, such a dialogue agent should \textit{not} pro-actively bring up new facts but act on human inquiries to ensure full human agency. While the agent may still occasionally introduce biases or errors, the meta-reviewer can verify the information and enhance their understanding through iterative refinement~\cite{ke2024mitigating}.  
Specifically, following \citet{balke5}, the dialogue agent should \emph{ground} their answers in the reviews without hallucinating or obfuscating information. Secondly, the answers should be highly \emph{technically specific}, avoiding generic and unrelated statements. We refer to these two desiderata as \textsc{groundedness} and \textsc{specificity}.

Our experiments in Section~\S\ref{sec:synthetic_data} demonstrate the difficulty of this task. Out-of-the-box LLMs generate answers that fail on both desiderata. Thus, dedicated fine-tuning and nuanced evaluation of these agents are essential. However, data in this domain is costly as human experts and considerable effort is required to generate human-to-human example dialogues as proposed in \citet{balke5}. We explore these challenges in the remainder of this work.\looseness=-1

\begin{figure*}
    \centering
    \includegraphics[width =0.8\linewidth]{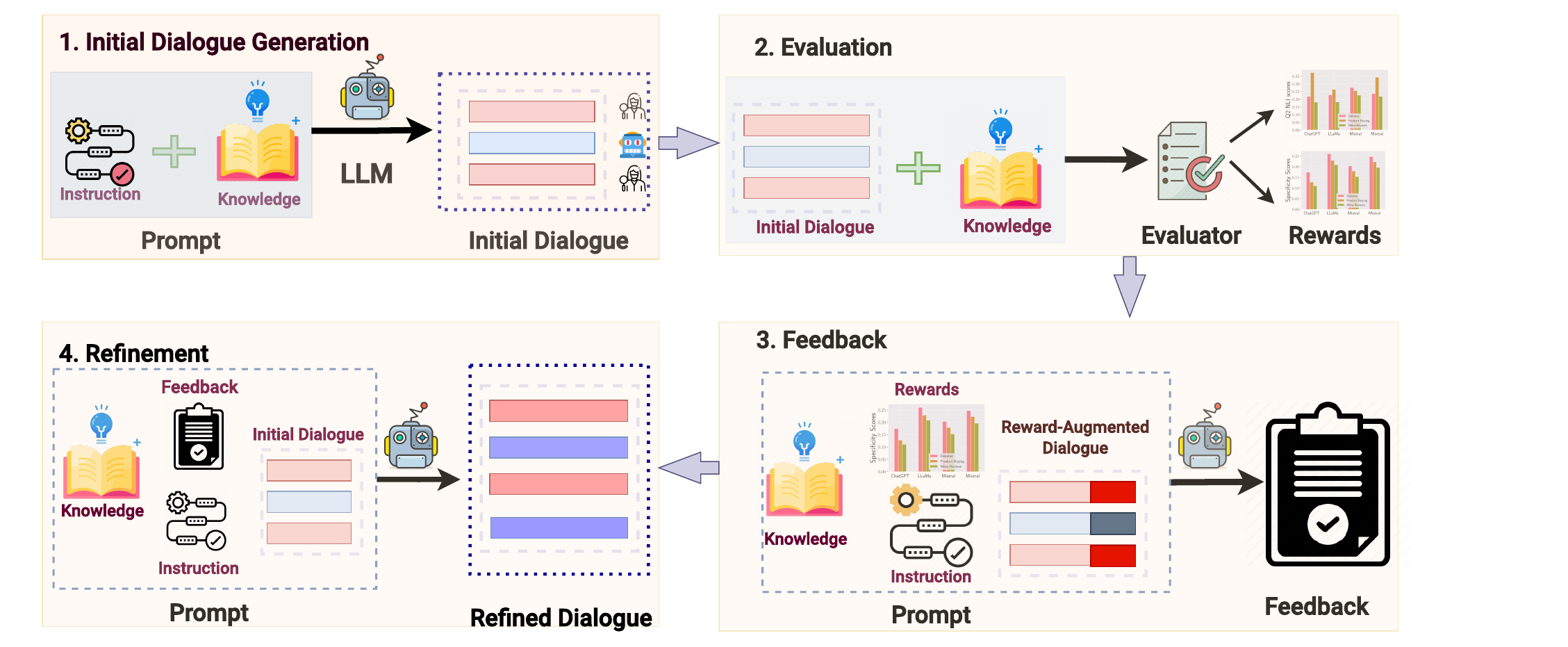}
    \caption{Overview of our \textbf{Re}ward-based \textbf{Mu}lti-aspect \textbf{S}elf \textbf{E}diting (ReMuSE) method. ReMuSE consists of four steps: 1. \textbf{Initial Dialogue Generation} in which we prompt an LLM with relevant documents (paper reviews) and instructions, 2. \textbf{Evaluation} of the dialogues by computing one or multiple measures (rewards), 3. Natural language \textbf{Feedback} Generation based on the computed rewards, 4. Self-\textbf{Refinement} of the dialogues based on the feedback.} 
    \label{fig:self-refine}
\end{figure*}

\section{Synthetic Data Generation for Fine-tuning Meta-review Dialogue Agents} \label{sec:synthetic_data}
Developing effective meta-reviewing dialogue agents necessitates dedicated fine-tuning, as off-the-shelf LLMs fall short of this specialized task (cf. Sec~\S\ref{sec:agents}). This fine-tuning process relies on high-quality training data tailored to meta-reviewing assistance. However, obtaining expert annotators and producing human gold data is resource-intensive since crowd-sourcing is not a viable option~\cite{stasaski-etal-2020-cima}. To address these limitations, leveraging LLMs to generate synthetic data has become a widely adopted approach in expert-driven domains such as education~\cite{macina-etal-2023-mathdial} and biomedicine~\cite{10.1093/jamiaopen/ooae114}. Following this trend, we employ LLMs to \emph{synthetically generate} a dataset of meta-reviewing dialogues, building on their proven utility in various tasks~\cite{møller2023prompt,chen-etal-2023-places}. However, meta-reviewing dialogues are particularly challenging to synthesize, as they need to account for the previously mentioned desiderata of \textsc{groundedness} and \textsc{specificity}. Our initial experiments show that out-of-the-box LLMs fail to generate dialogues that meet these criteria. To overcome this, we implement a self-refinement-based approach for data generation. In the following, we present our data generation approach (\S\ref{sec:approach}), assess dataset quality (\S\ref{sec:metrics}), fine-tune models (\S\ref{sec:agents}), and conduct a within-subject evaluation (\S\ref{sec:time_analysis}).\looseness=-1

\subsection{Approach} \label{sec:approach}

We propose ReMuSE (\textbf{Re}ward-based \textbf{Mu}lti-aspect \textbf{S}elf-\textbf{E}diting) based on the self-refinement paradigm ~\cite{madaan2023self, chen2023iterative}.
In this paradigm, an LLM $\mathcal{M}$ first produces the initial output $o$. Then, feedback $f$ is obtained on the output via $\mathcal{M}$ or other feedback mechanisms. The model $\mathcal{M}$ is then prompted to correct the initial output $o$ based on $f$.
All prior approaches~\cite{madaan2023self, chen2023iterative} either improve a single aspect of the output only, or use ground truth in-context exemplars to steer the generations towards a particular format. These factors restrict the diversity of the corrections to a small set of specific changes rather than profound revisions~\cite{wang2023enable}. However, as our experiments show (cf. Section~\S\ref{sec:metrics}), meta-reviewing dialogue generation gains from incorporating multiple aspects during the feedback loop of self-refinement. In this work, we thus propose a zero-shot refinement strategy that makes use of scoring functions to calculate dialogue quality metrics. These \textit{rewards} are used as feedback to improve the initial output across multiple dimensions.

ReMuSE generates dialogues based on a knowledge source, e.g. the collection of reviews. Given a knowledge source $\mathcal{K}$ and an LLM $\mathcal{M}$ ReMuSE operates in four steps (cf. Figure~\ref{fig:self-refine}):

\noindent \textbf{I. Initial Dialogue Generation} We prompt $\mathcal{M}$ with knowledge source $\mathcal{K}$ to generate an initial dialogue $\mathcal{D}$ in a zero-shot fashion. 
\vspace{0.2em}

\noindent \textbf{II. Evaluation} The generated dialogue is then evaluated to obtain rewards using an \textbf{evaluator}. The evaluator consists of various scoring functions. For example, one of the functions in our experiment calculates the token overlap-based groundedness score of utterance, $u$ with respect to $\mathcal{K}$. 
These rewards are concatenated to every utterance of $\mathcal{D}$, producing the reward-augmented dialogue $\mathcal{D'}$.\footnote{The evaluator is completely independent of $\mathcal{M}$.}
\vspace{0.2em}

\noindent \textbf{III. Feedback} $\mathcal{D'}$ and $\mathcal{K}$ are fed into $\mathcal{M}$ to obtain natural language feedback $\mathcal{F}$ on improving the rewards. Following \citet{madaan2023self}, we use the phrase `actionable feedback' in the prompt to obtain a concrete plan to improve the scores.
\vspace{0.2em}

\noindent \textbf{IV. Refinement} Finally, we feed the knowledge $\mathcal{K}$, the feedback $\mathcal{F}$, and the initial dialogue $\mathcal{D}$ to $\mathcal{M}$. The model is prompted to refine the dialogues based on the feedback provided.\looseness=-1

\subsection{Dataset Generation} \label{sec:data}
We generate a dataset of meta-reviewing dialogues for fine-tuning dialogue agents. For this purpose, we test several models and perform automated and human evaluations to select the best configuration.
\vspace{0.3em}

\noindent \hl{ \textbf{Models}} \label{sec:llm}
We use the following models namely, GPT 4o (dubbed `ChatGPT'), LlaMa-2 13B~\cite{touvron2023llama}, Mistral 7B~\cite{jiang2023mistral}, and Mixtral~\cite{jiang2024mixtral}.\footnote{Details about the models and prompts in \S\ref{sec:model_link} and \S\ref{sec:prompts}.}
\vspace{0.2em}

\noindent \hl{\textbf{Underlying Data} }\label{sec:datasets}
To generate a dialogue, ReMuSE receives a paper and its reviews as the input. As a basis to generate a dataset, we rely on the ORSUM~\cite{zeng2023scientific} dataset consisting of reviews and meta-reviews for papers from $39$ conferences collected using OpenReview.\footnote{There are no explicit licenses attached to OpenReview for third-party data processing. We use the dataset from the author's publicly available codebase.} We sample papers with exactly three reviews to fit most LLMs' context windows resulting in $\mathbf{3064}$ dialogues.
\vspace{0.3em}

\noindent \hl{\textbf{Reward Metrics}}
In the absence of ground truth dialogues to assess the quality of the synthetic dialogues, we use three reference-free metrics as rewards to assess the \textsc{Specificity} and \textsc{Groundedness} of generated dialogues. We use three feedback metrics as rewards to assess the specificity and groundedness of generated dialogues. \hldb{Knowledge Precision (K-Prec)} measures token overlap between an utterance and its knowledge source \citep{adlakha2023evaluating}. Because K-Precision cannot capture the semantic similarity between utterances, we additionally employ a semantics-based metric, \hldb{Q2-score}~\citep{honovich-etal-2021-q2}. This metric evaluates faithfulness via question generation, answer generation, and natural language inference, leveraging pre-trained T5 \citep{JMLR:v21:20-074}, ALBERT-Large \citep{Lan2020ALBERT:}, and DeBERTa-Large-MNLI \citep{he2021deberta} models, respectively, to capture dialogue groundedness. \hldb{Specificity} scores the level of detail in utterances using Specifiteller \citep{li2015fast}.

\noindent \hl{\textbf{Resulting Datasets}}
Using each LLM, we generate synthetic dialogues. Due to the associated API costs for using closed LLMs, we follow previous works~\cite{ bian2024chatgpt, chen2024reconcile} that rely on a lower number of examples (e.g., $100$) for evaluation. Consequently, we sample $200$ examples to analyze the performance of ReMuSE and publicly release $\mathbf{3064}$  meta-reviewing dialogues generated with the best ReMuSE configuration.\looseness=-1

\noindent \hl{\textbf{Baselines.}} We consider several baselines to compare our method against. (i) \textbf{Zero-shot}: LLM is prompted in a 1-pass setup to generate dialogues. (ii) \textbf{Self-Refine}~\cite{madaan2023self}: the LLM is prompted to refine the generations for the same number of iterations as in our approach, ReMuSE. (iii) \textbf{Self-Consistency and Refinement}~\cite{wang2023selfconsistency, chen2024reconcile}: The LLM first refines multiple sampled candidates and then the best candidate is selected by a majority vote on the reward criteria.\looseness=-1
\subsection{Dataset Evaluation} \label{sec:metrics}
We discuss various dialogue quality metrics employed in automated and human evaluation reflecting the desiderata for \textsc{groundedness} and \textsc{specificity}. We compare the synthetically generated dialogues to a human-generated dialogue dataset that we use as evaluation data for comparison along the various evaluation setups.\looseness=-1

\subsubsection{Human Dialogue Dataset} 
We compare our synthetically generated dialogues with real human meta-reviewing decision-making dialogues using the \textit{Dagstuhl} dataset \cite{balke5}. In this dataset, senior and junior scientists acted as meta-reviewers and assistants, respectively, in a wizard-of-oz setup, where the meta-reviewers used dialogue agents to gather information and decide on paper acceptance. The dataset is private and consists of 16 dialogues corresponding to 4 research papers.\footnote{Additional details about the dataset is in \S\ref{sec:dagastuhl}}\looseness=-1
%For the comparison of our synthetically generated dialogues with real human meta-reviewing decision making dialogues we employ the \textit{Dagstuhl} dataset \cite{balke5} kindly provided by the authors.
%For this dataset, $4$ senior and $4$ junior scientists acted as meta-reviewers and assistants, respectively, in a wizard-of-oz scenario. The study was conducted on $4$ papers selected from the \textsc{NLPeer} corpus \cite{dycke-etal-2023-nlpeer}. The assistants were provided with the paper and reviews in advance, whereas the meta-reviewers were instructed to skim the reviews within 10 minutes. The meta-reviewers were asked to converse with the dialogue agents to gather information and eventually decide on the acceptance of the paper.

\subsubsection{Automatic Evaluation}
%\todo{revise intro, point to previous section on metrics}
\hl{\noindent \textbf{Setup}} We compute K-Prec, Q2-score, Specificity automatically given the knowledge source $\mathcal{K}$, and a dialogue utterance $\mathit{u}$, part of the generated dialog. These metrics are also employed as \textbf{rewards} within ReMuSE. The evaluation is performed on an utterance level. The intuition behind using the same measures for automated evaluation and for refinement is to understand whether ReMuSE can, in fact, guide the LLMs towards better generations reflected in exactly these dimensions we optimize for. We complement the automatic evaluation with our human evaluation to verify the improved quality.

For the final evaluation, we compute the K-Prec and Q2-scores only over the dialogue agent's utterances since we require only the dialogue agent to be faithful. The meta-reviewer (as an information seeker) need not fulfill such criteria since they can bring up new facts independent of the knowledge source at any point in time~\cite{faithdial}. Nevertheless, all the utterances are supposed to be highly specific. This stems from our observation that $65.24\%$ utterances in the human dataset (cf. \S\ref{sec:datasets}) have a score of $0.5$ or above, indicating a significant presence of technical details. Hence, we measure the Specificity score over all utterances.

\begin{table}[!t]
\centering
\resizebox{0.42\textwidth}{!}{
\begin{tabular}{llllll}
\hline
 \textbf{Model} & \textbf{Method} & \textbf{K-Prec} & \textbf{Q2-F1} & \textbf{Q2-NL} & \textbf{Spec.} \\ \hline
ChatGPT        & Zero-Shot       & $0.34_{.01}$    & $0.17_{.01}$   & $0.17_{.01}$   & $0.46_{.13}$ \\
LLaMa          & Zero-Shot       & $0.37_{.01}$    & $0.15_{.13}$   & $0.16_{.01}$   & $0.46_{.01}$ \\
Mistral        & Zero-Shot       & $0.38_{.01}$    & $0.15_{.02}$   & $0.16_{.01}$   & $0.41_{.02}$ \\
Mixtral        & Zero-Shot       & $0.38_{.01}$    & $0.17_{.01}$   & $0.19_{.02}$   & $0.42_{.02}$ \\ \cline{1-6}

ChatGPT        & Self Ref. & $0.46_{.02}$   & $0.19_{.01}$   & $0.19_{.01}$   & $0.52_{.02}$ \\
LLaMa          & Self Ref. & $0.48_{.02}$   & $0.18_{.02}$   & $0.19_{.01}$   & $0.53_{.02}$ \\
Mistral        & Self Ref. & $0.49_{.01}$   & $0.19_{.01}$   & $0.20_{.01}$   & $0.50_{.02}$ \\
Mixtral        & Self Ref. & $0.50_{.01}$   & $0.20_{.01}$   & $0.21_{.02}$   & $0.51_{.02}$ \\ \cline{1-6}

ChatGPT        & Self Con. + Ref.     & $0.52_{.02}$   & $0.21_{.02}$   & $0.22_{.02}$   & $0.55_{.02}$ \\
LLaMa          & Self Con. + Ref.     & $0.53_{.02}$   & $0.21_{.01}$   & $0.22_{.02}$   & $0.56_{.02}$ \\
Mistral        & Self Con. + Ref.     & $0.55_{.01}$   & $0.23_{.02}$   & $0.25_{.01}$   & $0.58_{.02}$ \\
Mixtral        & Self Con. + Ref.     & $0.56_{.01}$   & $0.24_{.01}$   & $0.26_{.02}$   & $0.59_{.02}$ \\ \cline{1-6}

ChatGPT        & ReMuSE          & $0.76_{.12}$   & $0.24_{.02}$   & $0.24_{.18}$   & $0.61_{.02}$ \\
LLaMa          & ReMuSE          & $0.74_{.02}$   & $0.23_{.02}$   & $0.24_{.02}$   & $0.63_{.02}$ \\
Mistral        & ReMuSE          & $0.76_{.01}$   & $0.30_{.02}$   & $0.33_{.02}$   & $0.65_{.22}$ \\
Mixtral        & ReMuSE          & $\textbf{0.78}_{.01}$ & $\textbf{0.32}_{.01}$ & $\textbf{0.34}_{.02}$ & $\textbf{0.72}_{.02}$ \\ \hline              
\end{tabular}}
\caption{Performance with and without the refinement methods (Self-Consistency, Self-Refine, ReMuSE) in terms of K-Prec, Q2-F1, Q2-NLI, and Specificity (Spec) averaged over three prompts with standard deviation.}
%\vspace{-2em}
\label{tab:ReMuSE}
\end{table}

\noindent \hl{\textbf{Overall results and comparison to human dialogues}} We compare LLM performance across multiple baselines in Table~\ref{tab:ReMuSE}. \textbf{ReMuSE consistently improves all models and metrics}, with Mixtral achieving the best results. The largest gains are in \textbf{K-Precision} (+105\%), with substantial improvements in \textbf{Q2-F1 ($\sim$88\%)}, \textbf{Q2-NLI ($\sim$79\%)}, and \textbf{specificity ($\sim$72\%)}. \textbf{Self-Refinement} offers modest improvements, reflecting LLM limitations in self-correction~\cite{huang2024large}. Self-Consistency with refinement improves performance of the models, however the majority vote mainly selects the longest dialogue reflecting length bias in LLMs~\cite{hu2025explaininglengthbiasllmbased}.\footnote{Table~\ref{tab:ReMuSE_all} reports results for ReMuSE with all rewards.}\footnote{ \S\ref{sec:ablation} for ablations and \S\ref{sec:result_all} for cross-domain analysis.} We further assess \textbf{dialogue modeling} with the best performing model, Mixtral, dividing meta-review dialogues into \textbf{evidence gathering, conflict resolution, and decision-making stages} following prior approaches on task-specific deliberation~\cite{priya2025argueagreepersonalitydrivenargumentationbased}. ReMuSE significantly outperforms other strategies across stages, e.g., \textbf{$\sim$77\% vs. 28\%} for conflict resolution, demonstrating its effectiveness in structured meta-review dialogues.\footnote{Full experimental details in Sec~\S\ref{sec:dialog_model}. Sec~\S\ref{sec:dial_assist} lists dialogue excerpts exemplifying these stages.}\footnote{Further results using other open-source models in Sec~\S\ref{sec:other_open}}

\begin{figure}[!t]
\centering
    \begin{subfigure}[b]{0.20\textwidth}
    \centering
        \includegraphics[width=\linewidth]{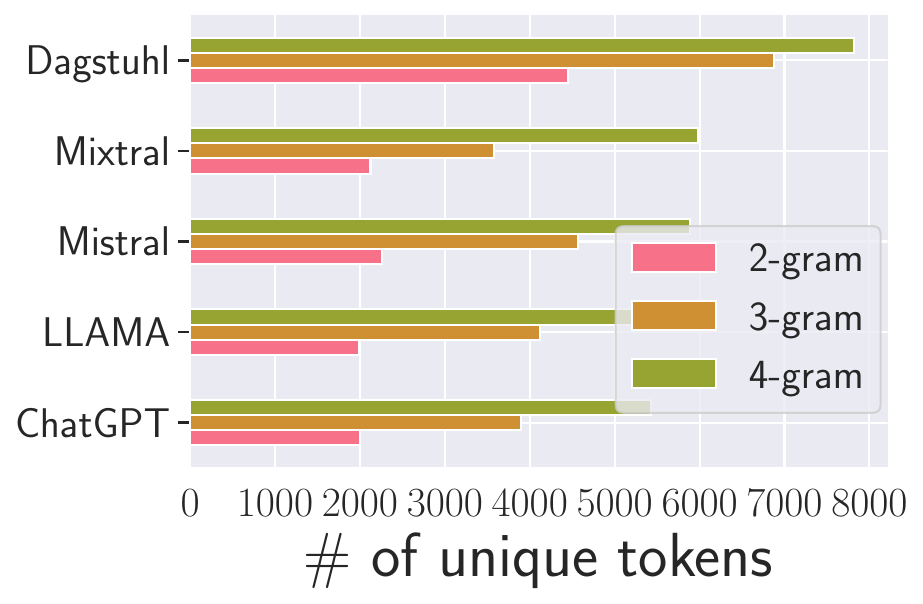}%
        \caption{Token Distribution}
        \label{fig:token_dist}
        \end{subfigure}%
        \begin{subfigure}[b]{0.20\textwidth}
        \includegraphics[width=\linewidth]{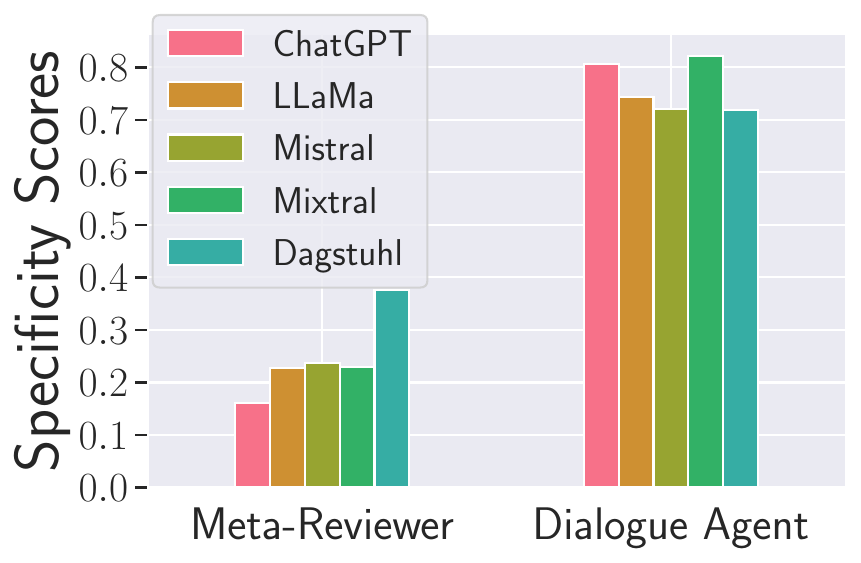}
        \caption{Specificity}
        \label{fig:spec_dagstuhl}
        \end{subfigure}%

        \begin{subfigure}[b]{0.20\textwidth}
         \includegraphics[width=\linewidth]{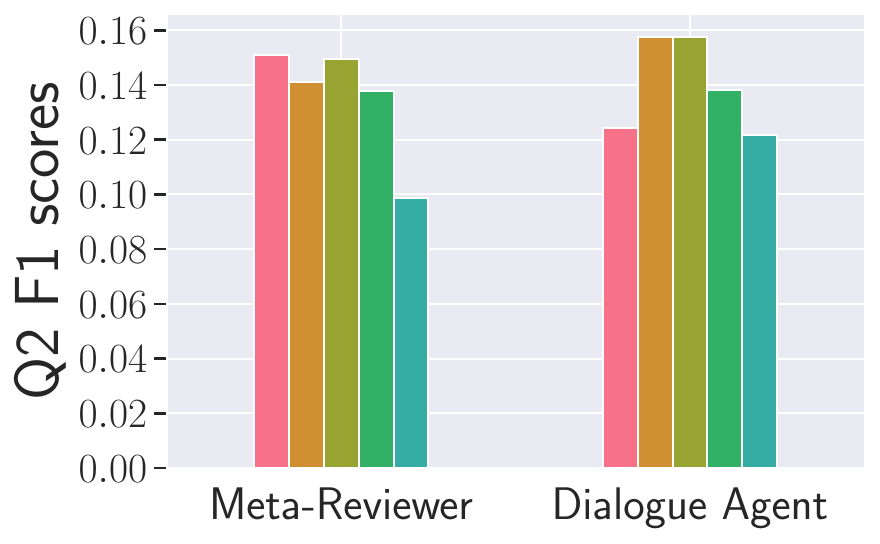}
        \caption{Q2 F1}
        \label{fig:q2_dagstuhl}
        \end{subfigure}%
        \begin{subfigure}[b]{0.20\textwidth}
        \includegraphics[width=\linewidth]{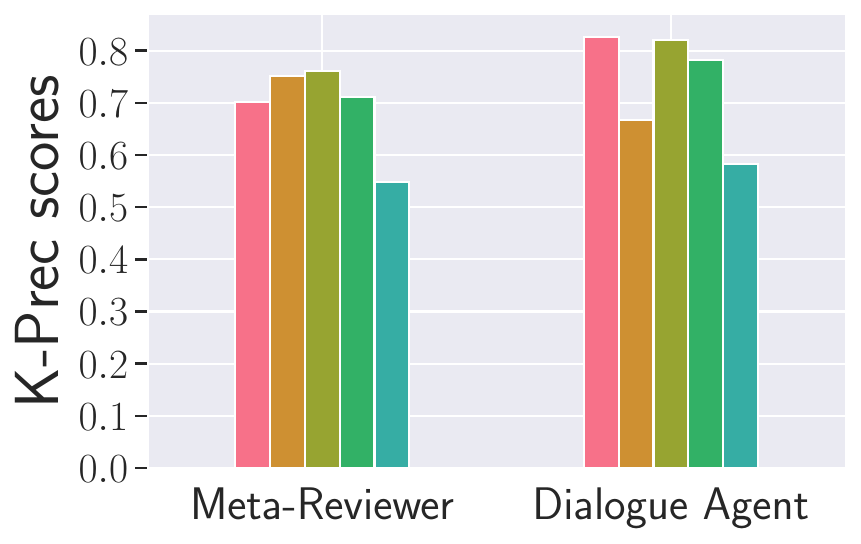}
        \caption{K-Prec}
        \label{fig:k_prec_dagstuhl}
    \end{subfigure}
    \caption{Comparison of utterances in human and synthetically generated dialogues in terms of (a) Token Distribution,  (b) Specificity, (c) Q2 F1, and (d) K-Prec.}
    %\vspace{-2mm}
\label{fig:llm_vs_human_meta_reviewing}
\end{figure}

\iffalse
\setlength{\tabcolsep}{6pt}
\begin{table}[!t]
\centering
\resizebox{0.34\textwidth}{!}{\begin{tabular}{llllll}
\hline
\textbf{Mode} & \textbf{Models} & \textbf{Coop.} & \textbf{Coh.} & \textbf{Eng.} & \textbf{Plau.} \\ \hline
\multirow{4}{*}{\begin{tabular}[c]{@{}c@{}}Pre-\\ Refine.\end{tabular}} 
& ChatGPT   & 2.65 & 2.71 & 2.74 & 2.15 \\
& LLaMa     & 2.53 & 2.13 & 2.11 & 2.45 \\
& Mistral   & 2.34 & 2.22 & 2.18 & 2.19 \\
& Mixtral   & 2.35 & 2.23 & 2.26 & 2.45 \\ \hline
\multirow{4}{*}{\begin{tabular}[c]{@{}c@{}}Post-\\ Refine.\end{tabular}} 
& ChatGPT   & 3.01 & \textbf{2.91} & \textbf{2.93} & 2.75 \\
& LLaMa     & 3.17 & 2.51 & 2.32 & 3.06 \\
& Mistral   & 3.28 & 2.63 & 2.41 & 3.53 \\
& Mixtral   & \textbf{3.37} & 2.79 & 2.85 & \textbf{3.58} \\ \hline
\end{tabular}}

\caption{Human Evaluation of the generated dialogues before and after-refinement across different dimensions. All the scores are on a scale of $1$ (worst) $- 4$ (best).
}
\label{tab:human_eval}
\end{table}
\fi
In order to compare \hldb{human vs synthetic dialogues}, we analyze K-Prec, Q2, and Specificity scores for all utterances in human and synthetic meta-reviewing dialogue datasets, along with \{2,3,4\}-gram token diversities (\{\textit{n}\}-gram vocabulary size) for meta-reviewer utterances in Fig~\ref{fig:llm_vs_human_meta_reviewing}. LLM-based dialogue agents outperform humans in Q2 and K-Prec scores (cf. Figures~\ref{fig:q2_dagstuhl} and \ref{fig:k_prec_dagstuhl}), demonstrating superior groundedness to the knowledge source. Human meta-reviewers exhibit lower groundedness, possibly due to off-topic divergence: for instance, in one of the dialogues, the meta-reviewer discusses datasets available for a related task not mentioned in the reviews. Regarding Specificity (Fig~\ref{fig:spec_dagstuhl}), LLM-based agents perform comparable to humans, indicating their feasibility as meta-reviewing assistants. However, a significant gap exists between real and LLM meta-reviewers in terms of token diversity and specificity (cf. Fig~\ref{fig:token_dist} and \ref{fig:spec_dagstuhl}), suggesting LLMs still lack the expert knowledge for simulating real-world meta-reviewers. \looseness=-1

\begin{table}[!t]
\centering
\resizebox{0.4\textwidth}{!}{
\begin{tabular}{llllll}
\hline
\textbf{Method} & \textbf{Coop.} & \textbf{Coh.} & \textbf{Eng.} & \textbf{Plau.} \\ \hline
Zero-Shot        & 2.35 & 2.23 & 2.26 & 2.45 \\
Self Ref.  & 2.70 & 2.45 & 2.50 & 2.75 \\
Self Con. + Ref.      & 3.00 & 2.60 & 2.70 & 3.05 \\
ReMuSE           & \textbf{3.37} & 2.79 & 2.85 & \textbf{3.58} \\ \hline
\end{tabular}}
\caption{Human evaluation of the best performing model, Mixtral across different refinement methods. Scores are on a scale of $1$ (worst) to $4$ (best).}
\label{tab:human_eval}
%\vspace{-6.5mm}
\end{table}

\subsubsection{Human Evaluation}
\hl{\noindent \textbf{Setup} }\label{sec:human_eval_metrics} Since using identical metrics for rewards and automated evaluation can be questioned for coincidental correlation, we further run a human evaluation. Following \cite{faithdial,wu-etal-2022-dg2}, we let humans evaluate the generated dialogs according to the following criteria: \textbf{Co-operativeness} evaluates the helpfulness of the dialogue agent's response to the query, \textbf{Coherence} specifies consistency of the overall dialogue, and \textbf{Engagingness} implies how engagingly the dialogue agent takes the discussion forward. Moreover, akin to faithfulness discussed before, we introduce a \textbf{Plausibility} score to measure the groundedness of the dialogue agent's responses to the reviews. All the dialogues are rated on a Likert scale of $1-4$ and the ratings are averaged over each metric as done in prior works~\cite{rashkin-etal-2021-increasing}.\footnote{The instructions to the annotators are in \S\ref{sec:hum_ins_full_dial}}\looseness=-1

\begin{figure}
    \centering
    \includegraphics[width=0.7\linewidth]{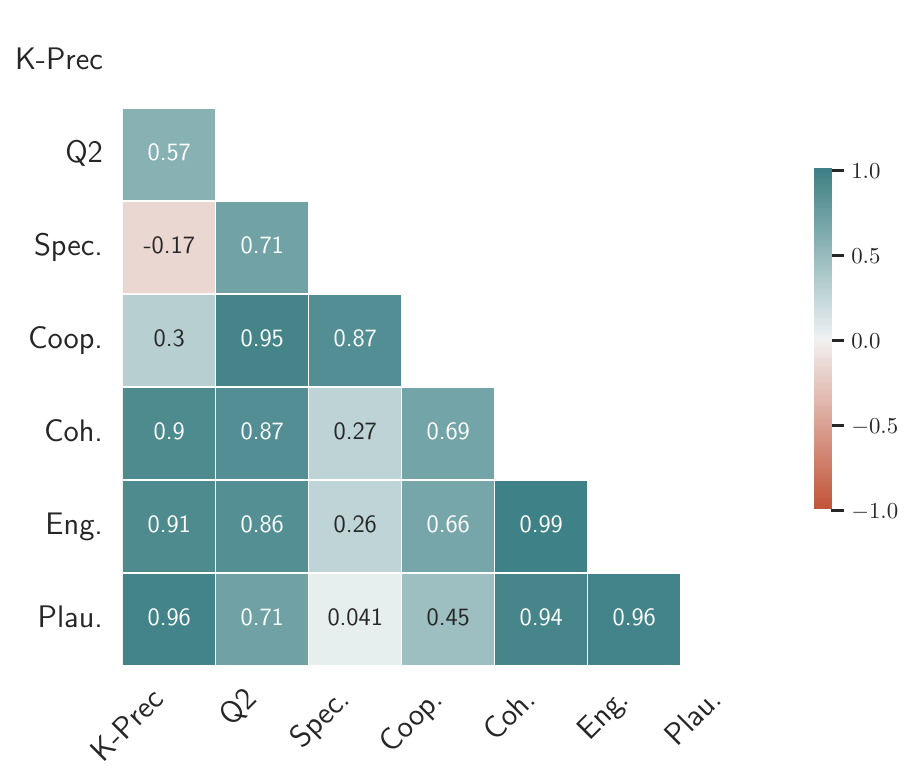}
    \caption{Correlation between human and automated evaluation metrics.}
    \label{fig:human_auto}
\end{figure}

\noindent \hl{ \textbf{Overall results and comparison to human dialogues}}\label{sec:human_eval}
Four Ph.D. students with NLP backgrounds evaluated 70 dialogues, split between the different baseline models and ReMuSE (post-refinement) with the best performing model, Mixtral using human evaluation metrics on a 1-4 Likert scale. Krippendorff’s $\alpha$ for the dialogues showed substantial agreement: co-operativeness (0.62), coherence (0.74), engagingness (0.74), and plausibility (0.76). We report the results in Table~\ref{tab:human_eval}. We observe that dialogues generated with ReMuSE are adjudged to be the best across all metrics. We observe a similar ranking of baselines as in our automated evaluation with zero-shot being the least effective method for dialogue generation. We further conducted correlation analysis (cf. Fig~\ref{fig:human_auto},\S\ref{sec:corr_human_auto}) of the automated and human evaluation metrics which revealed strong alignment between the metrics with similar objective (e.g., K-Prec and Plausibility), validating our evaluation approach.%We observe that post-refinement, all models improved, with Mixtral excelling in co-operativeness and plausibility consistent with the automated evaluation. However, ChatGPT excels in coherence and engagingness likely due to its reported use of convincing language in responses~\cite{fi15100336}. We further conducted correlation analysis (cf. Fig~\ref{fig:human_auto},\S\ref{sec:corr_human_auto}) of the automated and human evaluation metrics which revealed strong alignment between the metrics with similar objective (e.g., K-Prec and Plausibility), validating our evaluation approach.\looseness=-1%\footnote{We show some of such anecdotal examples in Appendix~\S\ref{sec:quant_eval}.}

Three NLP Ph.D. students evaluated the \hldb{synthetic and human-generated dialogues} on a $1$–$4$ scale (Sec.~\S\ref{sec:human_eval_metrics}). Krippendorff's $\alpha$ for co-operativeness, coherence, engagingness, and plausibility were $0.68$, $0.69$, $0.70$, and $0.74$, respectively. As shown in Table~\ref{tab:dagstuhl_vs_remuse}, synthetic dialogues scored higher for co-operativeness, coherence, and plausibility, while human dialogues scored higher for engagingness. Lower human ratings may stem from off-topic responses and grammatical errors, whereas engagingness benefited from diverse meta-reviewer questions producing longer, detailed responses, consistent with prior findings~\citep[e.g.,][]{wu-etal-2022-dg2}.

We evaluate generated dialogues using automated metrics for \textsc{Groundedness} and \textsc{Specificity}, and human assessments for flow and naturalness. Our results show LLMs are promising as meta-reviewing assistants, with high faithfulness and specificity, though low token diversity highlights room for improvement as meta-reviewers. Section~\S\ref{sec:agents} discusses how fine-tuning can address these limitations.\looseness=-1

%As one of the contributions of this work, we make the synthetically generated dataset for the best-performing ReMuSE configuration publicly available. The resulting dataset consists of $3,064$ meta-reviewing dialogues.\footnote{Detailed analysis of the dataset is in Appendix~\S\ref{sec:decidial}}  

\begin{table}[!t]
\centering
\resizebox{0.3\textwidth}{!}{\begin{tabular}{ccccc}
\hline
\textbf{Setup} & \textbf{Coop.} & \textbf{Coh.} & \textbf{Eng.} & \textbf{Plau.} \\ \hline
Dagstuhl       & 3.50             & 3.10                 & \textbf{3.10}            & 2.95                  \\
ReMuSE         & \textbf{3.52}**             & \textbf{3.86}**     & 3.00**                  & \textbf{3.10}**     \\ \hline      
\end{tabular}}

\caption{Human evaluation of dialogues from the human dialogue dataset (\textit{Dagstuhl}) and our proposed ReMuSE. The evaluation scale ranges from $1$ (worst) - $4$ (best). **(p$\leq$.001) statistically significant results}
\label{tab:dagstuhl_vs_remuse}
\end{table}
\iffalse
\begin{table}[t]
\centering
 \small{\begin{tabular}{lllll}
\hline
\textbf{Domain} & \textbf{\begin{tabular}[c]{@{}l@{}}\# dial.\end{tabular}}& \textbf{\begin{tabular}[c]{@{}l@{}}\# DA. \\ tok. (av.)\end{tabular}} & \textbf{\begin{tabular}[c]{@{}l@{}}\# Seek. \\ tok. (av.)\end{tabular}} & \textbf{\# tur.} \\ \hline
Meta-Rev.     &\textbf{3,064}    & \textbf{54.66}                                                                       & \textbf{30.95}                                                                       & \textbf{13.24}                 \\
Product-Buy.  & 123         & 28.87                                                                       & 13.49                                                                      & 11.92                 \\
Debates  & 108       & 48.51                                                             & 18.16                                                              & 10.29                \\
%Mixtral         & 59.63                                                                       & 25.44                                                                       & 10.52                \\ 
\hline
\end{tabular}}
\caption{Basic statistics for the developed dataset. The table shows the number of dialogues, the average number of tokens for the dialogue agent and seeker utterances, and the number of turns.}
\label{tab:dataset}
\end{table}
\fi

\looseness=-1

%\vspace{0.5em}

\section{Dialogue Agents for Meta-Reviewing Assistance} \label{sec:agents}
With the synthetically generated dataset for the best-performing ReMuSE configuration, we now shift our focus to the task of finetuning dialogue agents for meta-reviewing assistance.\footnote{See \S\ref{sec:decidial} and \S\ref{sec:examples} for the synthetic dataset and example dialogues.} \looseness=-1
\subsection{Experimental Setup}
\hl{\noindent \textbf{Task}} Given a knowledge source $\mathcal{K}$ and a dialogue history $[u_1,u_2,\dots,u_t]$, the task is to generate a response $u_{t+1}$ grounded in $\mathcal{K}$. The title and reviews for every paper serve as the knowledge source.
\vspace{0.2em}

\noindent \hl{\textbf{Models}} Following \citet{daheim2023elastic}, we select Flan-T5~\cite{pmlr-v202-longpre23a} (xl), T5~\cite{JMLR:v21:20-074} (3B), and OPT~\cite{zhang2022opt} (2.7B) for experiments. For further comparison, we also prompt ChatGPT in a 0-shot setup and use the best model from the ReMuSE setup, Mixtral.\footnote{The hyper-parameter details are provided in \S\ref{sec:hyper-parameters}}
\vspace{0.2em}
\noindent \hl{\textbf{Data and Evaluation} }
We split the $\mathbf{3064}$ meta-reviewing dialogues having an average of $\textbf{12.24}$ utterances per dialogue in our dataset into $60\%$ for training, $20\%$ for validation, and $20\%$ for testing, respectively. In terms of evaluation, following \citet{daheim2023elastic}, we use automated metrics - SacreBLEU \citep{post2018clarity} and BERTScore \citep{Zhang*2020BERTScore:} to measure response similarity to ground truth. We also report K-Prec and Q2 scores to assess faithfulness to $\mathcal{K}$. Additionally, following \citet{rashkin-etal-2021-increasing}, we perform human evaluation on fluency, relevance, faithfulness, and objectivity, plus a helpfulness score reflecting decision-making utility. All human ratings use a 1–5 scale, and the ratings are averaged over each metric.\looseness=-1

\begin{table}[!t]
\centering
\resizebox{0.35\textwidth}{!}{\begin{tabular}{llllll}
\hline
\multirow{2}{*}{\textbf{Models}} & \multicolumn{2}{c}{($y,\hat{y}$)}                           & \multicolumn{3}{c}{($y,\mathcal{K}$)}                         \\ \cmidrule(lr){2-3} \cmidrule(lr){4-6}
                        & \multicolumn{1}{c}{\textbf{BL.}} & \multicolumn{1}{c}{\textbf{BT Sc.}} & \multicolumn{1}{c}{\textbf{K-Prec}} & \multicolumn{1}{c}{\textbf{Q2-F1}} & \multicolumn{1}{c}{\textbf{Q2-NL}} \\ \cmidrule(lr){1-1} \cmidrule(lr){2-3} \cmidrule(lr){4-6}
OPT                    & 15.4                    & 54.5                         & 57.6                      & 36.2  & 40.2                \\
T5                      & 18.4                    & 58.1                         & 63.4                      & 39.1    &  45.4            \\
Flan-T5                 & \textbf{20.4}                    & \textbf{60.3}                         & \textbf{68.2}                      & \textbf{42.2}  & \textbf{47.3}                \\ \hdashline
ChatGPT & 12.1 & 45.0 & 40.1 & 23.1 & 22
.4 \\ 
%ReMuSE + K-Prec         & 13.5 & 47.0 & 65.8 & 40.4 & 46.2 \\
%ReMuSE + Q2             & 13.7 & 47.2 & 65.1 & 41.1 & 46.6 \\
%ReMuSE + Spec.          & 13.9 & 47.5 & 66.9 & 41.3 & 45.9 \\
%ReMuSE + Spec. + K-Prec & 14.2 & 48.1 & 66.0 & 41.5 & 46.5 \\
%ReMuSE + K-Prec + Q2    & 14.3 & 48.4 & 67.2 & 42.0 & 46.8 \\
%ReMuSE + Q2 + Spec      & 14.1 & 48.2 & 67.4 & 42.1 & 46.9 \\
ReMuSE            & \underline{14.5} & \underline{49.0} & \underline{67.6} & \underline{42.2} & \underline{47.1}
\\
\hline
\end{tabular}}
\caption{Performance of models in the response generation task. BLEU (BL.) and BERT Score (BT Sc.) measure the similarity of the response with the ground truth. K-Prec and Q2 scores evaluate the factual consistency of the response with the Knowledge Source.}
\label{tab:dialogue_agent}
\end{table}

\subsection{Results and Discussion} \label{sec:pre_refinement}
\begin{table}[!t]
\centering
\resizebox{0.35\textwidth}{!}{\begin{tabular}{llllll}
\hline
\textbf{Models} & \textbf{Fl.} & \textbf{Rel.} & \textbf{Faith.} & \textbf{Obj.} &\textbf{Help.} \\ \hline
OPT       &       3.82       &   3.84               &      4.04      &   4.06 &    4.21          \\
T5         &    3.91        &  3.98    &    4.12             &   4.21 &  4.32\\     
Flan-T5 &    \textbf{4.21}          &  \textbf{4.18}    &   \textbf{4.58}               &  \textbf{4.62} &  \textbf{4.71} \\   \hdashline 
ChatGPT & 3.80 & 3.80 & 3.24 & 3.87 & 3.82 \\  
ReMuSE & 3.92 & 3.94 & 4.11 & 4.11 & 4.24 \\ \hline
%\hdashline

%Ground-truth &  4.89           &  4.72   &   4.98               &  4.88 &    \\ \hline
\end{tabular}}

\caption{Human evaluation of model responses on \textbf{Fl}uency, \textbf{Rel}evance, \textbf{Faith}fulness, \textbf{Obj}ectivity, and \textbf{Help}fulness scored 1 (worst) to 5 (best). }
%\vspace{-6mm}
\label{tab:response_eval}
\end{table}
\hl{\noindent \textbf{Automated Evaluation}} 
Table~\ref{tab:dialogue_agent} lists model performances. Flan-T5 achieves the highest K-Prec (68.2), consistent with prior work~\citep{daheim2023elastic}, while ChatGPT performs significantly lower (K-Prec 42.1), highlighting the importance of high-quality task-specific supervision. We show that ReMuSE maintains strong faithfulness with a K-Prec of 67.6, though its BLEU and BERT scores are lower than fine-tuned models because it is optimized for faithful and diverse generation rather than surface-level alignment. In contrast, fine-tuned models benefit from task-specific supervision, improving alignment with reference responses~\cite{hsieh-etal-2023-distilling, fu-etal-2024-tiny}. We further show in ~\S\ref{sec:data_scale} that even with only 10\% of the data, models trained on ReMuSE significantly outperform zero-shot variants, demonstrating the high quality of the generated dialogues. \looseness=-1
\begin{figure}[!t]
    \centering
    \includegraphics[width=0.3\textwidth]{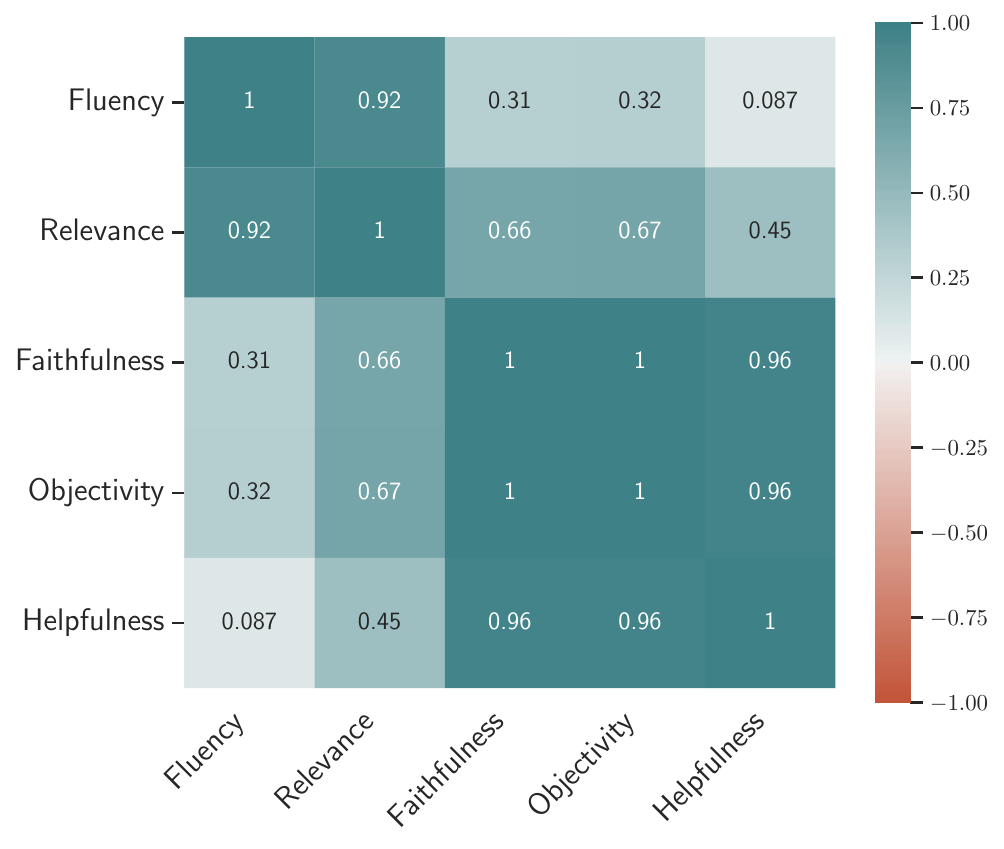}
    \caption{Correlation of the human evaluation metrics. We observe the strongest correlation between Helpfulness, Faithfulness, and Objectivity. } 
    \label{fig:human_eval}
\end{figure}
\vspace{0.2em}

\noindent \hl{\textbf{Human Evaluation}} Following \citet{santu2024prompting}, we evaluate dialogue agents on their ability to assist meta-reviewers by rating responses to key queries on contributions, strengths, weaknesses, improvements, and literature review. Two senior NLP Ph.D. students with over three years of reviewing experience rated 60 responses per model on a 1–5 Likert scale~\cite{purkayastha-etal-2023-exploring}.\footnote{See \S\ref{sec:hum_ins_resp} for annotator instructions.} Krippendorff’s $\alpha$ for fluency, relevance, faithfulness, objectivity, and helpfulness are 0.58, 0.64, 0.69, 0.66, and 0.74, respectively, substantial given task subjectivity~\cite{kennard-etal-2022-disapere}.\footnote{Annotator disagreements are analyzed in \S\ref{sec:disagree}} Table~\ref{tab:response_eval} shows that Flan-T5 scores highest, while ChatGPT scores lowest on faithfulness and objectivity. ReMuSE outperforms ChatGPT but lags behind fine-tuned models, likely due to its diversity-driven objectives and verbose responses, often ending with follow-up questions. Helpfulness correlates strongly with faithfulness and objectivity (Fig.~\ref{fig:human_eval}), supporting the quality of our synthetic training data.\footnote{Meta-reviewer question distribution and error analysis of the dialogue agent in Sec\S\ref{sec:met_rev_q} and Sec~\S\ref{sec:err_ann}} \looseness=-1

%\subsection{Time Analysis of the Dialogue Agent Usage}
\section{Deploying Meta-Reviewing Dialogue Agents: A Within-subject Study}
\label{sec:time_analysis}
We analyze the trained dialogue agent's effectiveness in real-world meta-reviewing, focusing on time and meta-review quality.

\noindent \hl{\textbf{Setup}} We conduct a within-subject controlled experiment where participants write meta-reviews with and without dialogue agent support.\footnote{The interaction interface is in Fig~\ref{fig:dial_agent_inter}, \S\ref{sec:interface}} Following prior work on user studies in peer review~\cite{santu2024prompting}, we ask fifteen Ph.D. students with peer review experience to evaluate three papers each under both conditions (45 papers total).\footnote{The participant details is in \S\ref{sec:annotators}} Following \citet{shen-etal-2022-mred}, we assess meta-review quality using (i)~\textit{Content Relevance} (alignment with review aspects, rated 1-4), (ii)~\textit{Decision Correctness} (alignment with the gold decision, binary 0/1), and (iii)~\textit{Coverage} (review diversity, rated 1-4). Two senior Ph.D. students evaluate the meta-reviews alongside an automatically generated meta-review across the chosen metrics. We employ the Level-1 prompting framework introduced in \citet{santu2024prompting}, wherein the LLM is tasked with generating a meta-review conditioned on the reviews of a specific paper. \footnote{Operation guidelines for deployment in Sec~\S\ref{sec:operational}.}%We explain the conference-specific guidelines to the participants in order to ensure consistency with the ground-truth meta-reviews.\looseness=-1
\vspace{0.2em}

\begin{table}[!t]
\centering
\resizebox{0.5\textwidth}{!}{\begin{tabular}{lccc}
\hline
\textbf{Metrics} & \textbf{w/o Dial. Agent} & \textbf{w. Dial. Agent} & \textbf{Auto Meta-Rev} \\ \hline
\multicolumn{4}{l}{\textbf{Expert}} \\ 
Content Relevance & 3.42 & \textbf{3.87} & 3.55 \\
Decision Correctness & \textbf{0.70} & 0.62 & 0.65 \\
Coverage & 3.21 & \textbf{3.76} & 3.40 \\ \hline
\multicolumn{4}{l}{\textbf{Non-expert}} \\ 
Content Relevance & 3.10 & \textbf{3.65} & 3.32 \\
Decision Correctness & 0.58 & \textbf{0.60} & 0.59 \\
Coverage & 2.95 & \textbf{3.50} & 3.20 \\ \hline
\end{tabular}}
\caption{Human evaluation of meta-reviews written by experts and non-experts, with and without dialogue agent assistance, and those generated automatically (Auto Meta-Rev). Content Relevance and Coverage are rated on a scale of 1–4. Decision Correctness is binary (0/1).}
\label{tab:dial_eval}
\end{table}
\noindent \hl{\textbf{Results.}} As shown in Table~\ref{tab:dial_eval}, reviews generated with the dialogue agent scored higher in Content Relevance (3.87 vs. 3.42) and Coverage (3.76 vs. 3.21), with Decision Correctness comparable (0.62 vs. 0.70). The automated meta-review baseline performed better than the no-agent condition (Content Relevance 3.55, Coverage 3.40, Decision Correctness 0.65) but below the dialogue-assisted setup. Interaction with the agent took an average of \textbf{20 minutes}, compared to \textbf{35 minutes} without it, demonstrating improved efficiency without compromising quality. The latency of the agent is 14.4 seconds for 12 turns ($\sim$ 1.2 seconds per turn) %As shown in Table~\ref{tab:dial_eval} (\S\ref{sec:interface}), meta-reviews generated with the dialogue agent outperform those written without it in Content Relevance (3.82 vs. 3.40) and Coverage (3.72 vs.3.20), with a slight improvement in Decision Correctness (0.70 vs.0.61). Interaction with the agent takes \textbf{20 minutes} on average, compared to \textbf{35 minutes} while writing after reading the reviews, offering a more efficient reviewing process without compromising reviewer autonomy. We further analyze the 30 papers — 15 borderline, 7 accepted, and 8 rejected. We observe that the borderline cases produce longer dialogues (12.2 turns) than accepted (8.3) or rejected (6.1) ones, indicating greater deliberation. While the discussions on accepted/rejected papers focus on content and related work, borderline cases emphasize reviewer scores and confidence. We find that the agent provides clarification when prompted but refrains from final decisions. While accepted and rejected outcomes generally align with ground truth, borderline cases show variability, likely due to external factors such as acceptance rates.
\looseness=-1

\noindent \hl{\textbf{Case Study with Non-Experts}} While meta-reviewing is usually done by experienced researchers, we test whether individuals from interdisciplinary backgrounds can use the tool effectively. Ten PhD students from non-CS fields, with no prior peer-review experience, wrote meta-reviews for three papers each under both setups.\footnote{Participant details in \S\ref{sec:annotators}} Agent-assisted reviews outperformed the no-agent condition in Content Relevance (3.65 vs. 3.10), Coverage (3.50 vs. 2.95), and Decision Correctness (0.60 vs. 0.58). The automated baseline improved over no-agent (3.32, 3.20, 0.59) but was still weaker than agent-assisted reviews. Participants spent an average of \textbf{35 minutes} with the agent versus \textbf{80 minutes} without it, showing the agent’s efficiency. Dialogues were longest for borderline papers (\textbf{12.2 turns}) with the highest conflict resolution coverage (\textbf{77\%}), demonstrating the agent’s support for structured deliberation and meta-review decision-making (Sec.~\S\ref{sec:deliberation}).
%While meta-reviewing is typically done by experienced researchers in the field, we also explore if individuals from interdisciplinary backgrounds can effectively use the tool. Ten PhD students from non-CS fields, with no prior peer-review experience, wrote meta-reviews for 3 papers each under both setups.\footnote{Participant details in \S\ref{sec:annotators}} Results in Table~\ref{tab:dial_eval} (\S\ref{sec:interface}) show that reviews with the dialogue agent scored higher in Content Relevance (3.62 vs 3.08), Coverage (3.47 vs 2.93), and Decision Correctness (0.61 vs 0.52). Participants spent an average of \textbf{35 minutes} using the agent versus \textbf{80 minutes} writing without it. This demonstrates the agent’s promise in helping newcomers learn meta-reviewing efficiently.

Paired t-tests and Wilcoxon signed-rank tests with Cohen's d show robust improvements~\cite{cleophas2016paired}. For 15 PhD students, the dialogue agent increased content relevance by 0.42 and coverage by 0.52, saving 15 minutes. For 10 non-PhD participants, relevance and coverage rose by 0.54 each, with a 45-minute reduction. Gains exceed those of automated meta-reviewing (+0.13–0.22, +0.19–0.25), and are statistically significant, likely generalizing to a larger population.\footnote{Detailed statistics and effect sizes in \S\ref{sec:appendix_stats}}

\section{Related Work} 
\hl{\noindent \textbf{Meta-Review Generation}} Meta-reviewing is commonly framed as multi-document summarization on the reviews. \citet{cikm} treat reviews, author responses, and reviewer discussions as an argumentative graph to generate comprehensive meta-reviews. \citet{li-etal-2023-summarizing} encode entire reviewer conversation threads using a multi-task approach to meta-review generation. \citet{shen-etal-2022-mred} introduce control codes for generating diverse meta-reviews. \citet{zeng2023scientific} use LLMs in a self-refinement paradigm based on task-specific checklists. Recently, \citet{santu2024prompting} analyzed the performance of LLMs for generating various meta-reviewing aspects. However, we redefine meta-reviewing as a dialogue that aids meta-reviewers in decision-making, rather than automatically generating text, preserving their scientific autonomy.

\noindent \hl{\textbf{AI-assisted decision making}} In this work, we frame meta-reviewing as a decision-making process. The role of AI in supporting human decisions has expanded across fields like finance, law, and technology \cite{10.1145/3415227, 10.1145/3593013.3594087}, following the collaborative decision-making paradigm \cite{ lin2023decisionoriented}. While \citet{lin2023decisionoriented} explore reward-based models with fixed solutions, we focus on meta-reviewing, where no single best solution exists, and decisions depend on the meta-reviewer's expertise.
%\vspace{0.2em}

\noindent \hl{\textbf{Synthetic Dialogue Generation}} We fine-tune meta-review agents on synthetically generated, review-grounded dialogues that reflect reasoning behind meta-review decisions. Prior synthetic data approaches~\cite{dai2022dialog, bao-etal-2023-synthetic} are limited by scale. We leverage LLMs as cost-effective curators to produce full-length, knowledge-grounded dialogues with high faithfulness and specificity. Using a zero-shot self-refinement strategy~\cite{madaan2023self}, we iteratively enhance multiple aspects of dialogue quality, outperforming prior refinement methods~\cite{wang2023selfconsistency} and enabling high-quality training data for complex tasks.\looseness=-1

\section{Conclusion}
In this work, we explored the potential of dialogue agents to assist meta-reviewers in the decision-making process, offering an alternative to traditional summarization approaches. We addressed the data scarcity challenge with a self-refinement method for generating diverse, faithful decision-making dialogues using LLMs. We found that models fine-tuned on our data significantly outperformed their zero-shot counterparts. Using a controlled experiment, we demonstrated that dialogue agents reduced the meta-reviewing time by almost half while maintaining reviewer autonomy. Our findings highlight the potential of AI-driven tools to improve peer-reviewing efficiency and encourage further research in this area. %We make the generated dialogue dataset, \textsc{Decidial}, publicly available.

\section*{Limitations}
%In this work, we propose a new perspective towards meta-reviewing and introduce a novel method along with a dataset to assist meta-reviewers and other decision-makers in relevant domains. However, our work comes with several limitations. 
\paragraph{Bias}
We employ both closed and open-source large language models that are standard for dialogue generation. These models have been criticized for producing harmful or biased content which may reflect in the generated synthetic data. We do not address biases such as socio-demographic factors, given the assumed neutrality of the decision-making domain. Additionally, we do not consider potential bias in the knowledge sources, such as paper reviews in ORSUM, which may include subjective content. Our focus is on generating faithful and diverse dialogues; analyzing the effect of bias in the underlying data is left to future work. We also emphasize that the agent cannot compensate for poor-quality reviews -- addressing review quality remains a critical, orthogonal research direction.

\paragraph{Practical Application}
LLMs often lack interpretability, making it difficult to trace how recommendations are generated. We do not aim to improve interpretability in this work, and the responsibility for validating suggestions lies with the meta-reviewer. Accountability is especially important when agent outputs influence decisions. While our agent is designed not to make final judgments, it may occasionally offer flawed suggestions. To mitigate this, reviewer training and clear usage guidelines are essential. 

\paragraph{Scope}
Our study focuses on English-language reviews in AI conferences. The limited availability of high-quality, annotated datasets outside well-studied domains—particularly in fields such as the humanities—currently restricts the generalizability and extension of this work to those areas. This data scarcity poses challenges for both model training and rigorous evaluation, thereby constraining cross-domain applicability. Future work could address this limitation by curating or leveraging domain-specific datasets and extending the proposed approach to multilingual settings and underexplored domains, including various subfields within the humanities, to assess its robustness and broader applicability.

\paragraph{Evaluation}
We further note that improvements in ``decision correctness'' 
should be interpreted as enhanced \emph{alignment} with historical meta-review outcomes rather than objective accuracy. 
Real conference decisions are shaped by multiple contextual factors --acceptance quotas, program‐chair policies, and field‐specific norms -- that are beyond the scope of our work.  Our evaluation isolates the deliberative quality of dialogue support rather than predicting actual acceptance decisions. Future work could incorporate contextual priors (e.g., acceptance‐rate distributions)  to model decision processes more realistically.

\section*{Ethics Statement}
In this work, we provide a detailed study of the dialogue generation capabilities of Large language models for meta-reviewing. Consequently, this work has an ethical dimension since it can be perceived as an attempt to replace human meta-reviewers with LLMs. We would like to emphasize that we solely focus on assisting the meta-reviewers in decision making using dialogue agents rather than providing them with any subjective opinions. Unlike previous work, we do not attempt in any way to generate meta-reviews and strongly believe that such a task is a \textbf{`Human-In-The-Loop'} process which requires highly skilled researchers (meta-reviewers) operating as humans to make the loop effective. 

\section*{Acknowledgements}
This work has been funded by the German Research Foundation (DFG) as part of the Research Training Group KRITIS No. GRK 2222, along with the German Federal Ministry of Education and Research and the Hessian Ministry of Higher Education, Research, Science and the Arts, within their joint support of the National Research Center for Applied Cybersecurity ATHENE. We gratefully acknowledge the support of Microsoft with a grant for access to OpenAI GPT models via the Azure cloud (Accelerate Foundation Model Academic Research). The work of Anne Lauscher is funded under the Excellence Strategy of the German Federal Government and the Federal States. The authors acknowledge the support of Schloss Dagstuhl – Leibniz Center for Informatics through the Dagstuhl Seminar  \textit{`22432: Towards a Unified Model of Scholarly Argumentation'}.

We thank Jonathan Tonglet, Federico Tiblias, Anmol Goel, Indraneil Paul, Aishik Mandal, Luke Bates, Aniket Pramanick, and Chen Cecilia Liu for their help with annotation; Hiba Arnout and Luke Bates for initial feedback on a draft of this paper; and Ming Tsun Lai, Andrew Eliseev, Jens Wala, Letizia Curreri, and Yannick Burmeister for fruitful discussions and evaluation of the dialogue agent.
\bibliography{acl_latex}

@Inbook{cleophas2016paired,
author="Cleophas, Ton J.
and Zwinderman, Aeilko H.",
title="Paired Continuous Data (Paired-t, Wilcoxon) (10 Patients)",
bookTitle="SPSS for Starters: SPSS for Starters",
year="2010",
publisher="Springer Netherlands",
address="Dordrecht",
pages="7--9",
isbn="978-90-481-9519-0",
doi="10.1007/978-90-481-9519-0_3",
url="https://doi.org/10.1007/978-90-481-9519-0_3",
note={\url{https://doi.org/10.1007/978-90-481-9519-0_3}}
}

@article{liu2019robertarobustlyoptimizedbert,
title = {Ro{BERT}a: {A} {R}obustly {O}ptimized {BERT} {P}retraining {A}pproach}, 
      author={Yinhan Liu and Myle Ott and Naman Goyal and Jingfei Du and Mandar Joshi and Danqi Chen and Omer Levy and Mike Lewis and Luke Zettlemoyer and Veselin Stoyanov},
      year={2019},
      journal={Arxiv preprint arXiv: 1907.11692},
      eprint={1907.11692},
      archivePrefix={arXiv},
      primaryClass={cs.CL},
      url={https://arxiv.org/abs/1907.11692}, 
}

@inproceedings{hu2025explaininglengthbiasllmbased,
    title = "Explaining Length Bias in {LLM}-Based Preference Evaluations",
    author = "Hu, Zhengyu  and
      Song, Linxin  and
      Zhang, Jieyu  and
      Xiao, Zheyuan  and
      Wang, Tianfu  and
      Chen, Zhengyu  and
      Yuan, Nicholas Jing  and
      Lian, Jianxun  and
      Ding, Kaize  and
      Xiong, Hui",
    editor = "Christodoulopoulos, Christos  and
      Chakraborty, Tanmoy  and
      Rose, Carolyn  and
      Peng, Violet",
    booktitle = "Findings of the Association for Computational Linguistics: EMNLP 2025",
    month = nov,
    year = "2025",
    address = "Suzhou, China",
    publisher = "Association for Computational Linguistics",
    url = "https://aclanthology.org/2025.findings-emnlp.358/",
    doi = "10.18653/v1/2025.findings-emnlp.358",
    pages = "6763--6794",
    ISBN = "979-8-89176-335-7"
}

@inproceedings{
huang2024large,
title={Large Language Models Cannot Self-Correct Reasoning Yet},
author={Jie Huang and Xinyun Chen and Swaroop Mishra and Huaixiu Steven Zheng and Adams Wei Yu and Xinying Song and Denny Zhou},
booktitle={12th International Conference on Learning Representations, {ICLR} 2024,
Vienna, Austria, May 07-11, 2024},
year={2024},
url={https://openreview.net/forum?id=IkmD3fKBPQ},
publisher="Openreview.net"
}

@inproceedings{
    wang2023selfconsistency,
    title={Self-Consistency Improves Chain of Thought Reasoning in Language Models},
    author={Xuezhi Wang and Jason Wei and Dale Schuurmans and Quoc  V Le and Ed H. Chi and Sharan Narang and Aakanksha Chowdhery and Denny Zhou},
    booktitle={Proceedings of the 11th International Conference on Learning Representations, ICLR 2023, Kigali, Rwanda, May 01-05, 2023 },
    year={2023},
    url={https://openreview.net/forum?id=1PL1NIMMrw},
    publisher="Openreview.net"
}

@article{golinkoff2019language,
    author = {Golinkoff, Roberta Michnick and Hoff, Erika and Rowe, Meredith L and Tamis-LeMonda, Catherine S and Hirsh-Pasek, Kathy},
    journal = {Child development},
    number = {3},
    pages = {985--992},
    publisher = {Wiley Online Library},
    title = {Language matters: Denying the existence of the 30-million-word gap has serious consequences},
    volume = {90},
    year = {2019},
    note={\url{https://doi.org/10.1111/cdev.13128}}
}

@article{lever2011discussing,
    author = {Lever, Rosemary and S{\'e}n{\'e}chal, Monique},
    journal = {Journal of experimental child psychology},
    number = {1},
    pages = {1--24},
    publisher = {Elsevier},
    title = {Discussing stories: On how a dialogic reading intervention improves kindergartners’ oral narrative construction},
    volume = {108},
    year = {2011},
    note={\url{https://doi.org/10.1016/j.jecp.2010.07.002}}
}

@inproceedings{gamzu-etal-2021-identifying,
    address = {Online},
    author = {Gamzu, Iftah  and
Gonen, Hila  and
Kutiel, Gilad  and
Levy, Ran  and
Agichtein, Eugene},
    booktitle = {Proceedings of the 2021 Conference of the North American Chapter of the Association for Computational Linguistics: Human Language Technologies},
    doi = {10.18653/v1/2021.naacl-main.55},
    pages = {678--691},
    publisher = {Association for Computational Linguistics},
    title = {Identifying Helpful Sentences in Product Reviews},
    url = {https://aclanthology.org/2021.naacl-main.55},
    year = {2021}
}

@inproceedings{zhang-etal-2016-conversational,
    address = {San Diego, California},
    author = {Zhang, Justine  and
Kumar, Ravi  and
Ravi, Sujith  and
Danescu-Niculescu-Mizil, Cristian},
    booktitle = {Proceedings of the 2016 Conference of the North {A}merican Chapter of the Association for Computational Linguistics: Human Language Technologies},
    doi = {10.18653/v1/N16-1017},
    pages = {136--141},
    publisher = {Association for Computational Linguistics},
    title = {Conversational Flow in {O}xford-style Debates},
    url = {https://aclanthology.org/N16-1017},
    year = {2016}
}

@article{lin2023decisionoriented,
    title = "Decision-Oriented Dialogue for Human-{AI} Collaboration",
    author = "Lin, Jessy  and
      Tomlin, Nicholas  and
      Andreas, Jacob  and
      Eisner, Jason",
    journal = "Transactions of the Association for Computational Linguistics",
    volume = "12",
    year = "2024",
    address = "Cambridge, MA",
    publisher = "MIT Press",
    url = "https://aclanthology.org/2024.tacl-1.50/",
    doi = "10.1162/tacl_a_00679",
    pages = "892--911"
}

@inproceedings{10.1145/3593013.3594087,
    address = {New York, NY, USA},
    author = {Lai, Vivian and Chen, Chacha and Smith-Renner, Alison and Liao, Q. Vera and Tan, Chenhao},
    booktitle = {Proceedings of the 2023 ACM Conference on Fairness, Accountability, and Transparency},
    doi = {10.1145/3593013.3594087},
    isbn = {9798400701924},
    location = {Chicago, IL, USA},
    numpages = {17},
    pages = {1369–1385},
    publisher = {Association for Computing Machinery},
    series = {FAccT '23},
    title = {Towards a Science of Human-AI Decision Making: An Overview of Design Space in Empirical Human-Subject Studies},
    url = {https://doi.org/10.1145/3593013.3594087},
    year = {2023}
}

@article{10.1145/3415227,
title = {Intelligent financial system: How AI is transforming finance},
journal = {Journal of Financial Stability},
volume = {81},
pages = {101472},
year = {2025},
issn = {1572-3089},
doi = {https://doi.org/10.1016/j.jfs.2025.101472},
url = {https://www.sciencedirect.com/science/article/pii/S1572308925001019},
author = {I. Aldasoro and L. Gambacorta and A. Korinek and V. Shreeti and M. Stein},
note    = {\url{https://doi.org/10.1016/j.jfs.2025.101472}}
}

@inproceedings{shen-etal-2022-mred,
    address = {Dublin, Ireland},
    author = {Shen, Chenhui  and
Cheng, Liying  and
Zhou, Ran  and
Bing, Lidong  and
You, Yang  and
Si, Luo},
    booktitle = {Findings of the Association for Computational Linguistics: ACL 2022},
    doi = {10.18653/v1/2022.findings-acl.198},
    pages = {2521--2535},
    publisher = {Association for Computational Linguistics},
    title = {{MR}e{D}: A Meta-Review Dataset for Structure-Controllable Text Generation},
    url = {https://aclanthology.org/2022.findings-acl.198},
    year = {2022}
}

@inproceedings{zeng2023scientific,
author = {Zeng, Qi and Sidhu, Mankeerat and Blume, Ansel and Chan, Hou Pong and Wang, Lu and Ji, Heng},
title = {Scientific Opinion Summarization: Paper Meta-review Generation Dataset, Methods, and Evaluation},
year = {2025},
isbn = {978-981-97-9535-2},
publisher = {Springer-Verlag},
address = {Berlin, Heidelberg},
url = {https://doi.org/10.1007/978-981-97-9536-9_2},
doi = {10.1007/978-981-97-9536-9_2},
booktitle = {Artificial Intelligence for Research and Democracy: First International Workshop, AI4Research 2024, and 4th International Workshop, DemocrAI 2024, Held in Conjunction with IJCAI 2024, Jeju, South Korea, August 5, 2024, Proceedings},
pages = {20–38},
numpages = {19},
location = {Jeju Island, Korea (Republic of)}
}

@inproceedings{stasaski-etal-2020-cima,
    title = "{CIMA}: A Large Open Access Dialogue Dataset for Tutoring",
    author = "Stasaski, Katherine  and
      Kao, Kimberly  and
      Hearst, Marti A.",
    editor = "Burstein, Jill  and
      Kochmar, Ekaterina  and
      Leacock, Claudia  and
      Madnani, Nitin  and
      Pil{\'a}n, Ildik{\'o}  and
      Yannakoudakis, Helen  and
      Zesch, Torsten",
    booktitle = "Proceedings of the Fifteenth Workshop on Innovative Use of NLP for Building Educational Applications",
    month = jul,
    year = "2020",
    address = "Seattle, WA, USA {\textrightarrow} Online",
    publisher = "Association for Computational Linguistics",
    url = "https://aclanthology.org/2020.bea-1.5/",
    doi = "10.18653/v1/2020.bea-1.5",
    pages = "52--64"
}

@article{10.1093/jamiaopen/ooae114,
    author = {Smolyak, Daniel and Bjarnadóttir, Margrét V and Crowley, Kenyon and Agarwal, Ritu},
    title = {Large language models and synthetic health data: progress and prospects},
    journal = {JAMIA Open},
    volume = {7},
    number = {4},
    pages = {ooae114},
    year = {2024},
    month = {10},
    issn = {2574-2531},
    doi = {10.1093/jamiaopen/ooae114},
    url = {https://doi.org/10.1093/jamiaopen/ooae114},
    eprint = {https://academic.oup.com/jamiaopen/article-pdf/7/4/ooae114/60129986/ooae114.pdf},
   note={\url{https://doi.org/10.1093/jamiaopen/ooae114}}
}

@inproceedings{li-etal-2023-summarizing,
    address = {Singapore},
    author = {Li, Miao  and
Hovy, Eduard  and
Lau, Jey},
    booktitle = {Findings of the Association for Computational Linguistics: EMNLP 2023},
    doi = {10.18653/v1/2023.findings-emnlp.472},
    editor = {Bouamor, Houda  and
Pino, Juan  and
Bali, Kalika},
    pages = {7089--7112},
    publisher = {Association for Computational Linguistics},
    title = {Summarizing Multiple Documents with Conversational Structure for Meta-Review Generation},
    url = {https://aclanthology.org/2023.findings-emnlp.472},
    year = {2023}
}

@article{ware2015stm,
  title={An overview of scientific and scholarly journal publishing},
  author={Ware, Mark and Mabe, Michael},
  journal={The STM report},
  url={https://digitalcommons.unl.edu/cgi/viewcontent.cgi?article=1008&context=scholcom},
  volume={1082},
  pages={1083},
  year={2009},
  note={\url{https://digitalcommons.unl.edu/cgi/viewcontent.cgi?article=1008&context=scholcom}}
}

@inproceedings{cikm,
    address = {New York, NY, USA},
    author = {Wu, Po-Cheng and Yen, An-Zi and Huang, Hen-Hsen and Chen, Hsin-Hsi},
    booktitle = {Proceedings of the 31st ACM International Conference on Information \& Knowledge Management},
    doi = {10.1145/3511808.3557360},
    isbn = {9781450392365},
    location = {Atlanta, GA, USA},
    numpages = {10},
    pages = {2189–2198},
    publisher = {Association for Computing Machinery},
    series = {CIKM '22},
    title = {Incorporating Peer Reviews and Rebuttal Counter-Arguments for Meta-Review Generation},
    url = {https://doi.org/10.1145/3511808.3557360},
    year = {2022}
}

@inproceedings{chen-etal-2023-places,
    address = {Dubrovnik, Croatia},
    author = {Chen, Maximillian  and
Papangelis, Alexandros  and
Tao, Chenyang  and
Kim, Seokhwan  and
Rosenbaum, Andy  and
Liu, Yang  and
Yu, Zhou  and
Hakkani-Tur, Dilek},
    booktitle = {Findings of the Association for Computational Linguistics: EACL 2023},
    pages = {844--868},
    publisher = {Association for Computational Linguistics},
    title = {{PLACES}: Prompting Language Models for Social Conversation Synthesis},
    url = {https://aclanthology.org/2023.findings-eacl.63},
    year = {2023}
}

@article{møller2023prompt,
    author = {Anders Giovanni Møller and Jacob Aarup Dalsgaard and Arianna Pera and Luca Maria Aiello},
    title = {Is a {P}rompt and a Few Samples All You Need? Using {GPT-4} for Data Augmentation in Low-Resource Classification Tasks},
    url = {https://arxiv.org/abs/2304.13861},
    journal={Arxiv preprint arXiv: 2304.13861},
    year = {2023}
}

@article{faithdial,
    author = {Dziri, Nouha and Kamalloo, Ehsan and Milton, Sivan and Zaiane, Osmar and Yu, Mo and Ponti, Edoardo M. and Reddy, Siva},
    doi = {10.1162/tacl_a_00529},
    eprint = {https://direct.mit.edu/tacl/article-pdf/doi/10.1162/tacl\_a\_00529/2065956/tacl\_a\_00529.pdf},
    issn = {2307-387X},
    journal = {Transactions of the Association for Computational Linguistics},
    pages = {1473-1490},
    title = {{FaithDial: A Faithful Benchmark for Information-Seeking Dialogue}},
    url = {https://doi.org/10.1162/tacl\_a\_00529},
    volume = {10},
    year = {2022}
}

@article{adlakha2023evaluating,
    title = "Evaluating Correctness and Faithfulness of Instruction-Following Models for Question Answering",
    author = "Adlakha, Vaibhav  and
      BehnamGhader, Parishad  and
      Lu, Xing Han  and
      Meade, Nicholas  and
      Reddy, Siva",
    journal = "Transactions of the Association for Computational Linguistics",
    volume = "12",
    year = "2024",
    address = "Cambridge, MA",
    publisher = "MIT Press",
    url = "https://aclanthology.org/2024.tacl-1.38/",
    doi = "10.1162/tacl_a_00667",
    pages = "681--699"
}

@inproceedings{honovich-etal-2021-q2,
    address = {Online and Punta Cana, Dominican Republic},
    author = {Honovich, Or  and
Choshen, Leshem  and
Aharoni, Roee  and
Neeman, Ella  and
Szpektor, Idan  and
Abend, Omri},
    booktitle = {Proceedings of the 2021 Conference on Empirical Methods in Natural Language Processing},
    doi = {10.18653/v1/2021.emnlp-main.619},
    pages = {7856--7870},
    publisher = {Association for Computational Linguistics},
    title = {$Q^{2}$: {E}valuating Factual Consistency in Knowledge-Grounded Dialogues via Question Generation and Question Answering},
    url = {https://aclanthology.org/2021.emnlp-main.619},
    year = {2021}
}

@inproceedings{li2015fast,
    author = {Junyi Jessy Li and
Ani Nenkova},
    bibsource = {dblp computer science bibliography, https://dblp.org},
    booktitle = {Proceedings of the Twenty-Ninth {AAAI} Conference on Artificial Intelligence,
January 25-30, 2015, Austin, Texas, {USA}},
    editor = {Blai Bonet and
Sven Koenig},
    pages = {2281--2287},
    publisher = {{AAAI} Press},
    title = {Fast and Accurate Prediction of Sentence Specificity},
    url = {http://www.aaai.org/ocs/index.php/AAAI/AAAI15/paper/view/9941},
    year = {2015}
}

@article{balke5,
          volume = {12},
            issn = {2192-5283},
       publisher = {Schloss Dagstuhl - Leibniz-Zentrum f{\"u}r Informatik},
        language = {en},
           title = {5.4 Scholarly Argumentation as a Community Dialogue},
           month = {May},
           pages = {202},
          number = {10},
            year = {2023},
          author = {Balke, Wolf-Tilo and Vlachos, Andreas and Ceolin, Davide and Alshomary, Milad and Dycke, Nils and Purkayastha, Sukannya and Gurevych, Iryna and Lauscher, Anne and Beck, Tilman},
         journal = {Dagstuhl Reports},
             doi = {https://doi.org/10.4230/DagRep.12.10.175},
             url = {http://tubiblio.ulb.tu-darmstadt.de/137720/},
note={\url{{https://doi.org/10.4230/DagRep.12.10.175}}}
}

@inproceedings{
ouyang2022training,
title={Training language models to follow instructions with human feedback},
author={Long Ouyang and Jeffrey Wu and Xu Jiang and Diogo Almeida and Carroll Wainwright and Pamela Mishkin and Chong Zhang and Sandhini Agarwal and Katarina Slama and Alex Gray and John Schulman and Jacob Hilton and Fraser Kelton and Luke Miller and Maddie Simens and Amanda Askell and Peter Welinder and Paul Christiano and Jan Leike and Ryan Lowe},
booktitle={Proceedings of the Thirty-sixth
Conference on Neural Information Processing Sys-
tems, NeurIPS 2022, New Orleans, USA (Hybrid), Nov 28- Dec 09, 2022},
editor={Alice H. Oh and Alekh Agarwal and Danielle Belgrave and Kyunghyun Cho},
year={2022},
url={https://openreview.net/forum?id=TG8KACxEON},
publisher="Openreview.net"
}

@article{bai2022training,
    author = {Yuntao Bai and Andy Jones and Kamal Ndousse and Amanda Askell and Anna Chen and Nova DasSarma and Dawn Drain and Stanislav Fort and Deep Ganguli and Tom Henighan and Nicholas Joseph and Saurav Kadavath and Jackson Kernion and Tom Conerly and Sheer El-Showk and Nelson Elhage and Zac Hatfield-Dodds and Danny Hernandez and Tristan Hume and Scott Johnston and Shauna Kravec and Liane Lovitt and Neel Nanda and Catherine Olsson and Dario Amodei and Tom Brown and Jack Clark and Sam McCandlish and Chris Olah and Ben Mann and Jared Kaplan},
    eprint = {2204.05862},
    journal = {ArXiv preprint arXiv: 2204.05862},
    primaryclass = {cs.CL},
    title = {Training a Helpful and Harmless Assistant with Reinforcement Learning from Human Feedback},
    url = {https://arxiv.org/abs/2204.05862},
    year = {2022}
}

@article{touvron2023llama,
    author = {Hugo Touvron and Louis Martin and Kevin Stone and Peter Albert and Amjad Almahairi and Yasmine Babaei and Nikolay Bashlykov and Soumya Batra and Prajjwal Bhargava and Shruti Bhosale and Dan Bikel and Lukas Blecher others},
    journal = {ArXiv preprint arXiv: 2307.09288 },
    title = {Llama 2: Open Foundation and Fine-Tuned Chat Models},
    url = {https://arxiv.org/abs/2307.09288},
    year = {2023}
}

@article{jiang2023mistral,
    author = {Albert Q. Jiang and Alexandre Sablayrolles and Arthur Mensch and Chris Bamford and Devendra Singh Chaplot and Diego de las Casas and Florian Bressand and Gianna Lengyel and others},
    journal = {ArXiv preprint arXiv: 2310.06825},
    title = {Mistral 7B},
    url = {https://arxiv.org/abs/2310.06825},
    eprint={arxiv:2310.06825},
    year = {2023}
}

@article{jiang2024mixtral,
    author = {Albert Q. Jiang and Alexandre Sablayrolles and Antoine Roux and Arthur Mensch and Blanche Savary and Chris Bamford and Devendra Singh Chaplot and Diego de las Casas and others},
    journal = {ArXiv preprint arXiv: 2401.04088},
    title = {Mixtral of {E}xperts},
    url = {https://arxiv.org/abs/2401.04088},
    year = {2024}
}

@inproceedings{
madaan2023self,
title = {Self-Refine: {I}terative {R}efinement with {S}elf-{F}eedback},
author={Aman Madaan and Niket Tandon and Prakhar Gupta and Skyler Hallinan and Luyu Gao and Sarah Wiegreffe and Uri Alon and Nouha Dziri and Shrimai Prabhumoye and Yiming Yang and Shashank Gupta and Bodhisattwa Prasad Majumder and Katherine Hermann and Sean Welleck and Amir Yazdanbakhsh and Peter Clark},
booktitle={Proceedings of the Thirty-seventh Conference on Neural Information Processing Systems, NeurIPS 2023, New Orleans, USA},
year={2023},
url={https://openreview.net/forum?id=S37hOerQLB},
publisher="Openreview.net"
}

@inproceedings{chen2023iterative,
    title = "Iterative {T}ranslation {R}efinement with {L}arge {L}anguage {M}odels",
    author = "Chen, Pinzhen  and
      Guo, Zhicheng  and
      Haddow, Barry  and
      Heafield, Kenneth",
    editor = "Scarton, Carolina  and
      Prescott, Charlotte  and
      Bayliss, Chris  and
      Oakley, Chris  and
      Wright, Joanna  and
      Wrigley, Stuart  and
      Song, Xingyi  and
      Gow-Smith, Edward  and
      Bawden, Rachel  and
      S{\'a}nchez-Cartagena, V{\'i}ctor M  and
      Cadwell, Patrick  and
      Lapshinova-Koltunski, Ekaterina  and
      Cabarr{\~a}o, Vera  and
      Chatzitheodorou, Konstantinos  and
      Nurminen, Mary  and
      Kanojia, Diptesh  and
      Moniz, Helena",
    booktitle = "Proceedings of the 25th Annual Conference of the European Association for Machine Translation (Volume 1)",
    month = jun,
    year = "2024",
    address = "Sheffield, UK",
    publisher = "European Association for Machine Translation (EAMT)",
    url = "https://aclanthology.org/2024.eamt-1.17/",
    pages = "181--190"
}

@inproceedings{dycke-etal-2023-nlpeer,
    address = {Toronto, Canada},
    author = {Dycke, Nils  and
Kuznetsov, Ilia  and
Gurevych, Iryna},
    booktitle = {Proceedings of the 61st Annual Meeting of the Association for Computational Linguistics (Volume 1: Long Papers)},
    doi = {10.18653/v1/2023.acl-long.277},
    editor = {Rogers, Anna  and
Boyd-Graber, Jordan  and
Okazaki, Naoaki},
    pages = {5049--5073},
    publisher = {Association for Computational Linguistics},
    title = {{NLP}eer: A Unified Resource for the Computational Study of Peer Review},
    url = {https://aclanthology.org/2023.acl-long.277},
    year = {2023}
}

@article{CHEN2022100001,
    author = {Xieling Chen and Haoran Xie and Xiaohui Tao},
    doi = {https://doi.org/10.1016/j.nlp.2022.100001},
    issn = {2949-7191},
    journal = {Natural Language Processing Journal},
    pages = {100001},
    title = {Vision, status, and research topics of Natural Language Processing},
    url = {https://www.sciencedirect.com/science/article/pii/S2949719122000012},
    volume = {1},
    year = {2022},
    note={\url{https://doi.org/10.1016/j.nlp.2022.100001}}
}

@inproceedings{wu-etal-2022-dg2,
    address = {Edinburgh, UK},
    author = {Wu, Qingyang  and
Feng, Song  and
Chen, Derek  and
Joshi, Sachindra  and
Lastras, Luis  and
Yu, Zhou},
    booktitle = {Proceedings of the 23rd Annual Meeting of the Special Interest Group on Discourse and Dialogue},
    pages = {204--216},
    publisher = {Association for Computational Linguistics},
    title = {{DG}2: Data Augmentation Through Document Grounded Dialogue Generation},
    url = {https://aclanthology.org/2022.sigdial-1.21},
    year = {2022}
}

@inproceedings{bao-etal-2023-synthetic,
    address = {Toronto, Canada},
    author = {Bao, Jianzhu  and
Wang, Rui  and
Wang, Yasheng  and
Sun, Aixin  and
Li, Yitong  and
Mi, Fei  and
Xu, Ruifeng},
    booktitle = {Proceedings of the 61st Annual Meeting of the Association for Computational Linguistics (Volume 1: Long Papers)},
    doi = {10.18653/v1/2023.acl-long.608},
    editor = {Rogers, Anna  and
Boyd-Graber, Jordan  and
Okazaki, Naoaki},
    pages = {10866--10882},
    publisher = {Association for Computational Linguistics},
    title = {A Synthetic Data Generation Framework for Grounded Dialogues},
    url = {https://aclanthology.org/2023.acl-long.608},
    year = {2023}
}

@inproceedings{dai2022dialog,
    author = {Zhuyun Dai and
Arun Tejasvi Chaganty and
Vincent Y. Zhao and
Aida Amini and
Qazi Mamunur Rashid and
Mike Green and
Kelvin Guu},
    bibsource = {dblp computer science bibliography, https://dblp.org},
    booktitle = {International Conference on Machine Learning, {ICML} 2022, 17-23 July
2022, Baltimore, Maryland, {USA}},
    editor = {Kamalika Chaudhuri and
Stefanie Jegelka and
Le Song and
Csaba Szepesv{\'{a}}ri and
Gang Niu and
Sivan Sabato},
    pages = {4558--4586},
    publisher = {{PMLR}},
    series = {Proceedings of Machine Learning Research},
    title = {Dialog Inpainting: Turning Documents into Dialogs},
    url = {https://proceedings.mlr.press/v162/dai22a.html},
    volume = {162},
    year = {2022}
}

@inproceedings{farag-etal-2022-opening,
    address = {Abu Dhabi, United Arab Emirates},
    author = {Farag, Youmna  and
Brand, Charlotte  and
Amidei, Jacopo  and
Piwek, Paul  and
Stafford, Tom  and
Stoyanchev, Svetlana  and
Vlachos, Andreas},
    booktitle = {Findings of the Association for Computational Linguistics: EMNLP 2022},
    pages = {4569--4582},
    publisher = {Association for Computational Linguistics},
    title = {Opening up Minds with Argumentative Dialogues},
    url = {https://aclanthology.org/2022.findings-emnlp.335},
    year = {2022}
}

@inproceedings{wang-etal-2019-persuasion,
    address = {Florence, Italy},
    author = {Wang, Xuewei  and
Shi, Weiyan  and
Kim, Richard  and
Oh, Yoojung  and
Yang, Sijia  and
Zhang, Jingwen  and
Yu, Zhou},
    booktitle = {Proceedings of the 57th Annual Meeting of the Association for Computational Linguistics},
    doi = {10.18653/v1/P19-1566},
    pages = {5635--5649},
    publisher = {Association for Computational Linguistics},
    title = {Persuasion for Good: Towards a Personalized Persuasive Dialogue System for Social Good},
    url = {https://aclanthology.org/P19-1566},
    year = {2019}
}

@inproceedings{daheim2023elastic,
    title = "Elastic Weight Removal for Faithful and Abstractive Dialogue Generation",
    author = "Daheim, Nico  and
      Dziri, Nouha  and
      Sachan, Mrinmaya  and
      Gurevych, Iryna  and
      Ponti, Edoardo",
    editor = "Duh, Kevin  and
      Gomez, Helena  and
      Bethard, Steven",
    booktitle = "Proceedings of the 2024 Conference of the North American Chapter of the Association for Computational Linguistics: Human Language Technologies (Volume 1: Long Papers)",
    month = jun,
    year = "2024",
    address = "Mexico City, Mexico",
    publisher = "Association for Computational Linguistics",
    url = "https://aclanthology.org/2024.naacl-long.393/",
    doi = "10.18653/v1/2024.naacl-long.393",
    pages = "7096--7112"
}

@inproceedings{post2018clarity,
    address = {Brussels, Belgium},
    author = {Post, Matt},
    booktitle = {Proceedings of the Third Conference on Machine Translation: Research Papers},
    doi = {10.18653/v1/W18-6319},
    pages = {186--191},
    publisher = {Association for Computational Linguistics},
    title = {A Call for Clarity in Reporting {BLEU} Scores},
    url = {https://aclanthology.org/W18-6319},
    year = {2018}
}

@inproceedings{pmlr-v202-longpre23a,
    author = {Longpre, Shayne and Hou, Le and Vu, Tu and Webson, Albert and Chung, Hyung Won and Tay, Yi and Zhou, Denny and Le, Quoc V and Zoph, Barret and Wei, Jason and Roberts, Adam},
    booktitle = {Proceedings of the 40th International Conference on Machine Learning, ICML 2023, Honolulu Hawaii USA, July 23 - 29, 2023},
    editor = {Krause, Andreas and Brunskill, Emma and Cho, Kyunghyun and Engelhardt, Barbara and Sabato, Sivan and Scarlett, Jonathan},
    publisher = {PMLR},
    title = {The {F}lan {C}ollection: {D}esigning {D}ata and {M}ethods for {E}ffective {I}nstruction {T}uning},
    url = {https://proceedings.mlr.press/v202/longpre23a.html},
    year = {2023}
}

@article{JMLR:v21:20-074,
    author = {Colin Raffel and
Noam Shazeer and
Adam Roberts and
Katherine Lee and
Sharan Narang and
Michael Matena and
Yanqi Zhou and
Wei Li and
Peter J. Liu},
    bibsource = {dblp computer science bibliography, https://dblp.org},
    journal = {Journal of Machine Learning Research},
    pages = {140:1--140:67},
    title = {Exploring the Limits of Transfer Learning with a Unified Text-to-Text
Transformer},
    url = {http://jmlr.org/papers/v21/20-074.html},
    volume = {21},
    year = {2020},
    note={\url{https://dl.acm.org/doi/10.5555/3455716.3455856}}
}

@inproceedings{Zhang*2020BERTScore:,
    author = {Tianyi Zhang and
Varsha Kishore and
Felix Wu and
Kilian Q. Weinberger and
Yoav Artzi},
    bibsource = {dblp computer science bibliography, https://dblp.org},
    booktitle = {Proceedings of the 8th International Conference on Learning Representations, {ICLR} 2020,
Addis Ababa, Ethiopia, April 26-30, 2020},
    publisher = {OpenReview.net},
    title = {BERTScore: Evaluating Text Generation with {BERT}},
    url = {https://openreview.net/forum?id=SkeHuCVFDr},
    year = {2020}
}

@inproceedings{rashkin-etal-2021-increasing,
    address = {Online},
    author = {Rashkin, Hannah  and
Reitter, David  and
Tomar, Gaurav Singh  and
Das, Dipanjan},
    booktitle = {Proceedings of the 59th Annual Meeting of the Association for Computational Linguistics and the 11th International Joint Conference on Natural Language Processing (Volume 1: Long Papers)},
    doi = {10.18653/v1/2021.acl-long.58},
    pages = {704--718},
    publisher = {Association for Computational Linguistics},
    title = {Increasing Faithfulness in Knowledge-Grounded Dialogue with Controllable Features},
    url = {https://aclanthology.org/2021.acl-long.58},
    year = {2021}
}

@inproceedings{macina-etal-2023-mathdial,
    address = {Singapore},
    author = {Macina, Jakub  and
Daheim, Nico  and
Chowdhury, Sankalan  and
Sinha, Tanmay  and
Kapur, Manu  and
Gurevych, Iryna  and
Sachan, Mrinmaya},
    booktitle = {Findings of the Association for Computational Linguistics: EMNLP 2023},
    doi = {10.18653/v1/2023.findings-emnlp.372},
    editor = {Bouamor, Houda  and
Pino, Juan  and
Bali, Kalika},
    pages = {5602--5621},
    publisher = {Association for Computational Linguistics},
    title = {{M}ath{D}ial: A Dialogue Tutoring Dataset with Rich Pedagogical Properties Grounded in Math Reasoning Problems},
    url = {https://aclanthology.org/2023.findings-emnlp.372},
    year = {2023}
}

@inproceedings{holtzman2020curious,
    author = {Ari Holtzman and
Jan Buys and
Li Du and
Maxwell Forbes and
Yejin Choi},
    bibsource = {dblp computer science bibliography, https://dblp.org},
    booktitle = {Proceedings of the 8th International Conference on Learning Representations, {ICLR} 2020,
Addis Ababa, Ethiopia, April 26-30, 2020},
    publisher = {OpenReview.net},
    title = {The Curious Case of Neural Text Degeneration},
    url = {https://openreview.net/forum?id=rygGQyrFvH},
    year = {2020}
}

@inproceedings{wang2023enable,
  title={Enabling Lanuguage Models to Implicitly Learn Self-Improvement},
  author={Wang, Ziqi and Hou, Le and Lu, Tianjian and Wu, Yuexin and Li, Yunxuan and Yu, Hongkun and Ji, Heng},
  booktitle={12th International Conference on Learning Representations, {ICLR} 2024,
Vienna, Austria, May 07-11, 2024},
year={2024},
publisher="OpenReview.Net",
url={https://openreview.net/forum?id=2tVHNRZuCs},
}

@inproceedings{wolf-etal-2020-transformers,
    address = {Online},
    author = {Wolf, Thomas  and
Debut, Lysandre  and
Sanh, Victor  and
Chaumond, Julien  and
Delangue, Clement  and
Moi, Anthony  and
Cistac, Pierric  and
Rault, Tim  and
Louf, Remi  and
Funtowicz, Morgan  and
Davison, Joe  and
Shleifer, Sam  and
von Platen, Patrick  and
Ma, Clara  and
Jernite, Yacine  and
Plu, Julien  and
Xu, Canwen  and
Le Scao, Teven  and
Gugger, Sylvain  and
Drame, Mariama  and
Lhoest, Quentin  and
Rush, Alexander},
    booktitle = {Proceedings of the 2020 Conference on Empirical Methods in Natural Language Processing: System Demonstrations},
    doi = {10.18653/v1/2020.emnlp-demos.6},
    pages = {38--45},
    publisher = {Association for Computational Linguistics},
    title = {Transformers: State-of-the-Art Natural Language Processing},
    url = {https://aclanthology.org/2020.emnlp-demos.6},
    year = {2020}
}

@inproceedings{bian2024chatgpt,
    title = "{C}hat{GPT} Is a Knowledgeable but Inexperienced Solver: An Investigation of Commonsense Problem in Large Language Models",
    author = "Bian, Ning  and
      Han, Xianpei  and
      Sun, Le  and
      Lin, Hongyu  and
      Lu, Yaojie  and
      He, Ben  and
      Jiang, Shanshan  and
      Dong, Bin",
    editor = "Calzolari, Nicoletta  and
      Kan, Min-Yen  and
      Hoste, Veronique  and
      Lenci, Alessandro  and
      Sakti, Sakriani  and
      Xue, Nianwen",
    booktitle = "Proceedings of the 2024 Joint International Conference on Computational Linguistics, Language Resources and Evaluation (LREC-COLING 2024)",
    month = may,
    year = "2024",
    address = "Torino, Italia",
    publisher = "ELRA and ICCL",
    url = "https://aclanthology.org/2024.lrec-main.276/",
    pages = "3098--3110"
}

@article{gemmateam2025gemma3technicalreport,
      title={Gemma 3 Technical Report}, 
      author={Aishwarya Kamath and Johan Ferret and Shreya Pathak and Nino Vieillard and Ramona Merhej and Sarah Perrin and Tatiana Matejovicova and others},
      year={2025},
      eprint={2503.19786},
      archivePrefix={arXiv},
      journal={Arxiv preprint arXiv: 2503.19786},
      primaryClass={cs.CL},
      url={https://arxiv.org/abs/2503.19786}, 
}

@article{yang2025qwen3technicalreport,
      title={Qwen3 Technical Report}, 
      author={An Yang and Anfeng Li and Baosong Yang and Beichen Zhang and Binyuan Hui and Bo Zheng and Bowen Yu and Chang Gao and Chengen Huang and Chenxu Lv and Chujie Zheng and others},
      year={2025},
      eprint={2505.09388},
      archivePrefix={arXiv},
      primaryClass={cs.CL},
      url={https://arxiv.org/abs/2505.09388}, 
      journal={Arxiv preprint arXiv: 2505.09388}
}

@inproceedings{priya2025argueagreepersonalitydrivenargumentationbased,
    title = "We Argue to Agree: Towards Personality-Driven Argumentation-Based Negotiation Dialogue Systems for Tourism",
    author = "Priya, Priyanshu  and
      Dudhate, Saurav  and
      Yasheshbhai, Desai Vishesh  and
      Ekbal, Asif",
    editor = "Christodoulopoulos, Christos  and
      Chakraborty, Tanmoy  and
      Rose, Carolyn  and
      Peng, Violet",
    booktitle = "Findings of the Association for Computational Linguistics: EMNLP 2025",
    month = nov,
    year = "2025",
    address = "Suzhou, China",
    publisher = "Association for Computational Linguistics",
    url = "https://aclanthology.org/2025.findings-emnlp.1390/",
    doi = "10.18653/v1/2025.findings-emnlp.1390",
    pages = "25504--25536",
    ISBN = "979-8-89176-335-7"
}

@inproceedings{chen2024reconcile,
    title = "{R}e{C}oncile: {R}ound-{T}able {C}onference {I}mproves {R}easoning via {C}onsensus among {D}iverse {LLM}s",
    author = "Chen, Justin  and
      Saha, Swarnadeep  and
      Bansal, Mohit",
    editor = "Ku, Lun-Wei  and
      Martins, Andre  and
      Srikumar, Vivek",
    booktitle = "Proceedings of the 62nd Annual Meeting of the Association for Computational Linguistics (Volume 1: Long Papers)",
    month = aug,
    year = "2024",
    address = "Bangkok, Thailand",
    publisher = "Association for Computational Linguistics",
    url = "https://aclanthology.org/2024.acl-long.381/",
    doi = "10.18653/v1/2024.acl-long.381",
    pages = "7066--7085"
}

@article{zhang2022opt,
title = {{OPT}: {O}pen {P}re-trained {T}ransformer {L}anguage {M}odels}, 
      author={Susan Zhang and Stephen Roller and Naman Goyal and Mikel Artetxe and Moya Chen and others},
      year={2022},
      eprint={2205.01068},
      archivePrefix={arXiv},
      primaryClass={cs.CL},
journal = {ArXiv preprint arXiv: 2205.01068},
    url= {https://arxiv.org/abs/2205.01068}
}

@article{kuznetsov2024natural,
      title={What Can Natural Language Processing Do for Peer Review?}, 
      author={Ilia Kuznetsov and Osama Mohammed Afzal and Koen Dercksen and Nils Dycke and Alexander Goldberg and Tom Hope and Dirk Hovy and others},
      year={2024},
      eprint={2405.06563},
      archivePrefix={arXiv},
      primaryClass={cs.CL},
journal = {ArXiv preprint arXiv: 2405.06563 },
    url= {https://arxiv.org/abs/2405.06563},
}

@inproceedings{
Lan2020ALBERT:,
title = {{ALBERT}: {A} Lite {BERT} for {S}elf-supervised {L}earning of {L}anguage {R}epresentations},
author={Zhenzhong Lan and Mingda Chen and Sebastian Goodman and Kevin Gimpel and Piyush Sharma and Radu Soricut},
year={2020},
url={https://openreview.net/forum?id=H1eA7AEtvS},
publisher = {OpenReview.net},
booktitle = {Proceedings of the 8th International Conference on Learning Representations, {ICLR} 2020,
Addis Ababa, Ethiopia, April 26-30, 2020}
}

@inproceedings{
he2021deberta,
title={{DEBERTA}: {D}ecoding-{E}nhanced {BERT} with {D}isentangled {A}ttention},
author={Pengcheng He and Xiaodong Liu and Jianfeng Gao and Weizhu Chen},
booktitle={9th International Conference on
Learning Representations, ICLR 2021, Virtual, May 03-07, 2021},
year={2021},
url={https://openreview.net/forum?id=XPZIaotutsD},
publisher="Openreview.net"
}

@inproceedings{santu2024prompting,
    title = "{LLM}s as Meta-Reviewers' Assistants: A Case Study",
    author = "Hossain, Eftekhar  and
      Sinha, Sanjeev Kumar  and
      Bansal, Naman  and
      Knipper, R. Alexander  and
      Sarkar, Souvika  and
      Salvador, John  and
      Mahajan, Yash  and
      Guttikonda, Sri Ram Pavan Kumar  and
      Akter, Mousumi  and
      Hassan, Md. Mahadi  and
      Freestone, Matthew  and
      Jr., Matthew C. Williams  and
      Feng, Dongji  and
      Karmaker, Santu",
    editor = "Chiruzzo, Luis  and
      Ritter, Alan  and
      Wang, Lu",
    booktitle = "Proceedings of the 2025 Conference of the Nations of the Americas Chapter of the Association for Computational Linguistics: Human Language Technologies (Volume 1: Long Papers)",
    month = apr,
    year = "2025",
    address = "Albuquerque, New Mexico",
    publisher = "Association for Computational Linguistics",
    url = "https://aclanthology.org/2025.naacl-long.395/",
    doi = "10.18653/v1/2025.naacl-long.395",
    pages = "7763--7803",
    ISBN = "979-8-89176-189-6"
}

@Article{fi15100336,
AUTHOR = {Lozić, Edisa and Štular, Benjamin},
title = {Fluent but Not Factual: {A} Comparative {A}nalysis of {C}hatGPT and Other {AI} Chatbots’ {P}roficiency and {O}riginality in {S}cientific {W}riting for {H}umanities},
JOURNAL = {Future Internet},
VOLUME = {15},
YEAR = {2023},
NUMBER = {10},
ARTICLE-NUMBER = {336},
URL = {https://www.mdpi.com/1999-5903/15/10/336},
ISSN = {1999-5903},
DOI = {10.3390/fi15100336},
note={\url{https://doi.org/10.3390/fi15100336}}
}

@inproceedings{li-etal-2023-autoconv,
    title = "{A}uto{C}onv: Automatically Generating Information-seeking Conversations with Large Language Models",
    author = "Li, Siheng  and
      Yang, Cheng  and
      Yin, Yichun  and
      Zhu, Xinyu  and
      Cheng, Zesen  and
      Shang, Lifeng  and
      Jiang, Xin  and
      Liu, Qun  and
      Yang, Yujiu",
    editor = "Rogers, Anna  and
      Boyd-Graber, Jordan  and
      Okazaki, Naoaki",
    booktitle = "Proceedings of the 61st Annual Meeting of the Association for Computational Linguistics (Volume 2: Short Papers)",
    month = jul,
    year = "2023",
    address = "Toronto, Canada",
    publisher = "Association for Computational Linguistics",
    url = "https://aclanthology.org/2023.acl-short.149",
    doi = "10.18653/v1/2023.acl-short.149",
    pages = "1751--1762",
}

@article{schintler2023critical,
  title = {A {C}ritical {E}xamination of the Ethics of {AI}-{M}ediated {P}eer {R}eview},
  author={Schintler, Laurie A and McNeely, Connie L and Witte, James},
  journal={Arxiv preprint arXiv: 2309.12356 },
url={https://arxiv.org/abs/2309.12356},
  year={2023}
}

@article{donker2023dangers,
  title={The dangers of using large language models for peer review},
  author={Donker, Tjibbe},
  journal={The Lancet Infectious Diseases},
  volume={23},
  number={7},
  pages={781},
  year={2023},
  note={\url{https://doi.org/10.1016/S1473-3099(23)00290-6}}
  
}

@article{goddard2014automation,
title = {Automation bias: Empirical results assessing influencing factors},
journal = {International Journal of Medical Informatics},
volume = {83},
number = {5},
pages = {368-375},
year = {2014},
issn = {1386-5056},
doi = {https://doi.org/10.1016/j.ijmedinf.2014.01.001},
url = {https://www.sciencedirect.com/science/article/pii/S1386505614000148},
author = {Kate Goddard and Abdul Roudsari and Jeremy C. Wyatt},
note={\url{https://doi.org/10.1016/j.ijmedinf.2014.01.001}}
}

@inproceedings{schemmer2022influence,
  title={On the Influence of Explainable AI on Automation Bias},
  author={Schemmer, Max and K{\"u}hl, Niklas and Benz, Carina and Satzger, Gerhard},
  booktitle={Proceedings of the 30th European Conference on Information Systems (ECIS), Timișoara, RO, June 18-24, 2022},
  year={2022},
  url ={https://aisel.aisnet.org/ecis2022_rip/51/},
  publisher= "aisel.aisnet.org"
}

@article{ke2024mitigating,
  title={Mitigating cognitive biases in clinical decision-making through multi-agent conversations using large language models: simulation study},
  author={Ke, Yuhe and Yang, Rui and Lie, Sui An and Lim, Taylor Xin Yi and Ning, Yilin and Li, Irene and Abdullah, Hairil Rizal and Ting, Daniel Shu Wei and Liu, Nan},
  journal={Journal of Medical Internet Research},
  volume={26},
  pages={e59439},
  year={2024},
  publisher={JMIR Publications Toronto, Canada},
  note={\url{https://doi.org/10.2196/59439}}
}

@inproceedings{hsieh-etal-2023-distilling,
    title = "Distilling Step-by-Step! Outperforming Larger Language Models with Less Training Data and Smaller Model Sizes",
    author = "Hsieh, Cheng-Yu  and
      Li, Chun-Liang  and
      Yeh, Chih-kuan  and
      Nakhost, Hootan  and
      Fujii, Yasuhisa  and
      Ratner, Alex  and
      Krishna, Ranjay  and
      Lee, Chen-Yu  and
      Pfister, Tomas",
    editor = "Rogers, Anna  and
      Boyd-Graber, Jordan  and
      Okazaki, Naoaki",
    booktitle = "Findings of the Association for Computational Linguistics: ACL 2023",
    month = jul,
    year = "2023",
    address = "Toronto, Canada",
    publisher = "Association for Computational Linguistics",
    url = "https://aclanthology.org/2023.findings-acl.507/",
    doi = "10.18653/v1/2023.findings-acl.507",
    pages = "8003--8017"
}

@inproceedings{fu-etal-2024-tiny,
    title = "Tiny Titans: Can Smaller Large Language Models Punch Above Their Weight in the Real World for Meeting Summarization?",
    author = "Fu, Xue-Yong  and
      Laskar, Md Tahmid Rahman  and
      Khasanova, Elena  and
      Chen, Cheng  and
      Tn, Shashi",
    editor = "Yang, Yi  and
      Davani, Aida  and
      Sil, Avi  and
      Kumar, Anoop",
    booktitle = "Proceedings of the 2024 Conference of the North American Chapter of the Association for Computational Linguistics: Human Language Technologies (Volume 6: Industry Track)",
    month = jun,
    year = "2024",
    address = "Mexico City, Mexico",
    publisher = "Association for Computational Linguistics",
    url = "https://aclanthology.org/2024.naacl-industry.33/",
    doi = "10.18653/v1/2024.naacl-industry.33",
    pages = "387--394"
}

@inproceedings{purkayastha-etal-2023-exploring,
    title = "Exploring Jiu-Jitsu Argumentation for Writing Peer Review Rebuttals",
    author = "Purkayastha, Sukannya  and
      Lauscher, Anne  and
      Gurevych, Iryna",
    editor = "Bouamor, Houda  and
      Pino, Juan  and
      Bali, Kalika",
    booktitle = "Proceedings of the 2023 Conference on Empirical Methods in Natural Language Processing",
    month = dec,
    year = "2023",
    address = "Singapore",
    publisher = "Association for Computational Linguistics",
    url = "https://aclanthology.org/2023.emnlp-main.894/",
    doi = "10.18653/v1/2023.emnlp-main.894",
    pages = "14479--14495"
}

@misc{ChatGPT,
  author       = {OpenAI},
  title        = {ChatGPT},
  howpublished = {\url{https://chat.openai.com/}},
  year         = {2026},
  note         = {Accessed: 2026‑01‑09}
}

@misc{ACL23ReviewPolicies,
  title        = {{ACL}’23 {P}eer {R}eview {P}olicies},
  author       = {Jordan Boyd‑Graber and Naoaki Okazaki and Anna Rogers},
  howpublished = {\url{https://2023.aclweb.org/blog/review-acl23/}},
  year         = {2023},
  note         = {Accessed: 2025-12-02}
}

@inproceedings{kennard-etal-2022-disapere,
    title = "{DISAPERE}: A Dataset for Discourse Structure in Peer Review Discussions",
    author = "Kennard, Neha Nayak  and
      O{'}Gorman, Tim  and
      Das, Rajarshi  and
      Sharma, Akshay  and
      Bagchi, Chhandak  and
      Clinton, Matthew  and
      Yelugam, Pranay Kumar  and
      Zamani, Hamed  and
      McCallum, Andrew",
    editor = "Carpuat, Marine  and
      de Marneffe, Marie-Catherine  and
      Meza Ruiz, Ivan Vladimir",
    booktitle = "Proceedings of the 2022 Conference of the North American Chapter of the Association for Computational Linguistics: Human Language Technologies",
    month = jul,
    year = "2022",
    address = "Seattle, United States",
    publisher = "Association for Computational Linguistics",
    url = "https://aclanthology.org/2022.naacl-main.89/",
    doi = "10.18653/v1/2022.naacl-main.89",
    pages = "1234--1249"
}

\appendix

\section{Appendix} \label{sec:appendix}

\subsection{Model Details and Computational Budget} \label{sec:model_link}
\noindent \paragraph{GPT 4o} We use the chat-tuned variant (dubbed \textit{`ChatGPT'}) of the Instruct-GPT class of models from OpenAI \cite{ouyang2022training}.\footnote{\url{https://chat.openai.com/}} This model was trained with internet data and finetuned using reinforcement learning from human feedback \cite{bai2022training}. The training and dataset details are not publicly available. We use `gpt4o'.

\noindent \paragraph{LLaMa-2} It is the upgraded version of the LLaMa family of models \cite{touvron2023llama} featuring a 40\% increase in pre-training models' data and doubled context length compared to its predecessor. We employ the 13B-chat version in our experiments.%\footnote{\url{https://huggingface.co/meta-llama/Llama-2-13b-chat-hf}}

\noindent \paragraph{Mistral} This model is trained on the grouped query and sliding window attention that helps focus on words even outside the context window~\cite{jiang2023mistral}. We use the 7B version.%\footnote{\url{https://huggingface.co/mistralai/Mistral-7B-v0.1}}

\noindent \paragraph{Mixtral} It is an open-source LLM based on a mixture of experts \cite{jiang2024mixtral}. Each feed-forward block has access to 8 parameter groups and utilizes two of them to process a token.%\footnote{\url{https://huggingface.co/mistralai/Mixtral-8x7B-v0.1}}
\noindent \paragraph{Gemma} Gemma is an open-source LLM designed for instruction-following and reasoning tasks \cite{gemmateam2025gemma3technicalreport}. We use the 12B-IT version. 

\noindent \paragraph{Qwen-3} Qwen-3 is a multilingual LLM capable of handling text and code across multiple languages \cite{yang2025qwen3technicalreport}. We use the 8B-IT version.

We use the Azure OpenAI service to prompt ChatGPT. The rest of the models are implemented using huggingface transformers~\cite{wolf-etal-2020-transformers}. Following previous work on data generation with LLMs \cite{adlakha2023evaluating}, we use a high temperature of $0.95$ and employ top-p sampling~\cite{holtzman2020curious} of $0.95$ to avoid sampling repetitions. 

\subsubsection{Computational Budget}
We ran all the experiments on Nvidia A100
80GB GPUs. None of the experiments consumed more than 36 hours.
\subsection{Dagstuhl dataset} \label{sec:dagastuhl}

 %\todo{put just one pointer to section with all prompts}. 
\vspace{0.2em}

\begin{figure}[!ht]
    \centering
    \includegraphics[width=0.8\linewidth]{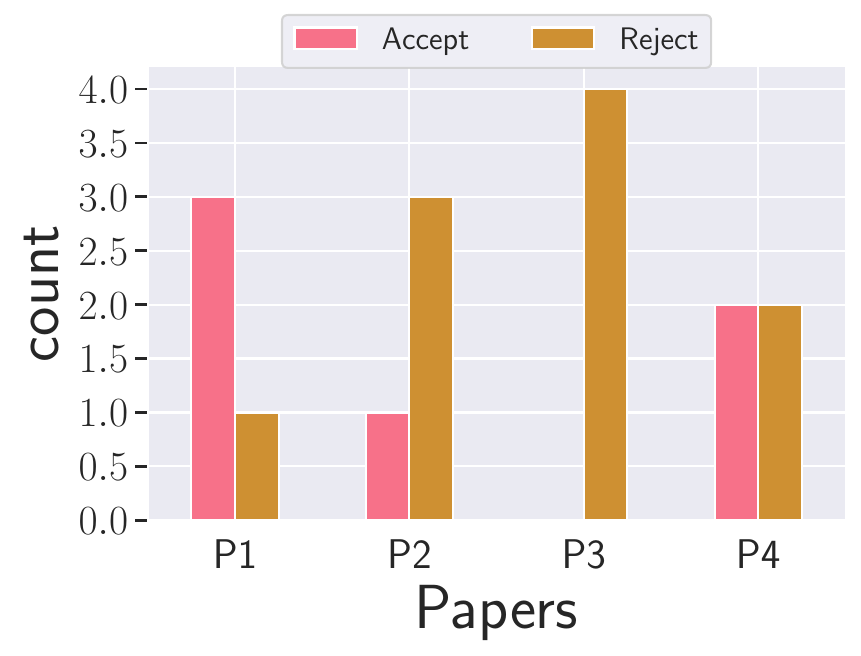}
    \caption{Diversity of Paper Acceptance Decisions in the Dagstuhl dataset}
    \label{fig:dagstuhl_dec}
\end{figure}

For the comparison of our synthetically generated dialogues with real human meta-reviewing decision making dialogues we employ the \textit{Dagstuhl} dataset \cite{balke5} kindly provided by the authors.
For this dataset, $4$ senior and $4$ junior scientists acted as meta-reviewers and assistants, respectively, in a wizard-of-oz scenario. The study was conducted on $4$ papers selected from the \textsc{NLPeer} corpus \cite{dycke-etal-2023-nlpeer}. The assistants were provided with the paper and reviews in advance, whereas the meta-reviewers were instructed to skim the reviews within 10 minutes. The meta-reviewers were asked to converse with the dialogue agents to gather information and eventually decide on the acceptance of the paper. The paper acceptance decisions for the dataset are displayed in Figure~\ref{fig:dagstuhl_dec}. Across the 4 papers, only 1 paper had a unanimous reject decision, while all the other papers received diverse verdicts.
\begin{figure}[!h]
    \centering
    \includegraphics[width=0.45\textwidth]{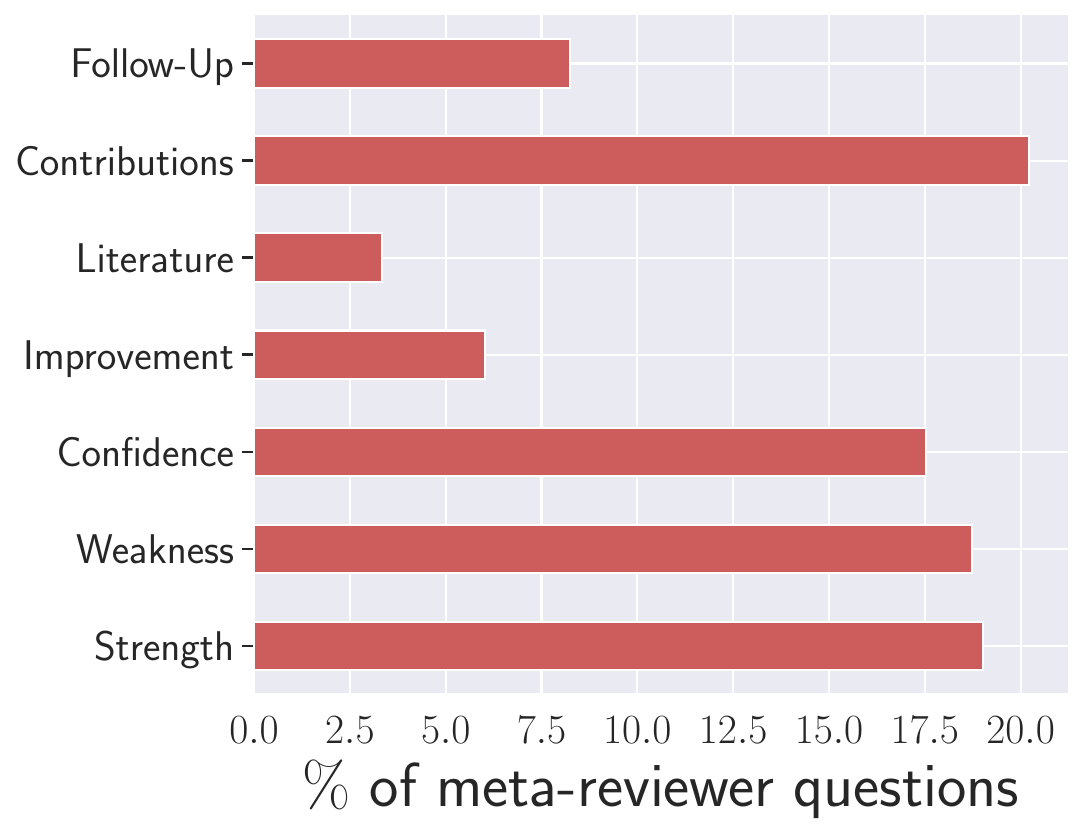}
    \caption{Distribution of meta-reviewer questions in our dataset. } 
    \label{fig:type_dist}
\end{figure}
\subsubsection{Diversity of decisions in Dagstuhl dataset} \label{sec:dagstuhl_dec}
The Dagstuhl Dataset consists of 4 papers meta-reviewed by 4 senior academicians. The paper acceptance decisions is displayed in Figure~\ref{fig:dagstuhl_dec}. Across the 4 papers, only 1 paper had a unanimous reject decision, while all the other papers received diverse verdicts.

\subsection{Prompts for all the experiments}\label{sec:prompts}
Our proposed method, ReMuSE, uses different prompting templates for the Initial Dialogue Generation, Feedback, and Refinement steps. The templates consist of an instruction specific to the particular ReMuSE step combined with our knowledge source, i.e., the paper's reviews, title, and type (long or short). We use multiple prompts for the initial dialogue generation to account for model sensitivity. We name these prompts as follows: extensive prompt (simulates the study in \citet{balke5}), paraphrased prompt (paraphrase of the extensive prompt), and tl:dr prompt (3-sentence summary of the extensive prompt). We provide full templates for these prompts along with the ones used for Feedback and Refinement in  Appendix~\S\ref{sec:init_dialogue}, ~\S\ref{sec:feedback} and \S\ref{sec:refinement}, respectively.
\subsubsection{Initial Dialogue Generation Prompts for all the experiments}\label{sec:init_dialogue}
We show the different prompts for initial dialogue generation below:
\clearpage

\noindent \paragraph{Extensive Prompt}
\begin{mybox}{Extensive Prompt (Meta-Reviewing)}
\small
\textbf{Instruction:} Generate a multi-turn dialogue between a meta-reviewer and a dialogue agent for reviews
about a paper. Every turn should alternate between ``Dialogue Agent" and ``Meta Reviewer"
utterances. The meta-reviewer must explicitly tell whether the paper is accepted or rejected
at the nal turn. The dialogue ends with the meta-reviewer saying whether they accept or
reject the paper. You need to simulate both the meta-reviewer and the dialogue agent. The
dialogue agent has access to only the reviews and type of the paper. A meta-reviewer mainly
asks questions and the dialogue agent should just answer from the reviews. A dialogue agent
should never recommend anything/ give any opinions/ decide anything for the paper. A
dialogue agent has no conscience of its own. A meta-reviewer can also ask follow-up
questions and grill the dialogue agent for more information on the reviews. As a meta-
reviewer, you should also weigh the importance of the condence of the reviewers
while making a decision. 

\vspace{3pt}
\textbf{Title}: [title]

\vspace{3pt}
\textbf{Knowledge Source:} \textit{ \textbf{Review 1}: The paper proposes a benchmark to measure the quality of the systems in the growing domain of code intelligence ..., \textbf{Review 2}: .., \textbf{Review 3}: ... }

\vspace{3pt}

%\textbf{Sentence:} $<$\textit{question}$>$

%\vspace{3pt}

%\textbf{Paraphrases:}

\end{mybox}

\begin{mybox}{Extensive Prompt (Product-Buying)}
\small
\textbf{Instruction:} Generate a multi-turn dialogue between a buyer who wants to buy a product and a
dialogue agent for reviews about that product. Every turn should start with either
"Dialogue Agent" followed by their utterance or "Buyer" followed by their utterance.
The dialogue agent should always remain neutral and take a neutral stand in any case.
The buyer should reach a decision to buy/not buy the paper at the final turn. You need to simulate both
the buyer and the dialogue agent. The buyer does not have access to the reviews but the
dialogue agent has access to the reviews and nothing else. It should answer any question about the
reviews. A buyer mainly asks questions and the dialogue agent should refrain from
offering advice if its opinion is sought. The dialogue agent should not recommend/advise anything
regarding the product.

\vspace{3pt}
\textbf{Title}: ``OP/TECH USA 1504372 Pro Loop Strap (Royal)".

\vspace{3pt}
\textbf{Knowledge Source:} \textit{ This strap is just what I needed
for lugging my D700 + battery pack + It is comfort-
able and can handle a dslr and heavy lens. I bought
this for the Nikon D700 camera and replaced the fac-
tory Nikon strap which is thinner. Very nice product
with great adaptability and not made in China!... }

\vspace{3pt}

%\textbf{Sentence:} $<$\textit{question}$>$

%\vspace{3pt}

%\textbf{Paraphrases:}
\end{mybox}

\begin{mybox}{Extensive Prompt (Debates)}
\small
\textbf{Instruction:} Generate a multi-turn dialogue between a debate decision maker who needs to take a
decision about which side wins a debate and a dialogue agent that has access to the
arguments put forward by both the sides. Every turn should start with either "Dialogue Agent"
followed by their utterance or "Decision Maker" followed by their utterance. You need to
simulate both the decision maker and the dialogue agent. The decision maker does not have
access to the arguments put forward by both the sides. The decision maker should just rely
on the dialogue agent to know about the arguments from both the sides. A decision maker
mainly asks questions and the dialogue agent answers. The dialogue agent should refrain
from answering if its opinion/decision is sought. The dialogue agent has access to only the
arguments of the debate. The dialogue agent should never decide on who wins the debate
and should always take a neutral stand when any opinion is sought.

\vspace{3pt}
\textbf{Title}: ``Abolish The Minimum Wage".

\vspace{3pt}
\textbf{Knowledge Source:} \textit{ \textbf{For arguments}: 'Let's talk about what would happen to real people in real families ...', 
\textbf{Against Arguments}: `We need to abolish the minimum wage. And there's only one argument that matters on this issue, and it's the moral argument. Does the minimum wage make the world a better...' }

\vspace{3pt}

%\textbf{Sentence:} $<$\textit{question}$>$

%\vspace{3pt}

%\textbf{Paraphrases:}
\end{mybox}

\noindent \paragraph{Paraphrased Prompt}
\begin{mybox}{Paraphrased Prompt (Meta-Reviewing)}
\small
\textbf{Instruction:} Create a multi-turn dialogue featuring a meta-reviewer and a dialogue agent discussing reviews
of a paper. Each turn should begin with either "Dialogue Agent" or "Meta Reviewer" followed by their respective statements. The meta-reviewer must make a final decision to "accept" or "reject" the paper and explicitly state their choice at the end. The dialogue concludes with the meta-reviewer indicating whether they accept or reject the paper. The meta-reviewer, aware
only of the paper's title, lacks access to reviews or information about the paper type and relies on the dialogue agent for this data. The dialogue agent has access solely to reviews and the
paper type, responding neutrally and abstaining from offering opinions, recommendations, or decisions. If asked for its opinion, the dialogue agent must respond with "I do not know." The
meta-reviewer primarily poses questions, and the dialogue agent answers based on the reviews, with the option to inquire about reviewer condence. The meta-reviewer may ask follow-up questions and press the dialogue agent for additional details from the reviews. It's important to note that the dialogue agent has no inherent conscience or personal opinions. 

\vspace{3pt}
\textbf{Title}: [title]

\vspace{3pt}
\textbf{Knowledge Source:} \textit{ \textbf{Review 1}: The paper proposes a benchmark to measure the quality of the systems in the growing domain of code intelligence ..., \textbf{Review 2}: .., \textbf{Review 3}: ... }

\end{mybox}

\begin{mybox}{Paraphrased Prompt (Product-Buying)}
\small
\textbf{Instruction:} Create a multi-turn conversation between a prospective buyer interested in purchasing a
specic product and a dialogue agent knowledgeable about reviews for that product. Each
turn should commence with either "Dialogue Agent" followed by their statement or "Buyer"
followed by theirs. The dialogue agent must maintain a neutral stance throughout and
respond solely based on the reviews available. The buyer, who has not read the reviews,
will make a denitive decision on whether to purchase the product or not in the nal turn.
The dialogue agent should refrain from expressing opinions and respond with 'I do not know'
if asked for personal input. It's essential to clarify that the dialogue agent is not a seller of the
product and possesses information solely from the reviews. 

\vspace{3pt}
\textbf{Title}: ``OP/TECH USA 1504372 Pro Loop Strap (Royal)".

\vspace{3pt}
\textbf{Knowledge Source:} \textit{ This strap is just what I needed
for lugging my D700 + battery pack + It is comfort-
able and can handle a dslr and heavy lens. I bought
this for the Nikon D700 camera and replaced the fac-
tory Nikon strap which is thinner. Very nice product
with great adaptability and not made in China!... }

%\textbf{Sentence:} $<$\textit{question}$>$

%\vspace{3pt}

%\textbf{Paraphrases:}
\end{mybox}

\begin{mybox}{Paraphrased Prompt (Debates)}
\small
\textbf{Instruction:} Craft a multi-turn conversation involving a decision maker tasked with determining the winning
side in a debate, and a dialogue agent equipped with information on the arguments presented
by both sides. Each turn should initiate with either "Dialogue Agent" or "Decision Maker" followed
by their respective statements. The decision maker, who is unaware of the arguments, must
make a conclusive decision on the debate's victor in the nal turn. The dialogue agent should
remain impartial, responding only based on the arguments, and stating "I do not know" if asked
for its opinion. The decision maker relies solely on the dialogue agent for information about the
arguments from both sides. The dialogue agent possesses access solely to the arguments
and is prohibited from making decisions or expressing opinions on the debate outcome.
\vspace{3pt}
\textbf{Title}: ``Abolish The Minimum Wage".

\vspace{3pt}
\textbf{Knowledge Source:} \textit{ \textbf{For arguments}: 'Let's talk about what would happen to real people in real families if the U.S. were to go down this untested and quite radical path of eliminating the minimum wage as our opponents urge...', 
\textbf{Against Arguments}: `We need to abolish the minimum wage. And there's only one argument that matters on this issue, and it's the moral argument. Does the minimum wage make the world a better place?...' }

\end{mybox}
\noindent \paragraph{TL:DR Prompt}
\begin{mybox}{TL:DR Prompt for Initial Dialogue Generation (Meta-Reviewing)}
\small
\textbf{Instruction:} Generate a multi-turn dialogue between a meta-reviewer and a dialogue agent
regarding a paper's reviews, with the meta-reviewer aiming to decide whether to accept
or reject the paper. The dialogue agent, relying on reviews and aware of the paper type,
must respond to the meta-reviewer's questions and refrain from providing opinions or
recommendations, while the meta-reviewer weighs the importance of reviewer confidence
in making the final decision

\vspace{3pt}
\textbf{Title}: [title].

\vspace{3pt}
\textbf{Knowledge Source:} \textit{ \textbf{Review 1}: The paper proposes a benchmark to measure the quality of the systems in the growing domain of code intelligence ..., \textbf{Review 2}: .., \textbf{Review 3}: ... }

\end{mybox}

\begin{mybox}{TL:DR Prompt for Initial Dialogue Generation (Product-Buying)}
\small
\textbf{Instruction:} Generate a multi-turn dialogue between a buyer and a neutral dialogue agent
regarding a product, with the buyer making a nal decision to buy or not. The dialogue
agent, equipped with access to reviews, answers the buyer's questions, refraining from
recommendations or opinions, while the buyer, without access to reviews, seeks information
to reach a purchasing decision.  

\vspace{3pt}
\textbf{Title}: ``OP/TECH USA 1504372 Pro Loop Strap (Royal)".

\vspace{3pt}
\textbf{Knowledge Source:} \textit{ This strap is just what I needed
for lugging my D700 + battery pack + It is comfort-
able and can handle a dslr and heavy lens. I bought
this for the Nikon D700 camera and replaced the fac-
tory Nikon strap which is thinner. Very nice product
with great adaptability and not made in China!... }

%\textbf{Sentence:} $<$\textit{question}$>$

%\vspace{3pt}

%\textbf{Paraphrases:}
\end{mybox}

\begin{mybox}{TL:DR Prompt for Initial Dialogue Generation (Debates)}
\small
\textbf{Instruction:} Simulate a multi-turn dialogue between a decision maker and a dialogue agent for a debate,
where the decision maker, aware only of the debate's topic, seeks information from the dialogue
agent, who has access to arguments from both sides. The decision maker aims to reach a
conclusion about which side wins, while the dialogue agent remains neutral, providing answers
solely based on the arguments presented, refraining from expressing opinions or making decisions
\vspace{3pt}

\textbf{Title}: ``Abolish The Minimum Wage".

\vspace{3pt}
\textbf{Knowledge Source:} \textit{ \textbf{For arguments}: 'Let's talk about what would happen to real people in real families if the U.S. were to go down this untested and quite radical path of eliminating the minimum wage as our opponents urge...', 
\textbf{Against Arguments}: `We need to abolish the minimum wage. And there's only one argument that matters on this issue, and it's the moral argument. Does the minimum wage make the world a better place?...' }

\end{mybox}

%\subsection{Additional System Prompts used for the LLaMa model}
%\noindent \textbf{Prompt for Feedback}: The system prompt to induce feedback agent behaviour. `You are an expert dialogue assistant that can provide feedback to improve the dialogues" so that the model behaves as a feedback-generating dialogue assistant."\\

 %\noindent \textbf{Prompt for Refinement}: We use this additional system prompt to induce the refinement behaviour ``You are a dialogue agent that can rewrite dialogues based on provided feedback."

%\subsection{Examples of dialogues generated by ReMuSE for all domains} \label{example_all}
%We provide example dialogues generated by ReMuSE in Fig~\ref{fig:dialog_all} for all the decision-making domains. We will release the generated dialogues for all the domains under Apache license.

\subsubsection{Prompt Templates for ReMuSE feedback} \label{sec:feedback}
We have different versions of the same prompt used for multiple metrics. The prompt for refinement using all the metrics is provided below: \\

\begin{mybox}{Prompt for ReMuSE feedback (Meta-Reviewing)}
\small
\textbf{Instruction:} Given the knowledge source and the dialogue, please provide actionable feedback to improve the dialogues. The feedback should just be for the overall dialogue and should start with `Feedback:'. A Q2 F1 score, Q2 NLI score, KPrecision, and specificity scores follow each utterance. The Q2 and Kprecision scores measure the groundedness of the response to the provided knowledge source. The Specificity scores measure the technicality of the response.  The feedback should try to improve all scores for the dialogue agent and the Specificity scores for the meta-reviewer.

\vspace{3pt}

\textbf{Knowledge Source:} \textit{ \textbf{Review 1}: The paper proposes a benchmark to measure the quality of the systems in the growing domain of code intelligence ..., \textbf{Review 2}: .., \textbf{Review 3}: ... }

\vspace{3pt}

\textbf{Dialogue:} \textit{Meta-Reviewer: Hello Dialogue Agent. Can you tell me more about this paper?, F1: 0.0, NLI: 0.0, Kprec: 0.0, Specificity: 0.1 \textbackslash n Dialogue Agent: Ofcourse! This is a paper about CodexGLUE, a machine learning benchmark dataset for code understanding and generation, F1: 0.12, NLI: 0.34, Kprec: 0.45, Specificity: 0.7 \textbackslash n MetaReviewer: ... }

%\vspace{3pt}

%\textbf{Sentence:} $<$\textit{question}$>$

%\vspace{3pt}

%\textbf{Paraphrases:}
\end{mybox}

%\noindent \textbf{Product-Buying}: \texttt{Given the knowledge source and the dialogue, please provide actionable feedback to improve the dialogues. The feedback should just be for the overall dialogue and should start with `Feedback:'. A Q2 F1 score, Q2 NLI score, KPrecision, and specificity scores follow each utterance. The Q2 and Kprecision scores measure the groundedness of the response to the provided knowledge source. The Specificity scores measure the technicality of the response.  The feedback should try to improve all scores for the dialogue agent and the Specificity scores for the buyer. The Knowledge Source is \{Review: the Product OP/TECH USA 1504372 Pro Loop Strap (Royal) is very comfortable ...\} and dialogues are \{Buyer: Hi there, I am interested in buying the OP/TECH USA 1504372 Pro Loop Strap (Royal). can you tell me more about it? F1: 0.0, NLI: 0.0 \textbackslash n Dialogue Agent: Sure! What would you like to know? F1: 0.0, NLI: 0.0, Kprec: 0.0, Spec: 0.1  \textbackslash n Buyer: How comfortable are these for long hours while travelling ? F1: 0.1, NLI: 0.1, Kprec: 0.05, Specificity: 0.4 \textbackslash n Dialogue Agent: According to the reviews, its comfortable to wear for long hours, F1: 0.2, NLI: 0.2, Kprec: 0.25, Specificity: 0.5 ... \}} \\
\begin{mybox}{Prompt for ReMuSE feedback (Debates)}
\small
\textbf{Instruction:} Given the knowledge source and the dialogue, please provide actionable feedback to improve the dialogues. The feedback should just be for the overall dialogue and should start with `Feedback:'. A Q2 F1 score, Q2 NLI score, KPrecision, and specificity scores follow each utterance. The Q2 and Kprecision scores measure the groundedness of the response to the provided knowledge source. The Specificity scores measure the technicality of the response.  The feedback should try to improve all scores for the dialogue agent and the Specificity scores for the decision-maker.

\vspace{3pt}

\textbf{Title}: ``Abolish The Minimum Wage".
\vspace{3pt}

\textbf{Knowledge Source:} \textit{\textbf{For arguments}: 'Let's talk about what would happen to real people in real families if the U.S. were to go down this untested and quite radical path of eliminating the minimum wage as our opponents urge...', 
\textbf{Against Arguments}: `We need to abolish the minimum wage. And there's only one argument that matters on this issue, and it's the moral argument. Does the minimum wage make the world a better place?...' }

\vspace{3pt}

\textbf{Dialogue:} \textit{Decision Maker: Hello, can you tell the arguments for and against debating for minimum wages? F1: 0.2, NLI: 0.2, KPrec:0.01, Specificity: 0.2 \textbackslash n Dialogue Agent: The debate over abolishing minimum wage centers on two perspectives... F1: 0.4, NLI: 0.39, KPrec: 0.45, Specificity: 0.6 ... }

%\textbf{Sentence:} $<$\textit{question}$>$

%\vspace{3pt}

%\textbf{Paraphrases:}
\end{mybox}
\begin{mybox}{Prompt for ReMuSE feedback (Product-Buying)}
\small
\textbf{Instruction:} Given the knowledge source and the dialogue, please provide actionable feedback to improve the dialogues. The feedback should just be for the overall dialogue and should start with `Feedback:'. A Q2 F1 score, Q2 NLI score, KPrecision, and specificity scores follow each utterance. The Q2 and Kprecision scores measure the groundedness of the response to the provided knowledge source. The Specificity scores measure the technicality of the response.  The feedback should try to improve all scores for the dialogue agent and the Specificity scores for the buyer.

\vspace{3pt}

\textbf{Knowledge Source:} \textit{This strap is just what I needed for lugging my D700 + battery pack + It is comfortable and can handle a dslr and heavy lens. I bought this for the Nikon D700 camera and replaced the factory Nikon strap which is thinner... }

\vspace{3pt}

\textbf{Title}: ``OP/TECH USA 1504372 Pro Loop Strap (Royal)".
\vspace{3pt}

\textbf{Dialogue:} \textit{Buyer: Hi there, I am interested in buying the OP/TECH USA 1504372 Pro Loop Strap (Royal). can you tell me more about it? F1: 0.0, NLI: 0.0 \textbackslash n Dialogue Agent: Sure! What would you like to know? F1: 0.0, NLI: 0.0, Kprec: 0.0, Spec: 0.1  \textbackslash n Buyer: How comfortable are these for long travel hours? F1: 0.1, NLI: 0.1, Kprec: 0.05, Specificity: 0.4 \textbackslash n Dialogue Agent: According to the reviews, its comfortable to wear for long hours, F1: 0.2, NLI: 0.2, Kprec: 0.25, Specificity: 0.5 \textbackslash n Buyer: ...}

\end{mybox}

\subsubsection{Prompt Templates for ReMuSE refinement } \label{sec:refinement}
We provide the refinement prompts here: \\

\begin{mybox}{Prompt for Refinement (Meta-Reviewing)}
\small
\textbf{Instruction:} Given the feedback, knowledge source, and dialogue improve the dialogue based on the feedback provided. The output should just be the new dialogue.

\vspace{3pt}

\textbf{Knowledge Source:} \textit{ \textbf{Review 1}: The paper proposes a benchmark to measure the quality of the systems in the growing domain of code intelligence ..., \textbf{Review 2}: .., \textbf{Review 3}: ... }

\vspace{3pt}
\textbf{Feedback}:  \textit{It would be helpful to provide a clearer explanation of the tasks' relevance to software development in response to reviewer one's concerns. Additionally, it may be useful to mention that the paper discusses the effectiveness of the baselines and provides an ablation study. Finally, it would be good to acknowledge reviewer three's comment explicitly and mention any plans to revise the descriptions of the datasets.}

\vspace{3pt}

\textbf{Dialogue:} \textit{Meta-Reviewer: Hello Dialogue Agent. Can you tell me more about this paper? \textbackslash n Dialogue Agent: Ofcourse! This is a paper about CodexGLUE, a machine learning benchmark dataset for code understanding and generation, \textbackslash n MetaReviewer: Interesting. Can you give me some more information about the reviews that have been
given for this paper? \textbackslash n Dialogue Agent: ... }

\vspace{3pt}

%\textbf{Sentence:} $<$\textit{question}$>$

%\vspace{3pt}

%\textbf{Paraphrases:}
\end{mybox}

\begin{mybox}{Prompt for Refinement (Product-Buying)}
\small
\textbf{Instruction:} Given the feedback, knowledge source, and dialogue, improve the dialogue based on the feedback provided. The output should just be the new dialogue.

\vspace{3pt}

\textbf{Knowledge Source:} \textit{This strap is just what I needed for lugging my D700 + battery pack + It is comfortable and can handle a dslr and heavy lens. I bought this for the Nikon D700 camera and replaced the factory Nikon strap which is thinner. Very nice product with great adaptability and not made in China!... }

\vspace{3pt}
\textbf{Feedback:} \textit{Here are some specific feedback points: 1. The buyer is hesitant about the price, and the dialogue agent could provide more information about the product's value and benefits to address these concerns. 2. The buyer asks about the reviews and ratings, and the dialogue agent could provide more specific examples or testimonials to showcase the product's positive feedback. 3. The dialogue agent could provide more information about the return policy and the retailer's customer service to alleviate any concerns the buyer may have. }

\vspace{3pt}
\textbf{Dialogue:} \textit{Buyer: Hi there, I am interested in buying the OP/TECH USA 1504372 Pro Loop Strap (Royal). can you tell me more about it? \textbackslash n Dialogue Agent: Sure! What would you like to know?  \textbackslash n Buyer: How comfortable are these for long travel hours?  \textbackslash n Dialogue Agent: According to the reviews, its comfortable to wear for long hours, \textbackslash n Buyer: ...}

\vspace{3pt}

%\textbf{Sentence:} $<$\textit{question}$>$

%\vspace{3pt}

%\textbf{Paraphrases:}
\end{mybox}
\begin{table*}
\centering

% Please add the following required packages to your document preamble:
% \usepackage{multirow}
 \resizebox{0.9\textwidth}{!}{\begin{tabular}{c|cccccc}
\hline
\textbf{Domain} & \textbf{Model} & \textbf{\begin{tabular}[c]{@{}c@{}}Rewards\end{tabular}} & \textbf{K-Prec} & \textbf{Q2-F1}            & \textbf{Q2-NLI}           & \textbf{Spec.} \\ \hline
\multirow{32}{*}{Meta-Reviewing} & ChatGPT        & -                                                                     & 0.387 $\pm$ 0.012          & \multicolumn{1}{l}{0.147 $\pm$ 0.011 } & \multicolumn{1}{l}{0.152 $\pm$ 0.011 } &  0.451 $\pm$ 0.127          \\
                            &     LLaMa          & -                                                                     &  0.341 $\pm$ 0.014           & \multicolumn{1}{l}{ 0.170 $\pm$ 0.130} & \multicolumn{1}{l}{ 0.170 $\pm$ 0.014} & 0.461 $\pm$ 0.081        \\
                             &      Mistral        & -                                                                     &  0.377 $\pm$ 0.012           & \multicolumn{1}{l}{ 0.151 $\pm$ 0.022} & \multicolumn{1}{l}{ 0.161 $\pm$ 0.014} &  0.407 $\pm$ 0.015        \\
                          &         Mixtral        & -                                                                     &  0.378 $\pm$ 0.014          & \multicolumn{1}{l}{ 0.171 $\pm$ 0.014} & \multicolumn{1}{l}{0.188 $\pm$ 0.021} &  0.422 $\pm$ 0.021          \\  \cline{2-7}
                          
                             &      ChatGPT        & K-Prec                                                                &      0.679 $\pm$ 0.022         &    0.177 $\pm$ 0.121                       &     0.197 $\pm$ 0.025                      & 0.552 $\pm$ 0.034               \\
                           &        LLaMa          & K-Prec                                                                &   0.633 $\pm$ 0.016            &  0.162 $\pm$ 0.141                         & 0.163 $\pm$ 0.012                          & 0.563 $\pm$ 0.021               \\
                           &        Mistral        & K-Prec                                                                &   0.681 $\pm$ 0.014              &   0.242 $\pm$ 0.014                        &    0.231 $\pm$ 0.021                       &      0.611 $\pm$ 0.024          \\
                          &         Mixtral        & K-Prec                                                                &  0.702 $\pm$ 0.011               &   0.268 $\pm$ 0.021                        &   0.263 $\pm$ 0.015                        &       0.646 $\pm$ 0.026         \\  \cline{2-7}
                            &       ChatGPT        & Q2 (F1 + NLI)                                                        &   0.689 $\pm$ 0.013              &   0.171 $\pm$ 0.011                        &      0.173 $\pm$ 0.013                     & 0.648 $\pm$ 0.018               \\
                               &    LLaMa          & Q2 (F1 + NLI)                                                        &  0.646 $\pm$ 0.017               &   0.178 $\pm$ 0.029                        &  0.189 $\pm$ 0.013                         &  0.571 $\pm$ 0.026             \\
                              &     Mistral        & Q2 (F1 + NLI)                                                        &  0.706 $\pm$ 0.017               &    0.268 $\pm$ 0.034                       &   0.272 $\pm$ 0.012                        & 0.626 $\pm$ 0.013               \\
                            &       Mixtral        & Q2 (F1 + NLI)                                                       & 0.721 $\pm$ 0.014                &     0.271 $\pm$ 0.016                      &   0.265 $\pm$ 0.012                       &   0.701 $\pm$ 0.028             \\ \cline{2-7}
                              &     ChatGPT        & Spec.                                                                 &  0.692 $\pm$ 0.015               &  0.180 $\pm$ 0.023                         &    0.181 $\pm$ 0.019                       &    0.652 $\pm$ 0.032            \\
                               &    LLaMa          & Spec.                                                                 & 0.681 $\pm$ 0.010                &     0.183 $\pm$ 0.021                      &      0.191 $\pm$ 0.016                     &   0.592 $\pm$ 0.035             \\
                              &     Mistral        & Spec.                                                                 & 0.703 $\pm$ 0.021                &   0.292 $\pm$ 0.021                        &     0.300 $\pm$ 0.021                      &   0.639 $\pm$ 0.020             \\
                              &     Mixtral        & Spec.                                                                 &  0.704 $\pm$ 0.012               &    0.285 $\pm$ 0.017                       &     0.269 $\pm$ 0.014                      &   0.689 $\pm$ 0.020             \\  \cline{2-7}
                             &       ChatGPT        & K-Prec + Q2                                                              & 0.701 $\pm$ 0.021            & 0.178 $\pm$ 0.014                          &  0.188 $\pm$ 0.067                         &   0.648 $\pm$ 0.024             \\
                             &      LLaMa          & K-Prec + Q2                                                                &  0.712 $\pm$ 0.023             &  0.177 $\pm$ 0.028                        & 0.189 $\pm$ 0.014                          &  0.555 $\pm$ 0.038              \\
                            &       Mistral        & K-Prec + Q2                                                               &   0.691 $\pm$ 0.013              &   0.288 $\pm$ 0.015                        & 0.290 $\pm$ 0.186                          &  0.611 $\pm$ 0.015              \\
                           &        Mixtral        & K-Prec + Q2                                                             &   0.731 $\pm$ 0.032              &   0.276 $\pm$ 0.189                       &    0.291 $\pm$ 0.178                      &  0.672 $\pm$ 0.112             \\  \cline{2-7}
                             &      ChatGPT        & Spec. + K-Prec                                                        &  0.711 $\pm$ 0.031               & 0.184 $\pm$ 0.012                          &  0.182 $\pm$ 0.131                         &   0.665 $\pm$ 0.021             \\
                             &      LLaMa          & Spec. + K-Prec                                                        &  0.704 $\pm$ 0.067              &   0.184 $\pm$ 0.034                        & 0.187 $\pm$ 0.024                          &     0.578 $\pm$ 0.024          \\
                            &       Mistral        & Spec. + K-Prec                                                        &  0.686 $\pm$ 0.021               &   0.281 $\pm$ 0.071                       & 0.299 $\pm$ 0.045                         &   0.591 $\pm$ 0.021             \\
                         &          Mixtral        & Spec. + K-Prec                                                        &  0.728 $\pm$ 0.021               & 0.282 $\pm$ 0.031                        & 0.321 $\pm$ 0.141                        & 0.711 $\pm$ 0.027               \\ \cline{2-7}
                            &       ChatGPT        & Spec. + Q2                                                                 &     0.721 $\pm$ 0.172            &  0.167 $\pm$ 0.021                        &  0.174 $\pm$ 0.019                        &  0.661 $\pm$ 0.027            \\

                             &      LLaMa          & Spec. + Q2                                                              &       0.718 $\pm$ 0.089         &   0.174 $\pm$ 0.056                        &   0.188 $\pm$ 0.065                        & 0.588 $\pm$ 0.041               \\
                             &      Mistral        & Spec. + Q2                                                                 &   0.688 $\pm$ 0.129              &   0.267 $\pm$ 0.014                        &  0.271 $\pm$ 0.011                         & 0.578 $\pm$ 0.018             \\
                               &    Mixtral        & Spec. + Q2                                                                &   0.741 $\pm$ 0.121              &   0.282 $\pm$ 0.073                       &   0.330 $\pm$ 0.021                      &  0.676 $\pm$ 0.043              \\  \cline{2-7}

                             &      ChatGPT        & K-Prec + Q2 + Spec.                                                   & 0.762 $\pm$ 0.014                &   0.241 $\pm$ 0.021                        & 0.241 $\pm$ 0.178                          &  0.614 $\pm$ 0.172              \\
                              &     LLaMa          & K-Prec + Q2 + Spec.                                                   & 0.740 $\pm$ 0.021                &  0.227 $\pm$ 0.018                         & 0.238 $\pm$ 0.019                          & 0.626 $\pm$ 0.016               \\
                              &     Mistral        & K-Prec + Q2 + Spec.                 &   0.762 $\pm$ 0.013
                                 &  0.304 $\pm$ 0.016               &  0.326 $\pm$ 0.018                         &          0.646 $\pm$ 0.221                                 \\
                              &     Mixtral        & K-Prec + Q2 + Spec.                                                   &  \textbf{0.781 $\pm$ 0.012}               &  \textbf{0.322 $\pm$ 0.012}                          &   \textbf{0.344 $\pm$ 0.017}                        & \textbf{0.721 $\pm$ 0.018} \\ \hline              
\end{tabular}}
\caption{Performance of ReMuSE with different combinations of rewards on the Meta-Reviewing Domain.}
\label{tab:ReMuSE_all}
\end{table*}

\begin{table*}[]
\centering
\begin{adjustbox}{width=0.9\linewidth}
% Please add the following required packages to your document preamble:
% \usepackage{multirow}
\begin{tabular}{c|cccccc}
\hline
\multicolumn{1}{l}{\textbf{Domain}}       & \textbf{Model} & \textbf{\begin{tabular}[c]{@{}c@{}}Rewards\end{tabular}} & \textbf{K-Prec} & \textbf{Q2-F1}            & \textbf{Q2-NLI}           & \textbf{Spec.} \\ \hline
\multirow{32}{*}{Product Buying} & ChatGPT        & -                                                                     & 0.497 $\pm$ 0.023          & \multicolumn{1}{l}{0.202 $\pm$ 0.014 } & \multicolumn{1}{l}{0.203 $\pm$ 0.011 } & 0.311 $\pm$ 0.024           \\
                                 & LLaMa          & -                                                                     & 0.462 $\pm$ 0.022            & \multicolumn{1}{l}{0.203 $\pm$ 0.034 } & \multicolumn{1}{l}{0.206 $\pm$ 0.044 } & 0.448 $\pm$ 0.017          \\
                                 & Mistral        & -                                                                     & 0.523 $\pm$ 0.015            & \multicolumn{1}{l}{0.225 $\pm$ 0.022 } & \multicolumn{1}{l}{0.225 $\pm$ 0.019 } & 0.427 $\pm$ 0.028          \\
                                 & Mixtral        & -                                                                     & 0.461 $\pm$ 0.026            & \multicolumn{1}{l}{0.263 $\pm$ 0.039 } & \multicolumn{1}{l}{0.323 $\pm$ 0.045 } & 0.424 $\pm$ 0.076           \\  \cline{2-7}
                                 
                                 & ChatGPT        & K-Prec                                                                & 0.539 $\pm$ 0.010                &   0.226 $\pm$ 0.012                        &  0.240 $\pm$ 0.024                         & 0.441 $\pm$ 0.012               \\
                                 & LLaMa          & K-Prec                                                                & 0.580 $\pm$ 0.032                &   0.236 $\pm$ 0.014                        &    0.246 $\pm$ 0.321                       &     0.541 $\pm$ 0.032           \\
                                 & Mistral        & K-Prec                                                                &  0.582 $\pm$ 0.036               &    0.267 $\pm$ 0.153                       &  0.267 $\pm$ 0.242                         &     0.475 $\pm$ 0.128           \\
                                 & Mixtral        & K-Prec                                                                &   0.622 $\pm$ 0.010               &  0.289 $\pm$ 0.011                         &    0.287 $\pm$0.152                       &    0.529 $\pm$ 0.009            \\  \cline{2-7}
                                 & ChatGPT        & Q2 (F1 + NLI)                                                        &  0.533 $\pm$ 0.029               &   0.287 $\pm$ 0.018                        &   0.294 $\pm$ 0.015                        & 0.438 $\pm$ 0.015               \\
                                 & LLaMa          & Q2 (F1 + NLI)                                                        &  0.544 $\pm$ 0.006             &    0.242 $\pm$ 0.018                       &   0.257 $\pm$ 0.125                        & 0.522 $\pm$ 0.025               \\
                                 & Mistral        & Q2 (F1 + NLI)                                                        &  0.593 $\pm$ 0.032                &  0.278 $\pm$ 0.036                         &   0.282 $\pm$ 0.036                        &  0.467 $\pm$ 0.042              \\
                                 & Mixtral        & Q2 (F1 + NLI)                                                       & 0.662 $\pm$ 0.004                &    0.335 $\pm$ 0.013                       &  0.347 $\pm$ 0.011                         &  0.525 $\pm$ 0.004              \\ \cline{2-7}
                                 & ChatGPT        & Spec.                                                                 & 0.643 $\pm$ 0.012                & 0.267 $\pm$ 0.019                          &  0.276 $\pm$ 0.019                         &  0.453 $\pm$ 0.039              \\
                                 & LLaMa          & Spec.                                                                 &  0.621 $\pm$ 0.023               & 0.226 $\pm$ 0.027                          &   0.239 $\pm$ 0.030                        &  0.589 $\pm$ 0.029              \\
                                 & Mistral        & Spec.                                                                 &  0.634 $\pm$ 0.037               &   0.258 $\pm$ 0.054                        &      0.276 $\pm$ 0.069                     &    0.602 $\pm$ 0.063            \\
                                 & Mixtral        & Spec.                                                                 & 0.642 $\pm$ 0.015                &      0.342 $\pm$ 0.032                     &     0.344 $\pm$ 0.035                      &   0.592 $\pm$ 0.001             \\  \cline{2-7}
                                 & ChatGPT & SR     & 0.510 $\pm$ 0.018 & 0.215 $\pm$ 0.020 & 0.220 $\pm$ 0.018 & 0.330 $\pm$ 0.030 \\
& LLaMa   & SR     & 0.485 $\pm$ 0.020 & 0.218 $\pm$ 0.022 & 0.225 $\pm$ 0.022 & 0.460 $\pm$ 0.018 \\
& Mistral & SR     & 0.540 $\pm$ 0.022 & 0.240 $\pm$ 0.020 & 0.245 $\pm$ 0.019 & 0.440 $\pm$ 0.025 \\
& Mixtral & SR     & 0.495 $\pm$ 0.025 & 0.270 $\pm$ 0.030 & 0.310 $\pm$ 0.032 & 0.435 $\pm$ 0.040 \\ \cline{2-7}
% Self-Consistency + Self-Refine (SC+SR)
& ChatGPT & SC+SR  & 0.520 $\pm$ 0.015 & 0.225 $\pm$ 0.025 & 0.230 $\pm$ 0.021 & 0.345 $\pm$ 0.032 \\
& LLaMa   & SC+SR  & 0.495 $\pm$ 0.018 & 0.230 $\pm$ 0.020 & 0.235 $\pm$ 0.018 & 0.470 $\pm$ 0.020 \\
& Mistral & SC+SR  & 0.550 $\pm$ 0.020 & 0.250 $\pm$ 0.025 & 0.260 $\pm$ 0.020 & 0.455 $\pm$ 0.030 \\
& Mixtral & SC+SR  & 0.505 $\pm$ 0.022 & 0.275 $\pm$ 0.028 & 0.315 $\pm$ 0.030 & 0.445 $\pm$ 0.035 \\ \cline{2-7}
% Original entries continue
                                 & ChatGPT        & K-Prec + Q2                                                              & 0.623 $\pm$ 0.017             &    0.243 $\pm$ 0.035                      &      0.260 $\pm$ 0.121                     &    0.463 $\pm$ 0.051          \\
                                 & LLaMa          & K-Prec + Q2                                                                &  0.631 $\pm$ 0.046             &    0.234 $\pm$ 0.129                     &  0.237 $\pm$ 0.038                         &        0.552 $\pm$ 0.042        \\
                                 & Mistral        & K-Prec + Q2                                                               &  0.641 $\pm$ 0.043             & 0.242 $\pm$ 0.117                          &  0.264 $\pm$ 0.015                         &    0.588 $\pm$ 0.052            \\
                                 & Mixtral        & K-Prec + Q2                                                             & 0.671 $\pm$ 0.432                &    0.326 $\pm$ 0.012                   &    0.341 $\pm$ 0.028                      &  0.602 $\pm$ 0.016             \\  \cline{2-7}
                                 & ChatGPT        & Spec. + K-Prec                                                        &  0.612 $\pm$ 0.015               &   0.241 $\pm$ 0.016                        &  0.311 $\pm$ 0.178                         &  0.524 $\pm$ 0.028             \\
                                 & LLaMa          & Spec. + K-Prec                                                        &  0.582 $\pm$ 0.016              &         0.248 $\pm$ 0.018                &   0.254 $\pm$ 0.048                        &  0.581 $\pm$ 0.063          \\
                                 & Mistral        & Spec. + K-Prec                                                        &  0.633 $\pm$ 0.061               &  0.241 $\pm$ 0.017                        &  0.263 $\pm$ 0.018                        &   0.572 $\pm$ 0.061           \\
                                 & Mixtral        & Spec. + K-Prec                                                        &   0.664 $\pm$ 0.038             &    0.327 $\pm$ 0.019            &       0.341 $\pm$ 0.026                  &  0.585 $\pm$ 0.042             \\ \cline{2-7}
                                 & ChatGPT        & Spec. + Q2                                                                 & 0.652 $\pm$ 0.041                &  0.267 $\pm$ 0.045                       &   0.287 $\pm$ 0.078                       &   0.535 $\pm$ 0.056           \\

                                 & LLaMa          & Spec. + Q2                                                              & 0.621 $\pm$ 0.064               &  0.257 $\pm$ 0.054                         &    0.268 $\pm$ 0.068                       & 0.591 $\pm$ 0.017              \\
                                 & Mistral        & Spec. + Q2                                                                 &  0.641 $\pm$ 0.074              &  0.256 $\pm$ 0.076                         &  0.322 $\pm$ 0.059                         &  0.622 $\pm$ 0.078            \\
                                 & Mixtral        & Spec. + Q2                                                                & 0.676 $\pm$ 0.051                & 0.358 $\pm$ 0.067                        &   0.361 $\pm$ 0.071                      &   0.621 $\pm$ 0.051            \\  \cline{2-7}
                                
                                 & ChatGPT        & K-Prec + Q2 + Spec.                                                   &   0.671 $\pm$ 0.014             &  0.284 $\pm$ 0.151                         &    0.314 $\pm$ 0.115                       &   0.555 $\pm$ 0.157             \\
                                 & LLaMa          & K-Prec + Q2 + Spec.                                                   & 0.637 $\pm$ 0.015                 &  0.286 $\pm$ 0.131                         &   0.299 $\pm$ 0.141                        &  0.581 $\pm$ 0.231              \\
                                 & Mistral        & K-Prec + Q2 + Spec.                                                   &  0.697 $\pm$ 0.051               &   0.339 $\pm$ 0.182                        &   0.351 $\pm$ 0.132                        & 0.604 $\pm$ 0.214               \\
                                 & Mixtral        & K-Prec + Q2 + Spec.                                                   &  \textbf{0.713 $\pm$ 0.141}               &  \textbf{0.372 $\pm$ 0.121}                         &   \textbf{0.404 $\pm$ 0.151}                        & \textbf{0.652 $\pm$ 0.191} \\ \hline              
\end{tabular}
\end{adjustbox}
\caption{Performance of ReMuSE on the Product Buying Domain. Best results are indicated in \textbf{bold}.}
\label{tab:products_results}
\end{table*}

\begin{table*}[]
\centering
\begin{adjustbox}{width=0.9\linewidth}
% Please add the following required packages to your document preamble:
% \usepackage{multirow}
\begin{tabular}{c|cccccc}
\hline
\multicolumn{1}{l}{\textbf{Domain}}       & \textbf{Model} & \textbf{\begin{tabular}[c]{@{}c@{}}Rewards\end{tabular}} & \textbf{K-Prec} & \textbf{Q2-F1}            & \textbf{Q2-NLI}           & \textbf{Spec.} \\ \hline
\multirow{32}{*}{ Debates} & ChatGPT        & -                                                                     & 0.374 $\pm$ 0.011          & \multicolumn{1}{l}{ 0.171 $\pm$ 0.121} & \multicolumn{1}{l}{0.202 $\pm$ 0.141
 } & 0.398 $\pm$ 0.141         \\
                                 & LLaMa          & -                                                                     & 0.377 $\pm$ 0.021            & \multicolumn{1}{l}{0.177
$\pm$ 0.214 } & \multicolumn{1}{l}{0.209 $\pm$ 0.151
 } &  0.427 $\pm$ 0.115        \\
                                 & Mistral        & -                                                                     & 0.406 $\pm$ 0.022            & \multicolumn{1}{l}{ 0.223 $\pm$ 0.113} & \multicolumn{1}{l}{ 0.246 $\pm$ 0.133
} &  0.443 $\pm$ 0.122         \\
                                 & Mixtral        & -                                                                     & 0.411 $\pm$ 0.015            & \multicolumn{1}{l}{ 0.208 $\pm$ 0.141} & \multicolumn{1}{l}{ 0.226 $\pm$ 0.115
} & 0.451 $\pm$ 0.114         \\  \cline{2-7}
& ChatGPT        & SR         & 0.390 $\pm$ 0.012 & 0.180 $\pm$ 0.120 & 0.215 $\pm$ 0.130 & 0.420 $\pm$ 0.120 \\
& LLaMa          & SR         & 0.395 $\pm$ 0.015 & 0.185 $\pm$ 0.125 & 0.220 $\pm$ 0.135 & 0.430 $\pm$ 0.110 \\
& Mistral        & SR         & 0.420 $\pm$ 0.018 & 0.230 $\pm$ 0.110 & 0.245 $\pm$ 0.120 & 0.450 $\pm$ 0.115 \\
& Mixtral        & SR         & 0.430 $\pm$ 0.016 & 0.235 $\pm$ 0.115 & 0.250 $\pm$ 0.118 & 0.460 $\pm$ 0.110 \\ \cline{2-7}
% Self-Consistency + Self-Refine (SC+SR)
& ChatGPT        & SC+SR      & 0.400 $\pm$ 0.010 & 0.190 $\pm$ 0.110 & 0.225 $\pm$ 0.125 & 0.440 $\pm$ 0.105 \\
& LLaMa          & SC+SR      & 0.405 $\pm$ 0.012 & 0.195 $\pm$ 0.112 & 0.230 $\pm$ 0.120 & 0.450 $\pm$ 0.100 \\
& Mistral        & SC+SR      & 0.430 $\pm$ 0.015 & 0.240 $\pm$ 0.108 & 0.255 $\pm$ 0.115 & 0.470 $\pm$ 0.108 \\
& Mixtral        & SC+SR      & 0.440 $\pm$ 0.014 & 0.245 $\pm$ 0.110 & 0.260 $\pm$ 0.112 & 0.480 $\pm$ 0.105 \\ \cline{2-7}
                                 & ChatGPT        & K-Prec                                                                &   0.444 $\pm$ 0.006              &    0.180 $\pm$ 0.131                       &     0.215 $\pm$ 0.151                      &   0.482 $\pm$ 0.221              \\
                                 & LLaMa          & K-Prec                                                                &   0.414 $\pm$ 0.012          &   0.187 $\pm$ 0.152                        &     0.214 $\pm$ 0.042                      & 0.529 $\pm$ 0.033               \\
                                 & Mistral        & K-Prec                                                                &   0.486 $\pm$ 0.004              & 0.235 $\pm$ 0.151                          &     0.244 $\pm$ 0.041                      &    0.547 $\pm$ 0.003            \\
                                 & Mixtral        & K-Prec                                                                &  0.483 $\pm$ 0.009               &  0.246  $\pm$ 0.015                          &  0.241 $\pm$ 0.141                         &    0.581 $\pm$ 0.033            \\  \cline{2-7}
                                 & ChatGPT        & Q2 (F1 + NLI)                                                        &  0.421 $\pm$ 0.028              & 0.194 $\pm$ 0.020                          &    0.228 $\pm$ 0.024                       &  0.481 $\pm$ 0.027              \\
                                 & LLaMa          & Q2 (F1 + NLI)                                                        &   0.412 $\pm$ 0.027            &   0.216 $\pm$ 0.039                        &  0.224 $\pm$ 0.045                         &  0.518 $\pm$ 0.033              \\
                                 & Mistral        & Q2 (F1 + NLI)                                                        &    0.487 $\pm$ 0.039              &   0.251 $\pm$ 0.043                        &    0.277 $\pm$ 0.045                       &    0.552 $\pm$ 0.015            \\
                                 & Mixtral        & Q2 (F1 + NLI)                                                       &   0.504 $\pm$ 0.009              &   0.268 $\pm$ 0.034                        & 0.288 $\pm$ 0.010                          &  0.574 $\pm$ 0.022              \\ \cline{2-7}
                                 & ChatGPT        & Spec.                                                                 & 0.414 $\pm$ 0.017               &  0.177 $\pm$ 0.018                          &   0.186 $\pm$ 0.027                        &   0.494 $\pm$ 0.018             \\
                                 & LLaMa          & Spec.                                                                 &  0.400 $\pm$ 0.031                & 0.181 $\pm$ 0.047                          &   0.190 $\pm$ 0.044                        &  0.540 $\pm$ 0.032              \\
                                 & Mistral        & Spec.                                                                 & 0.484 $\pm$ 0.024                &   0.249 $\pm$ 0.046                        &      0.250  $\pm$ 0.048                     &    0.581 $\pm$ 0.018            \\
                                 & Mixtral        & Spec.                                                                 & 0.491 $\pm$ 0.019                &     0.252 $\pm$ 0.012                      &     0.261 $\pm$ 0.017                      & 0.601 $\pm$ 0.010               \\  \cline{2-7}
                                 & ChatGPT        & K-Prec + Q2                                                              &   0.419 $\pm$ 0.057          &   0.208 $\pm$ 0.054                       &    0.214 $\pm$ 0.043                       &   0.526 $\pm$ 0.051           \\
                                 & LLaMa          & K-Prec + Q2                                                                & 0.411 $\pm$ 0.064              &    0.192 $\pm$ 0.051                     &   0.201 $\pm$ 0.054                        &    0.537 $\pm$ 0.018            \\
                                 & Mistral        & K-Prec + Q2                                                               & 0.473 $\pm$ 0.051              &   0.262 $\pm$ 0.043                        &    0.278 $\pm$ 0.052                       &    0.574 $\pm$ 0.087            \\
                                 & Mixtral        & K-Prec + Q2                                                             &  0.494 $\pm$ 0.064               &  0.268 $\pm$ 0.018                     &    0.267 $\pm$ 0.064                      &    0.628 $\pm$ 0.065           \\  \cline{2-7}
                                 & ChatGPT        & Spec. + K-Prec                                                        &   0.422 $\pm$ 0.052              &  0.201 $\pm$ 0.044                         &  0.209 $\pm$ 0.116                         &    0.484 $\pm$ 0.054           \\
                                 & LLaMa          & Spec. + K-Prec                                                        &      0.428 $\pm$ 0.089          &      0.194 $\pm$ 0.062                   & 0.205 $\pm$ 0.067                          &    0.542 $\pm$ 0.144        \\
                                 & Mistral        & Spec. + K-Prec                                                        &      0.489 $\pm$ 0.057           &   0.244 $\pm$ 0.067                       &   0.259 $\pm$ 0.089                       &    0.584 $\pm$ 0.056          \\
                                 & Mixtral        & Spec. + K-Prec                                                        &    0.484 $\pm$ 0.078            &  0.252 $\pm$ 0.056              &   0.252 $\pm$ 0.018                      &   0.624 $\pm$ 0.076            \\ \cline{2-7}
                                 & ChatGPT        & Spec. + Q2                                                                 &  0.448 $\pm$ 0.197               &  0.201 $\pm$ 0.069                       &  0.212 $\pm$ 0.068                        &  0.502 $\pm$ 0.074            \\

                                 & LLaMa          & Spec. + Q2                                                              &  0.421 $\pm$ 0.079              &   0.221 $\pm$ 0.168                        &  0.248 $\pm$ 0.156                         &  0.551 $\pm$ 0.018             \\
                                 & Mistral        & Spec. + Q2                                                                 &    0.486 $\pm$ 0.058              &  0.248 $\pm$ 0.188                       &   0.268 $\pm$  0.076                       &  0.591 $\pm$ 0.054            \\
                                 & Mixtral        & Spec. + Q2                                                                &  0.510 $\pm$ 0.098               &   0.278 $\pm$ 0.088                      &    0.281 $\pm$ 0.045                     & 0.621 $\pm$ 0.077              \\  \cline{2-7}

                                 & ChatGPT        & K-Prec + Q2 + Spec.                                                   &     0.501 $\pm$ 0.016            &  0.221 $\pm$ 0.114                         &  0.232 $\pm$ 0.168                         & 0.535 $\pm$ 0.132               \\
                                 & LLaMa          & K-Prec + Q2 + Spec.                                                   &  0.441 $\pm$ 0.131               &   0.234 $\pm$ 0.178                       &      0.244 $\pm$ 0.132                     & 0.565 $\pm$ 0.151               \\
                                 & Mistral        & K-Prec + Q2 + Spec.                                                   & 0.501 $\pm$ 0.112                & 0.274 $\pm$ 0.213                          & 0.291 $\pm$ 0.121                          &   0.627 $\pm$ 0.189             \\
                                 & Mixtral        & K-Prec + Q2 + Spec.                                                   & \textbf{0.522 $\pm$ 0.131}                & \textbf{0.277 $\pm$ 0.122}                           & \textbf{0.298 $\pm$ 0.152}                           & \textbf{0.628 $\pm$ 0.121} \\ \hline              
\end{tabular}
\end{adjustbox}
\caption{Performance of ReMuSE on the Debates Domain. Best results are indicated in \textbf{bold}.}
\label{tab:debates_results}
\end{table*}

\begin{mybox}{Prompt for Refinement (Debates)}
\small
\textbf{Instruction:} Given the feedback, knowledge source, and dialogue improve the dialogue based on the feedback provided. The output should just be the new dialogue.

\vspace{3pt}

\textbf{Knowledge Source:} \textit{\textbf{For arguments}: 'Let's talk about what would happen to real people in real families if the U.S. were to go down this untested and quite radical path of eliminating the minimum wage as our opponents urge...', 
\textbf{Against Arguments}: `We need to abolish the minimum wage. And there's only one argument that matters on this issue, and it's the moral argument. Does the minimum wage make the world a better place?...'}

\vspace{3pt}
\textbf{Feedback}:  \textit{It could benefit from more specific and concise answers to the questions asked by the decision-maker. 1. The Q2 F1 score could be improved by providing more specific examples and evidence to support the arguments presented. For example, when discussing the argument for abolishing the minimum wage, the dialogue agent could provide examples of how higher labor costs have negatively impacted businesses in the past. 2. The Q2 NLI score could be improved by ensuring that the dialogue agent's responses are coherent and make sense in the context of the conversation. For example, when discussing the argument for keeping the minimum wage, the dialogue agent could provide more specific examples of how a living wage would benefit low-income families.}

\vspace{3pt}

\textbf{Dialogue:} \textit{Decision Maker: Hello, can you tell the arguments for and against debating for minimum wages? \textbackslash n Dialogue Agent: The debate over abolishing the minimum wage centers on
two perspectives. Supporters claim it harms businesses
and argue for market-driven wages without government
interference. Opponents assert the minimum wage is
crucial to shield low-income workers from exploitation,
contending that without it, wages might be insufficient for
basic needs. \textbackslash n Decision Maker: .. }

\vspace{3pt}

%\textbf{Sentence:} $<$\textit{question}$>$

%\vspace{3pt}

%\textbf{Paraphrases:}
\end{mybox}
\begin{mybox}{Prompt for Generic feedback (Meta-Reviewing)}
\small
\textbf{Instruction:} Given the knowledge source and the dialogue, please provide \textbf{feedback} to improve the dialogues. The feedback should just be for the overall dialogue and should start with `Feedback:'. 

\vspace{3pt}

\textbf{Knowledge Source:} \textit{ \textbf{Review 1}: The paper proposes a benchmark to measure the quality of the systems in the growing domain of code intelligence ..., \textbf{Review 2}: .., \textbf{Review 3}: ... }

\vspace{3pt}

\textbf{Dialogue:} \textit{Meta-Reviewer: Hello Dialogue Agent. Can you tell me more about this paper? \textbackslash n Dialogue Agent: Ofcourse! This is a paper about CodexGLUE, a machine learning benchmark dataset for code understanding and generation, \textbackslash n MetaReviewer: ... }

\end{mybox}
\vspace{3pt}
\begin{mybox}{Prompt for Actionable feedback (Meta-Reviewing)}
\small
\textbf{Instruction:} Given the knowledge source and the dialogue, please provide \textbf{actionable feedback} to improve the dialogues. The feedback should be for the overall dialogue and start with `Feedback:.' The feedback should improve the groundedness of the dialogue agent's utterances to the knowledge source. The feedback should also increase the specificity (higher technical details) of utterances for the dialogue agent and the meta-reviewer.

\vspace{3pt}

\textbf{Knowledge Source:} \textit{ \textbf{Review 1}: The paper proposes a benchmark to measure the quality of the systems in the growing domain of code intelligence ..., \textbf{Review 2}: .., \textbf{Review 3}: ... }

\vspace{3pt}

\textbf{Dialogue:} \textit{Meta-Reviewer: Hello Dialogue Agent. Can you tell me more about this paper? \textbackslash n Dialogue Agent: Ofcourse! This is a paper about CodexGLUE, a machine learning benchmark dataset for code understanding and generation, \textbackslash n MetaReviewer: ... }

\vspace{3pt}

%\textbf{Sentence:} $<$\textit{question}$>$

%\vspace{3pt}

%\textbf{Paraphrases:}
\end{mybox}

\subsection{Cross-Domain analysis for the performance of ReMuSE}\label{sec:result_all}
Table~\ref{tab:ReMuSE_all} shows the detailed performance of ReMuSE with multiple combinations of rewards for the meta-reviewing domain. We discuss the cross-domain analysis of ReMuSE on two other decision-making domains namely debates and product reviews below.

\subsubsection{Domains} \label{sec:cross_dom}
In order to test the robustness of our approach, ReMuSE, we experiment with two other domains of decision-making based on lengthy input documents. In \textit{product-buying}, a buyer needs to decide for or against a product based on the provided customer reviews. In \textit{debates}, a listener must decide on one of the stances based on the provided arguments.

In this section, we initially discuss the datasets we employed to perform the analysis, followed by a discussion on our experimental findings.

\subsubsection{Data}For \textit{product-buying}, we use the Helpful Reviews Dataset \cite{gamzu-etal-2021-identifying}. The reviews are available for $123$ products, and each review comes annotated with a helpfulness score. Following the recommendation of \citet{gamzu-etal-2021-identifying}, we consider all the reviews scored above 1, which are the most helpful for making buying decisions.
For \textit{debates}, we use the IQ2 Debates Dataset \cite{zhang-etal-2016-conversational}. The dataset consists of Oxford-Style Debates on $108$ topics. Each debate is a dialogue between multiple parties namely, moderator, pro-side, and con-side. For ease of use, we only consider the opening arguments from the pro- and con- sides of the debate for each topic.

\subsubsection{Results} In the case of product buying, we use all the helpful review sentences available for a product as the knowledge source, whereas the arguments from pro and con-sides of a topic in case of debates. We use similar prompts as in Sec~\S\ref{sec:prompts} but simulate different scenarios for both domains and then apply our method as is. The detailed prompts for initial dialogue generation, feedback, and refinement for both the domains are provided in \S\ref{sec:init_dialogue}, \S\ref{sec:feedback} and \S\ref{sec:refinement} respectively. We report the results for the \textit{product-buying} and \textit{debates} domain in Tables~\ref{tab:products_results} and \ref{tab:debates_results} in \S\ref{sec:result_all} respectively. The best results are attained by using an amalgamation of all the refinement metrics, similar to the meta-reviewing domain. We observe a similar trend in terms of the best-performing model as well where the Mixtral model excels across all the domains. 

\subsection{Ablation Study} \label{sec:ablation}
\subsubsection{Single vs Joint Reward Optimization} \label{sec:app_single_vs_joint}
We show the scores for single vs joint score optimization in Table~\ref{tab:single_vs_joint}. We observe massive improvements when using the amalgamation of all the refinement scores.
\begin{table}[H]
\centering

% Please add the following required packages to your document preamble:
% \usepackage{multirow}
\small{\begin{tabular}{lllll}
% \resizebox{0.5\textwidth}{!}{\begin{tabular}{lllll}
 %\begin{tabular}{ccccc}
\hline
\textbf{Reward} & \textbf{K-Prec} & \textbf{Q2-F1}            & \textbf{Q2-NL}           & \textbf{Spec.} \\ \hline
                                 K-Prec                                                                &   $0.70_{.01}$              &   $0.27_{.02}$                        &    $0.26_{.02}$                       &      $0.65_{.03}$          \\
                                 
                                 Q2                                                       &  $0.72_{.01}$               &    $0.27_{.02}$                       &   $0.27_{.01}$                        & $0.70_{.03}$               \\
                                 
                                 Spec.                                                                 & $0.70_{.01}$                &   $0.29_{.02}$                        &     $0.27_{.01}$                      &   $0.69_{.02}$             \\ \hline
                                 
                                 K-Prec + Q2                                                               &   $0.73_{.03}$              &   $0.28_{.15}$                        & $0.29_{.18}$                          &  $0.67_{.11}$              \\
                                 
                                 Spec. + K-Prec                                                        &  $0.73_{.02}$               &   $0.28_{.03}$                       & $0.32_{.14}$                         &   $0.71_{.03}$             \\
                                 
                                 Spec. + Q2                                                                 &   $0.74_{.12}$              &   $0.28_{.07}$                        &  $0.33_{.02}$                         & $0.68_{.04}$             \\ \hline
                                 
                                 ReMuSE                 &   $\textbf{0.78}_{.01}$
                                 &  $\textbf{0.32}_{.01}$               &  $\textbf{0.34}_{.02}$                         &          $\textbf{0.72}_{.02}$                                 \\        \hline   
\end{tabular}}
%\end{tabular}

\caption{Single vs Joint-Score Optimization of ReMuSE for our best-performing model, \textbf{Mixtral} in terms of various refinement metrics.}
\label{tab:single_vs_joint}
\end{table}

\subsubsection{Multi-Iterative Refinement} \label{sec:multi_iter_app}
\begin{figure}[!htb]
\centering
    \begin{subfigure}[b]{0.22\textwidth}
        \includegraphics[width=\textwidth]{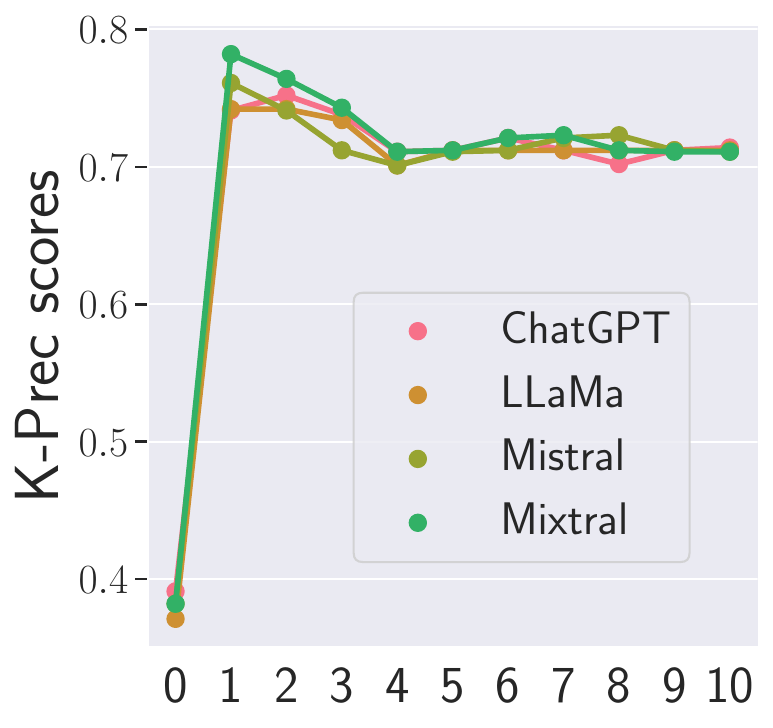}
        \caption{K-Precision}
        \end{subfigure}%
        \begin{subfigure}[b]{0.22\textwidth}
        \includegraphics[width=\textwidth]{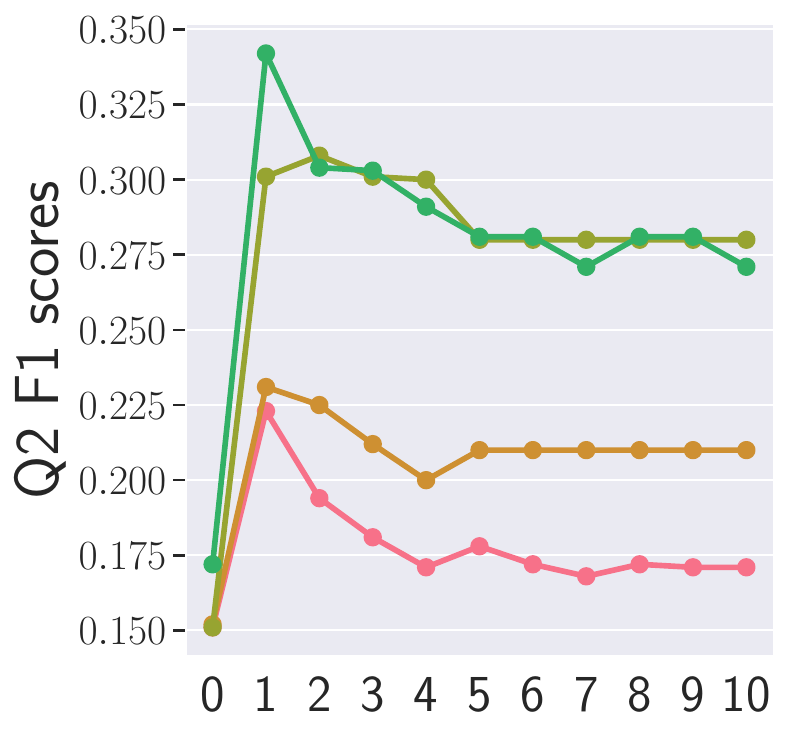}
        \caption{Q2-F1}
        \end{subfigure}%

        \begin{subfigure}[b]{0.22\textwidth}
         \includegraphics[width=\textwidth]{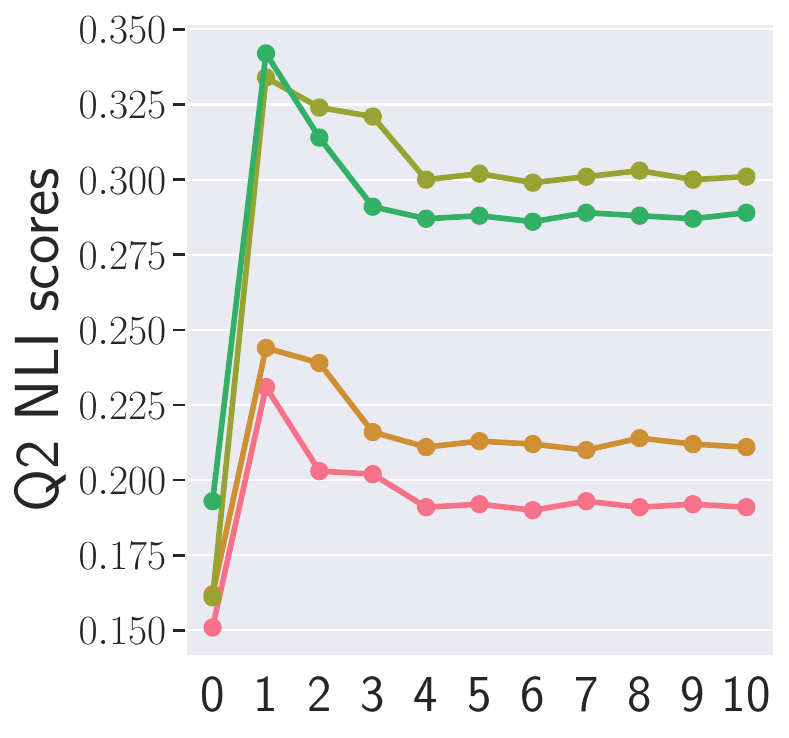}
        \caption{Q2-NLI}
        \end{subfigure}%
        \begin{subfigure}[b]{0.22\textwidth}
        \includegraphics[width=\textwidth]{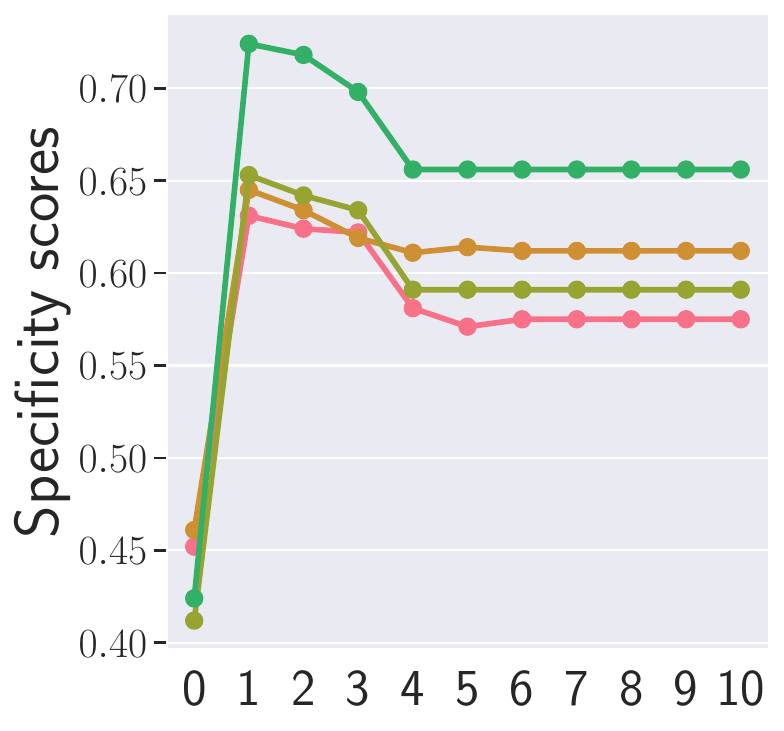}
        \caption{Specificity}
    \end{subfigure}
    \caption{For the iterative refinement version of ReMuSE, we plot the (a) K-Precision, (b) Q2-F1, (c) Q2-NLI, and, (d) Specificity scores with varying iterations ($0-4$). All the results are averaged over $3$ prompts.  }
    \label{fig:iter_vs_zero}
\end{figure}

In Fig~\ref{fig:iter_vs_zero}, we plot the scores for each iteration of ReMuSE up to 10 epochs and observe that the scores plateau after the first epoch, indicating superior feedback quality in the first round, consistent with prior work~\cite{madaan2023self}. 
\begin{figure}[!t]
\centering
    \begin{subfigure}[b]{0.22\textwidth}
        \includegraphics[width=\textwidth]{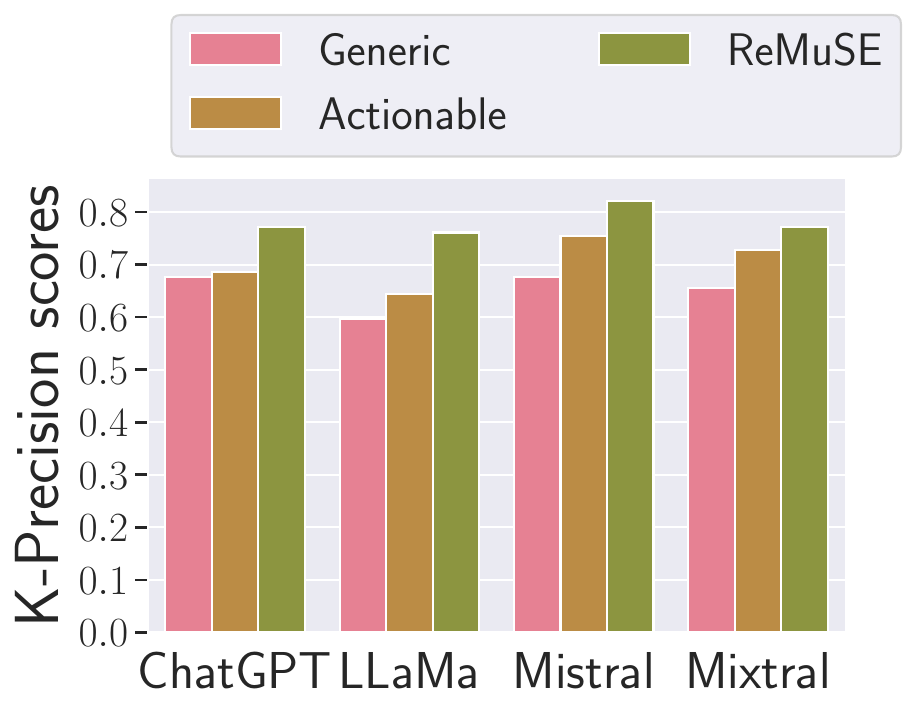}
        \caption{K-Precision}
        \end{subfigure}%
        \begin{subfigure}[b]{0.22\textwidth}
        \includegraphics[width=\textwidth]{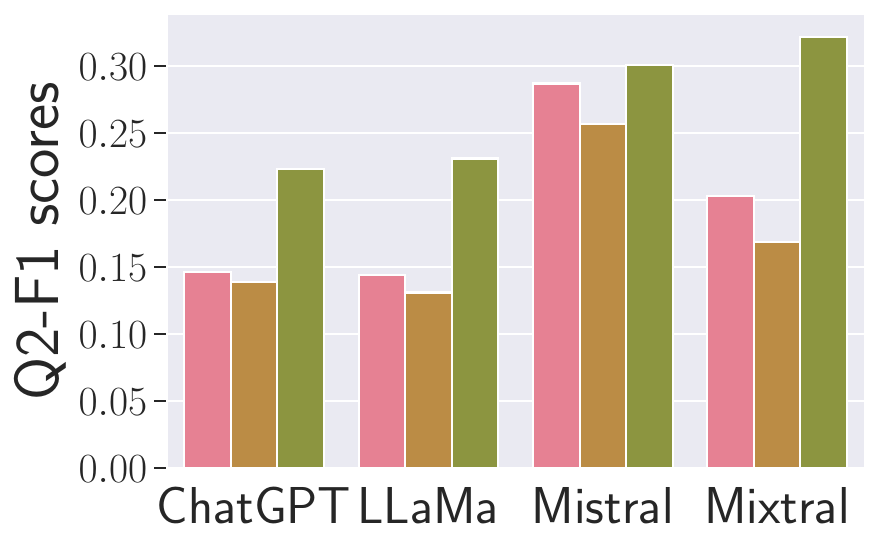}
        \caption{Q2-F1}
        \end{subfigure}%

        \begin{subfigure}[b]{0.22\textwidth}
         \includegraphics[width=\textwidth]{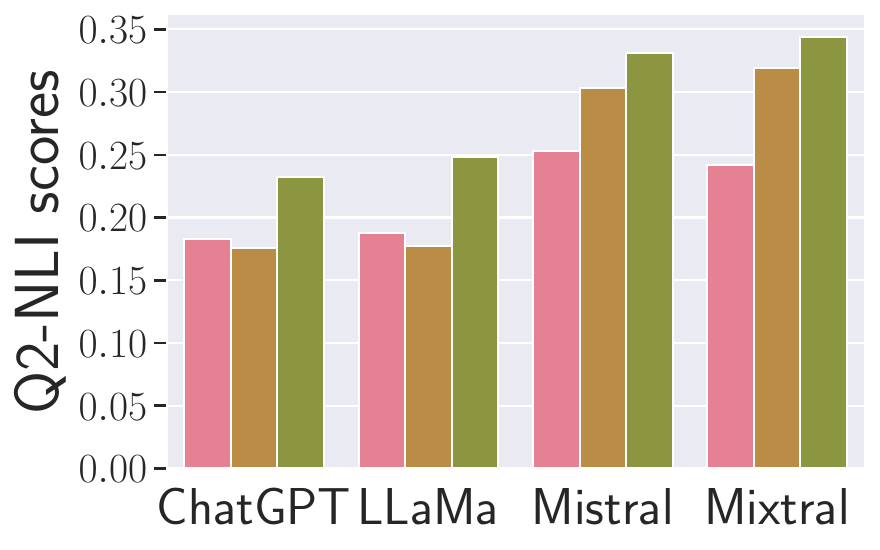}
        \caption{Q2-NLI}
        \end{subfigure}%
        \begin{subfigure}[b]{0.22\textwidth}
        \includegraphics[width=\textwidth]{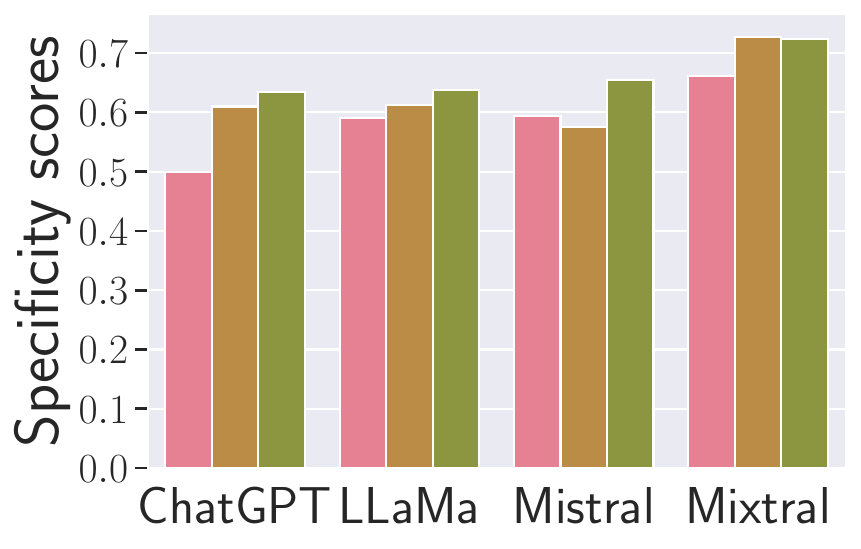}
        \caption{Specificity}
    \end{subfigure}
    \caption{Performances for the different feedback types Generic, Actionable, and ReMuSE. We plot the (a) K-Precision, (b) Q2-F1, (c) Q2-NLI, and, (d) Specificity scores for each (averaged over 3 prompts).}
    \label{fig:gen_vs_act}
\end{figure}

\subsubsection{Generic vs Actionable vs Multi-Reward FeedBack}
We plot the various metrics for these multiple feedback strategies in Fig~\ref{fig:gen_vs_act}. We observe that the feedback-prompting in ReMuSE leads to the maximum improvements. We also show a few examples of the feedback generated for a dialogue using the prompts discussed above and our proposed method, ReMuSE in Table~\ref{tab:feedback_eg}.\\
\begin{table*}[]

\centering
\begin{tabularx}{\textwidth}{@{}lX@{}}
\hline
\textbf{Strategy}       & \textbf{Feedback}                                                                                                                                                                                                                                                                                                                                                                                                                                                                                                                                                                                                                                                                                                                                                                                                                                                                                                                                                                                                               \\ \hline
Generic    & The agent mostly summarized the concerns raised by the reviewers without suggesting any specific actions that could be taken to address those concerns. In the future, the agent could provide more specific suggestions or questions to help the meta-reviewer make a more informed decision. Additionally, the agent could try to paraphrase or restate the concerns raised by the reviewers in a clearer or more concise way to facilitate the conversation.                                                                                                                                                                                                                                                                                                                                                                                                                                                                                     \\ \hline
Actionable & Overall, it is advisable for the authors to further clarify their proposed algorithm by addressing the issues raised by the reviewers. For instance, they can provide a better definition of transportation polytope in Eq. (2) and explain the process of generating two views of the input images in more detail. It is also recommended that they compare their method with the state-of-the-art consensus clustering methods that are not deep learning-based as suggested by Reviewer 1. To address the concerns over the quality of the experiments, the authors can provide a more thorough experimental investigation and comparison, as well as a discussion of the novelty and contributions of their method with respect to similar methods proposed in prior work.                                                                                                                                                                          \\ \hline
ReMuSE     & Here are some suggestions to improve the dialogue: 1. Provide a more detailed
summary of the reviewers' concerns, such as the specific areas where the paper is lacking in novelty or clarity.  For example, the dialogue could mention the reviewers' comments about the outdated literature, vague algorithm description, and lack of analysis of results. 2. Offer specific recommendations to address the reviewers' concerns. For instance, the dialogue could suggest that the authors update the literature review with more recent work on consensus clustering, provide a more detailed explanation of the algorithm, and perform additional analysis of the results. 3. Highlight the positive aspects of the paper, such as the novelty of combining ensemble methods with deep clustering. The dialogue could also mention the reviewers' comments about the potential effectiveness of the proposed ensemble deep clustering algorithm. \\ \hline
 \end{tabularx}

\caption{Different types of feedback generated for the same dialogue using multiple feedback prompting strategies.}
 \label{tab:feedback_eg}
\end{table*}
\begin{figure*}[!t]
    \centering
    \includegraphics[width=\textwidth]{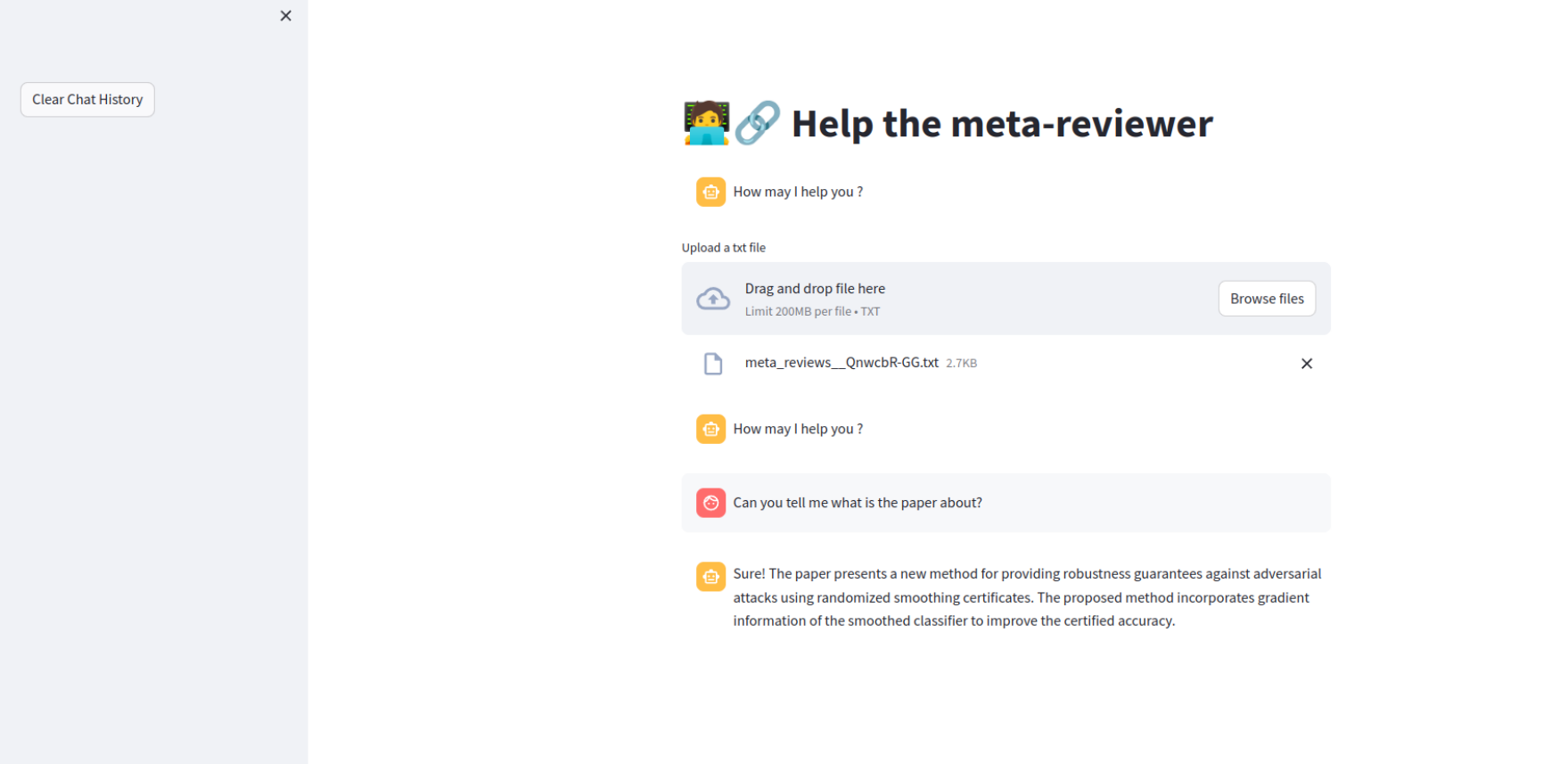}
    \caption{Screenshot of the Interface for assisting meta-reviewers with a dialogue agent} 
    \label{fig:dial_agent_inter}
\end{figure*}

\subsubsection{Correlation of human evaluation metrics}
\label{sec:human_corr}
We plot the correlation of the different metrics used for evaluating the dialogue agent's responses, namely, fluency, faithfulness, relevance, objectivity, and helpfulness, in Fig~\ref{fig:human_eval}. We observe a strong correlation between faithfulness, objectivity, and helpfulness. This further indicates a low tolerance for unfaithful responses in the decision-making process. The models finetuned on our data generate more faithful responses as demonstrated through the scores in human evaluation in Table~\ref{tab:response_eval} in Sec~\S\ref{sec:pre_refinement}.

\subsection{Resulting Dataset} \label{sec:decidial}
\begin{figure}
    \centering
    \includegraphics[width=\linewidth]{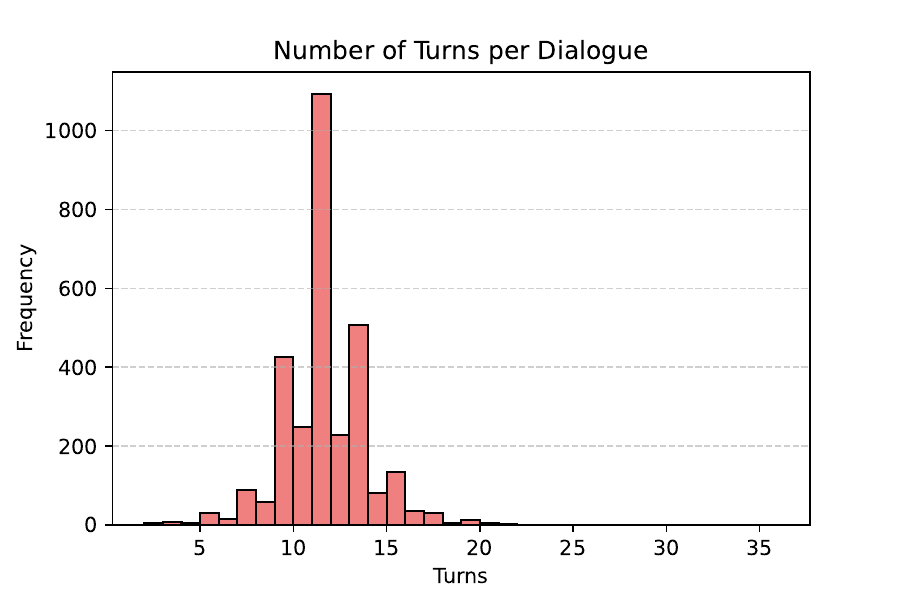}
    \caption{Distribution of Number of turns in Meta-Reviewing Dataset}
    \label{fig:meta_rev_turns}
\end{figure}

\begin{figure}
    \centering
    \includegraphics[width=\linewidth]{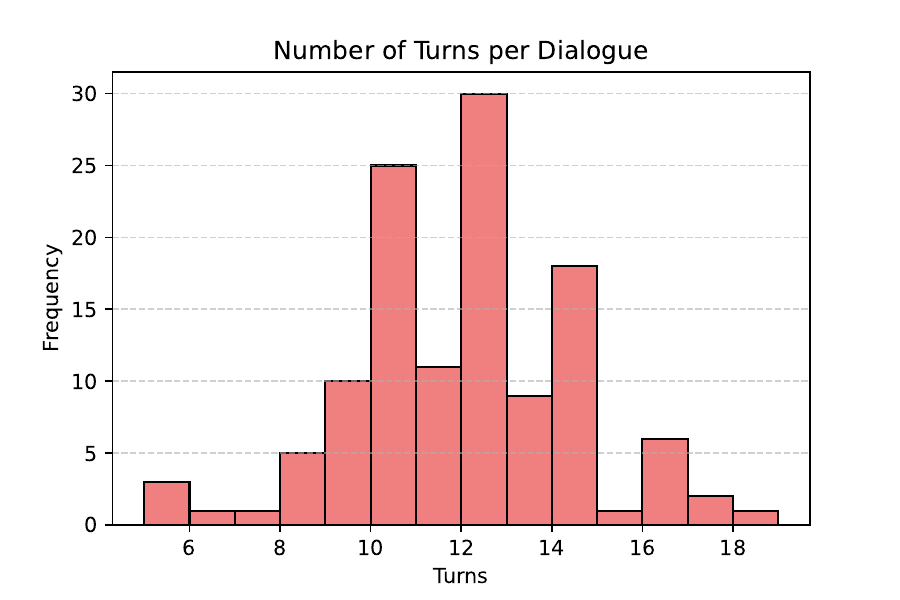}
    \caption{Distribution of Number of turns in Product Buying Dataset}
    \label{fig:product_turns}
\end{figure}

\begin{figure}
    \centering
    \includegraphics[width=\linewidth]{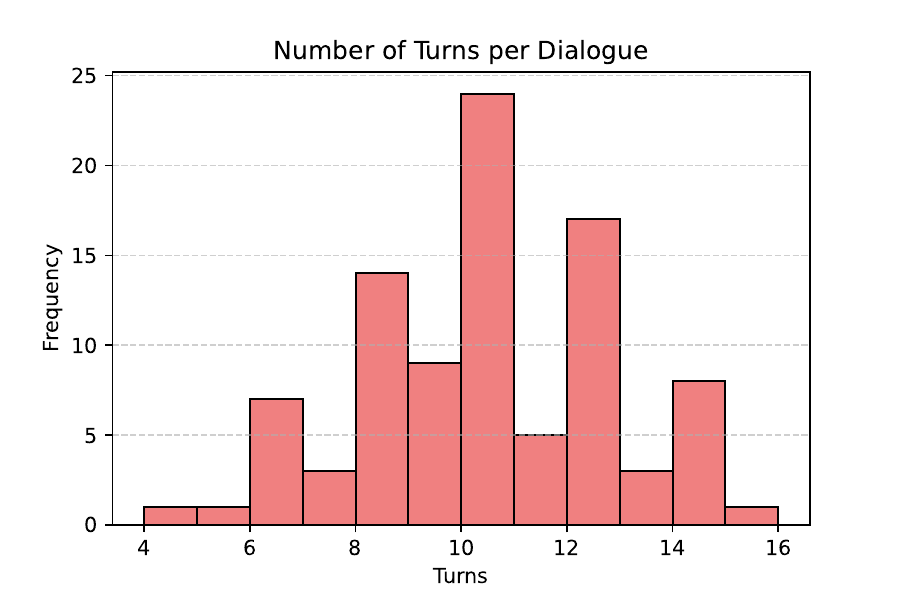}
    \caption{Distribution of Number of turns in Debates Dataset}
    \label{fig:debate_turns}
\end{figure}

As one of the contributions of this work, we make the synthetically generated dataset for the best-performing ReMuSE configuration publicly available. We release $3,064$ meta-reviewing, $128$ product-buying, and $108$ debate dialogues, respectively. We display the basic statistics for our resulting dataset in Table~\ref{tab:dataset}. Meta-Reviewing dialogues show the most diversity (unigram vocabulary size) in agent and meta-reviewer utterances, reflecting the specific terminology associated with the peer-reviewing domain. The debates domain closely follows due to diverse topics, while the product-buying domain exhibits the least diversity, reflecting the domain's simplicity. Meta-Reviewing dialogues are notably longer, with more turns depicting the longer reasoning steps in expert tasks. The turn distributions in our dataset are highlighted in Figures~\ref{fig:meta_rev_turns}, \ref{fig:product_turns}, \ref{fig:debate_turns} for meta-reviewing, product buying, and debates datasets, respectively.


\subsection{Hyperparameters  for Dialogue Response Generation} \label{sec:hyper-parameters}
All models are fine-tuned for 10 epochs. We perform a grid search on learning rates $\{1e-4, 3e-4, 5e-5, 6e-5\}$ and select the best model based on the validation set performance. The results are averaged over 3 random seeds. For ReMuSE, we use the best-performing model, Mixtral, and update the responses of ReMuSE for 1 iteration, given our prior findings. 
We adapt the same prompts as in \ref{sec:prompts} for the response generation task.

\vspace{2mm}
\begin{adjustbox}{width=0.7\linewidth}
\centering
\begin{mybox}{Response generation Prompt (Meta-Reviewing)}
\small
\textbf{Instruction:} Generate a response corresponding to a dialogue for a conversation between a meta-reviewer and a dialogue agent for reviews
about a paper. The
The dialogue agent has access to the dialogue history, reviews, and the type of paper. The dialogue agent should just answer from the reviews. A dialogue agent
should never recommend anything/ give any opinions/ decide anything for the paper. A
dialogue agent has no conscience of its own.

\vspace{3pt}
\textbf{Title}: [title]

\vspace{3pt}
\textbf{Knowledge Source:} \textit{ \textbf{Review 1}: The paper proposes a benchmark to measure the quality of the systems in the growing domain of code intelligence ..., \textbf{Review 2}: .., \textbf{Review 3}: ... }
\textbf{Dialogue History: } \textit{ \textbf{Meta-reviewer}: Hello, dialogue agent. Thank you for joining me today to discuss the reviews for the paper [title]. Can you please start by summarizing the main contributions of the paper and how it addresses the problem it sets out to solve?
}

\vspace{3pt}

%\textbf{Sentence:} $<$\textit{question}$>$

%\vspace{3pt}

%\textbf{Paraphrases:}

\end{mybox}
\end{adjustbox}
\vspace{2mm}
\begin{adjustbox}{width=0.7\linewidth}
\begin{mybox}{Prompt for Response generation ReMuSE feedback}
\small
\textbf{Instruction:} Given the knowledge source, dialogue history and the response, please provide actionable feedback to improve the response. The feedback should be for the response and should start with `Feedback:'. A Q2 F1 score, Q2 NLI score, KPrecision, and specificity scores. The Q2 and Kprecision scores measure the groundedness of the response to the provided knowledge source. The Specificity scores measure the technicality of the response.  The feedback should try to improve all scores for the dialogue agent

\vspace{1pt}

\textbf{Knowledge Source:} \textit{ \textbf{Review 1}: The paper proposes a benchmark to measure the quality of the systems in the growing domain of code intelligence ..., \textbf{Review 2}: .., \textbf{Review 3}: ... }

\vspace{1pt}

\textbf{Dialogue History:} \textit{ \textbf{Meta-reviewer}: Hello, dialogue agent. Thank you for joining me today to discuss the reviews for the paper [title]. Can you please start by summarizing the main contributions of the paper and how it addresses the problem it sets out to solve?
}\\
\textbf{Response: } \textit{ \textbf{Dialogue Agent}: Sure. The paper proposes a method based on importance and coherence to evaluate whether a partitioning of a neural network exhibits modular characteristics. F1:0.12, NLI:0.14 and Specificity: 0.1
}
\end{mybox}
\end{adjustbox}
\newpage
\begin{adjustbox}{width=0.7\linewidth}
\begin{mybox}{Prompt for Response generation ReMuSE Refinement}
\small
\textbf{Instruction:} Given the feedback, knowledge source, dialogue history and the previous response, improve the response based on the feedback provided. The output should just be the new dialogue.

\vspace{3pt}

\textbf{Knowledge Source:} \textit{ \textbf{Review 1}: The paper proposes a benchmark to measure the quality of the systems in the growing domain of code intelligence ..., \textbf{Review 2}: .., \textbf{Review 3}: ... }

\vspace{3pt}
\textbf{Feedback}:  \textit{It would be helpful to provide a clearer explanation of the tasks' relevance to software development in response to reviewer one's concerns. Additionally, it may be useful to mention that the paper discusses the effectiveness of the baselines and provides an ablation study. Finally, it would be good to acknowledge reviewer three's comment explicitly and mention any plans to revise the descriptions of the datasets.}

\vspace{3pt}

\textbf{Dialogue history:} \textit{ \textbf{Meta-reviewer}: Hello, dialogue agent. Thank you for joining me today to discuss the reviews for the paper [title]. Can you please start by summarizing the main contributions of the paper and how it addresses the problem it sets out to solve?
}\\
\textbf{Previous Response:} \textit{ \textbf{Dialogue Agent}: Sure. The paper proposes a method based on importance and coherence to evaluate whether a partitioning of a neural network exhibits modular characteristics. }

\vspace{3pt}

%\textbf{Sentence:} $<$\textit{question}$>$

%\vspace{3pt}

%\textbf{Paraphrases:}
\end{mybox}
\end{adjustbox}

\subsection{Meta-Reviewer Question Distribution} \label{sec:met_rev_q}
To quantify the different types of questions, we quantitatively analyze the different meta-reviewer questions in our dataset. To come up with the categories, we use the `Dagstuhl' dataset and annotate the questions in that dataset.  We find that most of the meta-reviewer questions are related to Contributions, Strengths, Weaknesses, Confidence, Avenues for Improvement, and Literature Survey and Follow-Up. We then prompt GPT 3.5-turbo to provide us with these labels for all the questions in our dataset. 
The distribution of questions in our dataset is provided in Fig~\ref{fig:type_dist}.  We observed that questions related to contributions, strengths, weaknesses, contributions, and confidence ($\sim$ 20\%) occur across all dialogues. These questions are the most important aspects for consideration while writing a meta-review, which is in line with the findings of \citet{santu2024prompting}. Thus, the questions in our dataset can serve as an effective resource for serving as a proxy meta-reviewer while training dialogue agents. 

\subsection{Error Analysis of the dialogue agent responses} \label{sec:err_ann}
Following \citet{li-etal-2023-autoconv}, we quantitatively analyze a set of 50 dialogues for error types in the dialogue agent's responses. Out of 150 dialogue agent turns, we observed 50 responses that may be considered erroneous. We identified four different kinds of errors in the responses: 1) \textbf{Irrelevant answers}: 75\% errors are incurred when the dialogue agent tries to answer questions that are based on external knowledge. For instance, for a meta-reviewer question on the literature survey of the paper, the dialogue agent answers about the novelty of the method in relation to the prior work as available in the reviewer comments. However, these responses are short, and the dialogue agent evades the answer by providing a partial/unsatisfactory answer; 2) \textbf{Inappropriate answers}: 10\% of the errors are incurred because of the lack of knowledge in using comparative adjectives. For instance, responses to questions regarding the confidence of a particular reviewer are often answered with the confidence of all the reviewers. This question is particularly asked as a follow-up to the severity of the weakness stated by the reviewers. However, the dialogue agent answers with the confidence of other reviewers using adjectives such as `low' or `high'. These responses are sometimes confusing since confidence levels, such as 4 out of 5, are not particularly low but are conceived as low because of the higher confidence of other reviewers. 3) \textbf{Repitition}: 10\% of responses have repeated facts as other answers and lack diversity. 4) \textbf{Lack of Neutrality}: In 5\% of the cases, the dialogue agent agrees with the meta-reviewer's evaluation of the paper, which is not intended since the dialogue agent should state its neutral stance. However, we did not observe any cases of disagreement with the meta-reviewer, which justifies the refinement metrics used in the dialogue generation. 

As a part of our research design, we induce faithfulness and technical specificity in the LLM generations. Thus, errors related to irrelevant answers occur due to increased faithfulness in the generations and the necessity to stick to the source content. The other errors are mainly related to limitations in LLMs in using comparative adjectives and the inherent pre-training-based alignment to agree with humans. However, none of the errors contradict our research design, and all the responses are still technical and faithful. This justifies the use of the dataset to train dialogue agents with higher faithfulness and technicality.

\begin{table}[t]
\centering
 \small{\begin{tabular}{lllll}
\hline
\textbf{Domain} & \textbf{\begin{tabular}[c]{@{}l@{}}\# dial.\end{tabular}}& \textbf{\begin{tabular}[c]{@{}l@{}}\# DA. \\ tok. (av.)\end{tabular}} & \textbf{\begin{tabular}[c]{@{}l@{}}\# Seek. \\ tok. (av.)\end{tabular}} & \textbf{\# tur.} \\ \hline
Meta-Rev.     &\textbf{3,064}    & \textbf{54.66}                                                                       & \textbf{30.95}                                                                       & \textbf{12.24}                 \\
Product-Buy.  & 123         & 28.87                                                                       & 13.49                                                                      & 11.55                 \\
Debates  & 108       & 48.51                                                             & 18.16                                                              & 10.0                \\
%Mixtral         & 59.63                                                                       & 25.44                                                                       & 10.52                \\ 
\hline
\end{tabular}}
\caption{Basic Statistics for the generated dataset. The table shows the number of dialogues, the average number of tokens for the dialogue agent and seeker utterances, and the number of turns.}
\label{tab:dataset}
\end{table}

\subsection{Human Evaluation Instruction for evaluating full dialogues} \label{sec:hum_ins_full_dial}
Please evaluate the dialogue in Table~\ref{tab:meta_rev_eg} with respect to the reviews in Sec~\S\ref{sec:meta-reviews} on a Likert Scale of $1-4$ over the following dimensions:\\
\noindent \textbf{Co-operativeness} means that the response of the
dialogue agent is coherent with the previous turn
and does not try to mislead the interlocutor or act
unhelpfully.\\
\noindent \textbf{Coherence} how coherent and consistent is the overall dialogue.\\
\noindent \textbf{Engagingness} implies the dialogue agent engages the interlocutor by prompting further replies and helps move the conversation forward.\\
\noindent \textbf{Plausibility} evaluate the groundedness of the dialogue agent’s response. The measure implies how reasonable the information is about the reviews and the paper as provided by the assistant.

\subsection{Human Evaluation Instruction for evaluating dialogue responses} \label{sec:hum_ins_resp}
Please evaluate the response, $u_t+1$ in connection to the Knowledge source, $\mathcal{K}$ and dialogue history $[u_1,u_2,\dots u_t]$ on a Likert scale of $1-5$ over the following dimensions:\\
\noindent \textbf{Fluency} How fluent is the language usage in the response? \\
\noindent \textbf{Relevance} How appropriate is the response to the dialogue history?\\
\noindent \textbf{Faithfulness} Is the response grounded/faithful to the knowledge source, $\mathcal{K}$. \\
\noindent \textbf{Objectivity} Is the answer factually correct rather than providing personal opinions and feelings? \\
\noindent \textbf{Helpfulness} How useful is this response if you had to decide the acceptance or rejection of this paper based on this response?\\

\noindent  \textbf{Knowledge Source, $\mathcal{K}$} 

\textit{Review 1}: This paper proposes a new type of generative models with a new inference method of latent variables. ..., \textit{Review 2}: The paper proposes GONs which seek to build a generative model with an “implicit” encoder that comes essentially for free with the use of a few re-parameterization tricks .., \textit{Review 3}: This paper introduces a "new" inference method for autoencoder-type models, where the encoder is taken as a gradient of the decoder with respect to a zero-initialized latent variable ... \\

\noindent \textbf{Dialogue History, $[u_1,u_2,\dots u_t]$}\\
\textit{Meta Reviewer}: Hello, can you summarize the main contributions of this paper "Gradient Origin Networks"?\\
\textit{Dialogue Agent}: Yes, of course. This paper proposes a new type of generative models called Gradient Origin Networks which use a new inference method of latent variables. Specifically, the gradient of latent variables with respect to the zero vector is taken as the inferred latent variables.\\
\textit{Meta Reviewer}: I see. Can you tell me more about the pros and cons mentioned in the reviews?]\\

\noindent \textbf{Response, $u_t+1$}

The reviews describe that the proposed method is easy and straightforward to implement. However, the model assumption that the one step gradient from zero vector equals to latent vector is quite limited and greatly constrains the model expressiveness. Also, the formulation needs to be carefully checked. The experimental results are not sufficient to demonstrate the efficacy. Need more quantitative analysis and experiments on more challenging datasets. The claim that it saves parameters compared to VAE is also confusing.

\subsection{Annotator Details} \label{sec:annotators}
Out of the 15 expert annotators in our study, 7 annotators are 2nd-year Ph.D students while 2 annotators are 1st-year Ph.D students, and the other 6 annotators are 3rd-year Ph.D students. 7 annotators are from India, 6 annotators are from Germany, 1 annotator is from Belgium and 1 annotator is from Italy. All of them are fluent English speakers who have experience in NLP peer-reviewing. The non-expert annotators are from the fields of Mathematics, Railway Engineering, Architecture, Civil Engineering, and Medieval History. Five of the annotators are from Germany, one from Hong Kong, one from India, two from France and the other from Italy. 
\subsection{Correlation between human and automated metrics} \label{sec:corr_human_auto}Since human evaluation metrics are widely used and easily interpretable~\cite{macina-etal-2023-mathdial, rashkin-etal-2021-increasing}, we analyze their correlation with automated metrics. We compute correlations between human and automated evaluations (Tables~\ref{tab:ReMuSE} and \ref{tab:human_eval}), with results shown in Fig~\ref{fig:human_auto}. We find that automated metrics closely aligned with human metrics exhibit strong correlations (e.g., K-Prec and Q2: 0.95), while those measuring different aspects, such as Specificity and Plausibility, show lower correlation (0.04). Lower correlations within the same category further highlight the diversity of captured aspects.

\subsection{Disagreement analysis for Human Evaluation in response generation} \label{sec:disagree}
We analyzed the instances where the scores varied by 2 or more points (e.g., one annotator scores a response 1 for fluency, while another scores it 4). Out of the 240 responses (60 from each model), the most significant disagreements occurred when rating ChatGPT’s responses (12 out of 60). Of these 12 responses, 7 are related to areas of improvement, 4 to the literature review, and 1 to a weakness. All of these responses are notably wordy, averaging 100 words, and often diverge off-topic, a pattern also observed in previous studies~\cite{fi15100336}. While these responses are fluent, the variation in objectivity scores between annotators suggests differing perspectives on the content. In contrast, we observed much higher agreement on responses from other models, excluding ChatGPT, with scores for relevance (0.75), faithfulness (0.78), objectivity (0.78), fluency (0.72), and helpfulness (0.70). For ChatGPT, the agreement scores were lower: fluency (0.52), relevance (0.65), faithfulness (0.68), objectivity (0.68), and helpfulness (0.68). Since these measures are used for evaluation rather than to generate gold standards, calibration is performed using averaging, as done in several prior works~\cite{faithdial, rashkin-etal-2021-increasing}.

\subsection{Interface for Conversation and Dialogue Agent Evaluation} \label{sec:interface}
We developed the interface using Streamlit to interact with the dialogue agent as shown in Fig~\ref{fig:dial_agent_inter}.\footnote{\url{https://streamlit.io/}} In Table~\ref{tab:dial_eval}, we observe higher scores for content relevance and Coverage while using a dialogue agent for both expert and non-expert meta-reviewers. However, decision correctness was lower while using a dialogue agent for the experts, though they perfectly aligned with the non-experts.

%\subsection{Statistical Significance Tests} \label{sec:stat_sig}
%Moreover, to corroborate our findings, we conducted paired t-tests and Wilcoxon signed-rank tests for both participant groups, calculating Cohen’s d for effect sizes and 95\% confidence intervals for mean differences. The results show highly significant improvements with the agent, indicating that even with these sample sizes, the effects are statistically robust. For the 15 PhD participants, content relevance increased by 0.39 points (t(14) = 10.84, p $\leq$ .001, d = 3.43), coverage by 0.38 points (t(14) = 10.56, p $\leq$ .001, d = 3.34), and time decreased by 30.1 minutes (t(14) = -19.45, p $\leq$ .001, d = -6.15). For the 10 non-PhD participants, content relevance increased by 0.48 points (t(9) = -13.50, p = .0002, d = 6.04), coverage by 0.55 points (t(9) = -10.59, p = .0004, d = 4.74), and time decreased by 45 minutes (t(9) = 8.51, p = .001, d = -3.81). Across both groups, effect sizes are very large to extremely large, demonstrating that the agent intervention produces robust and statistically significant improvements in user performance, even with modest sample sizes.

\subsection{Dialogue Process Modelling} \label{sec:dialog_model}
To assess whether ReMuSE captures the decision-making structure of meta-review dialogues, we perform a phase-level evaluation grounded in dialogue-system theory. Each dialogue turn is annotated as one of the three phases, following annotation schemes from task-oriented and debate systems: (i) \textbf{Evidence Gathering} – retrieving and summarizing reviewer claims; (ii) \textbf{Conflict Resolution} – identifying and weighing contradictory opinions; (iii) \textbf{Consensus Recommendation} –synthesizing points of agreement, summarizing trade-offs, and formulating a collective recommendation. We additionally measure \textbf{Full 3-Phase Completion}, defined as the proportion of dialogues that include all three phases in the correct sequential order, capturing the model's ability to simulate a complete meta-review reasoning process.

\noindent \textbf{Setup.} We train a lightweight RoBERTa~\cite{liu2019robertarobustlyoptimizedbert} classifier on 300 randomly sampled manually labeled dialogue turns to identify the three phases, reserving 20\% of the data as a hold-out set for evaluation. The classifier achieves a macro-F1 score of \textbf{0.83} on this held-out set. For each generated dialogue, we measure (i) the proportion of turns per phase and (ii) phase completeness, defined as the presence of all three phases in sequence. 

\noindent \textbf{Results.} We show in Table~\ref{tab:phase-completeness} that zero-shot dialogues rarely complete all phases (18\%), likely because the model lacks structured guidance for meta-review reasoning. Self-Consistency and Self-Refine yield moderate improvements, suggesting that iterative sampling and self-correction provide some structure but remain limited. \textbf{We show that ReMuSE achieves the highest coverage}, with 85\% evidence gathering, 77\% conflict resolution, 81\% consensus, and 71\% full phase completion, indicating that its training explicitly encourages the model to follow the multi-phase decision-making process and produce more structured, realistic meta-review dialogues.
    
We show the results for product buying and debates in Tables~\ref{tab:product-buying} and \ref{tab:debates} respectively. In product buying, dialogues emphasize Evidence Gathering (90\%) (researching options) with relatively lower Conflict Resolution (60\%), reflecting the focus on exploring alternatives before making a decision. In contrast, debates involve higher Conflict Resolution (counter-arguments) (85\%) and moderate Evidence Gathering (78\%), consistent with argumentation and synthesis processes. In both domains, ReMuSE consistently achieves the highest phase completion across all phases, demonstrating that modeling structured reasoning improves dialogue quality and generalizes effectively to tasks with different emphasis on information gathering, conflict handling, and consensus-building.

\begin{table*}[t]
\centering
\resizebox{0.8\textwidth}{!}{%
\begin{tabular}{lcccc}
\toprule
\textbf{Model} & \textbf{Evidence Gathering (\%)} & \textbf{Conflict Resolution (\%)} & \textbf{Consensus Phase (\%)} & \textbf{Full 3-Phase Completion (\%)} \\
\midrule
Zero-shot & 43 $\pm$ 4 & 28 $\pm$ 5 & 31 $\pm$ 6 & 18 $\pm$ 3 \\
Self Con. + Ref. & 52 $\pm$ 4 & 36 $\pm$ 5 & 42 $\pm$ 5 & 25 $\pm$ 3 \\
Self-Ref. & 60 $\pm$ 5 & 44 $\pm$ 4 & 51 $\pm$ 5 & 32 $\pm$ 4 \\
\textbf{ReMuSE (ours)} & \textbf{85 $\pm$ 5} & \textbf{77 $\pm$ 4} & \textbf{81 $\pm$ 5} & \textbf{71 $\pm$ 4} \\
\bottomrule
\end{tabular}%
}
\caption{Modeling of decision-making phases in generated meta-review dialogues. Each column shows the percentage of dialogue turns exhibiting the respective phase. Self-Consistency and Self-Refine provide intermediate improvements over zero-shot, while ReMuSE achieves the highest phase completeness.}
\label{tab:phase-completeness}
\end{table*}
\begin{table*}[t]
\centering
\resizebox{0.8\textwidth}{!}{%
\begin{tabular}{lcccc}
\toprule
\textbf{Model} & \textbf{Evidence Gathering (\%)} & \textbf{Conflict Resolution (\%)} & \textbf{Consensus Phase (\%)} & \textbf{Full 3-Phase Completion (\%)} \\
\midrule
Zero-shot & 55 $\pm$ 4 & 20 $\pm$ 5 & 30 $\pm$ 4 & 18 $\pm$ 3 \\
Self Con. + Ref. & 65 $\pm$ 4 & 28 $\pm$ 5 & 38 $\pm$ 4 & 25 $\pm$ 3 \\
Self-Refine & 72 $\pm$ 5 & 35 $\pm$ 4 & 45 $\pm$ 5 & 32 $\pm$ 4 \\
\textbf{ReMuSE (ours)} & \textbf{90 $\pm$ 5} & \textbf{60 $\pm$ 4} & \textbf{70 $\pm$ 5} & \textbf{65 $\pm$ 4} \\
\bottomrule
\end{tabular}%
}
\caption{Phase completion in product buying dialogues. Product buying emphasizes Evidence Gathering (researching options) with fewer conflicts, and ReMuSE achieves the highest completion across all phases.}
\label{tab:product-buying}
\end{table*}

\begin{table*}[t]
\centering
\resizebox{0.8\textwidth}{!}{%
\begin{tabular}{lcccc}
\toprule
\textbf{Model} & \textbf{Evidence Gathering (\%)} & \textbf{Conflict Resolution (\%)} & \textbf{Consensus Phase (\%)} & \textbf{Full 3-Phase Completion (\%)} \\
\midrule
Zero-shot & 35 $\pm$ 4 & 50 $\pm$ 5 & 28 $\pm$ 5 & 15 $\pm$ 3 \\
Self Con. + Ref. & 45 $\pm$ 4 & 60 $\pm$ 5 & 35 $\pm$ 5 & 22 $\pm$ 3 \\
Self-Refine & 50 $\pm$ 5 & 68 $\pm$ 4 & 42 $\pm$ 5 & 30 $\pm$ 4 \\
\textbf{ReMuSE (ours)} & \textbf{78 $\pm$ 5} & \textbf{85 $\pm$ 4} & \textbf{72 $\pm$ 5} & \textbf{68 $\pm$ 4} \\
\bottomrule
\end{tabular}%
}
\caption{Phase completion in debate dialogues. Debates involve higher Conflict Resolution (counter-arguments) with moderate Evidence Gathering and Consensus. ReMuSE achieves the highest completion across all phases.}
\label{tab:debates}
\end{table*}

\subsection{Effect of data scaling} \label{sec:data_scale}
\begin{figure}
    \centering
    \includegraphics[width=0.8\linewidth]{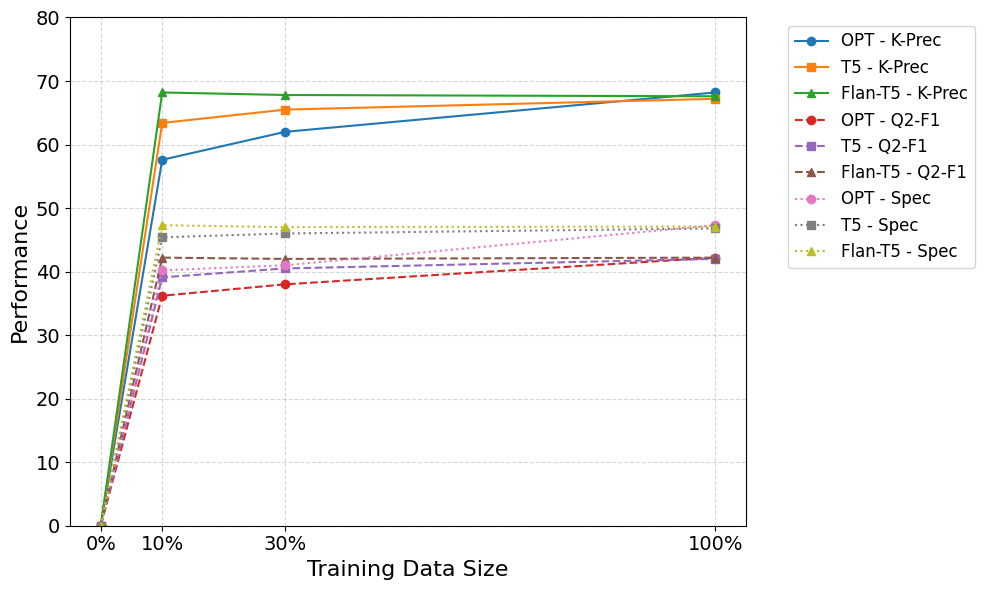}
    \caption{Performance of ReMuSE across different data scales.}
    \label{fig:data_scale}
\end{figure}

We show the performance of ReMuSE across different scales : 0,10,30 and 100 \% in Fig~\ref{fig:data_scale}. We find that even with smaller training data subsets (10–30\%), all the models achieve substantial gains in K-Precision, Q2-F1, and Specificity, with Flan-T5 already performing competitively at 10\% of the data. Increasing the training size to 100\% provides only modest additional improvements, suggesting that the algorithmic design and reward-guided fine-tuning, rather than sheer data scale, drive most of the performance gains. This highlights that ReMuSE efficiently leverages limited data to learn knowledge-grounded dialogue behavior.
\subsection{Dialogue Differences across paper types}\label{sec:deliberation}
\begin{figure}[!t]
    \centering
    \includegraphics[width=0.3\textwidth]{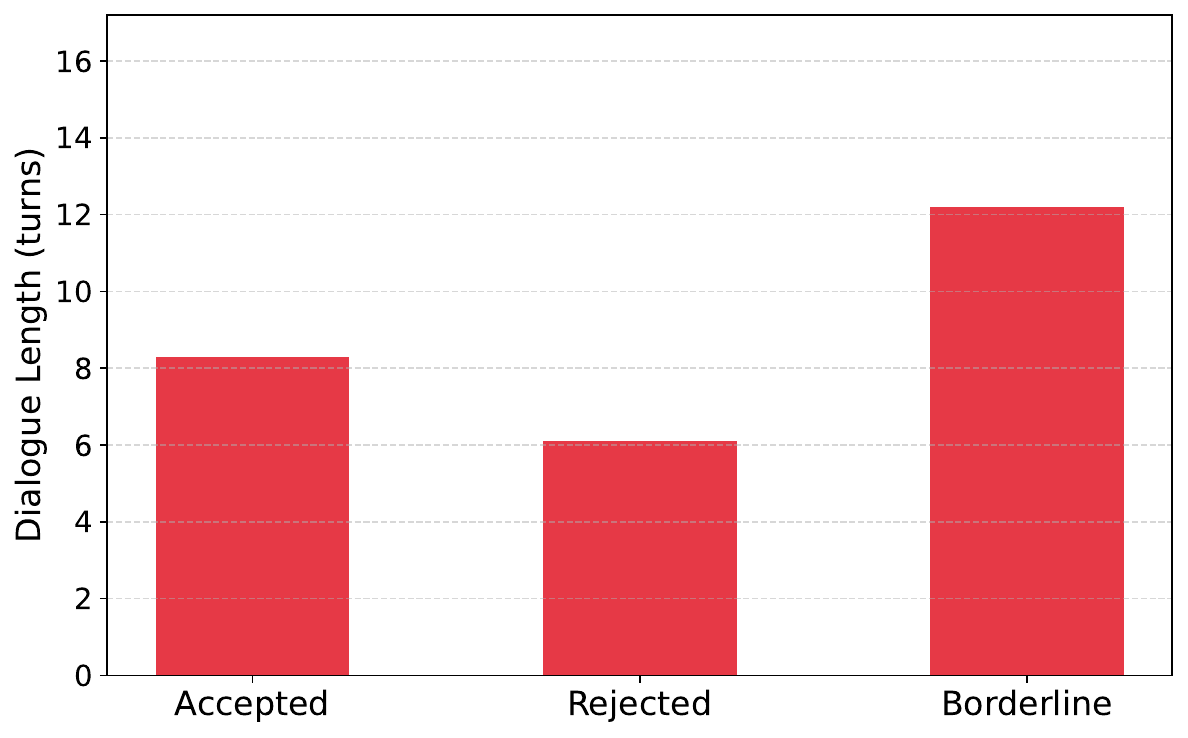}
    \caption{Distribution of dialogue length across paper types.}
    \label{fig:dial_length}
\end{figure}

\begin{figure}[!t]
    \centering
    \includegraphics[width=0.3\textwidth]{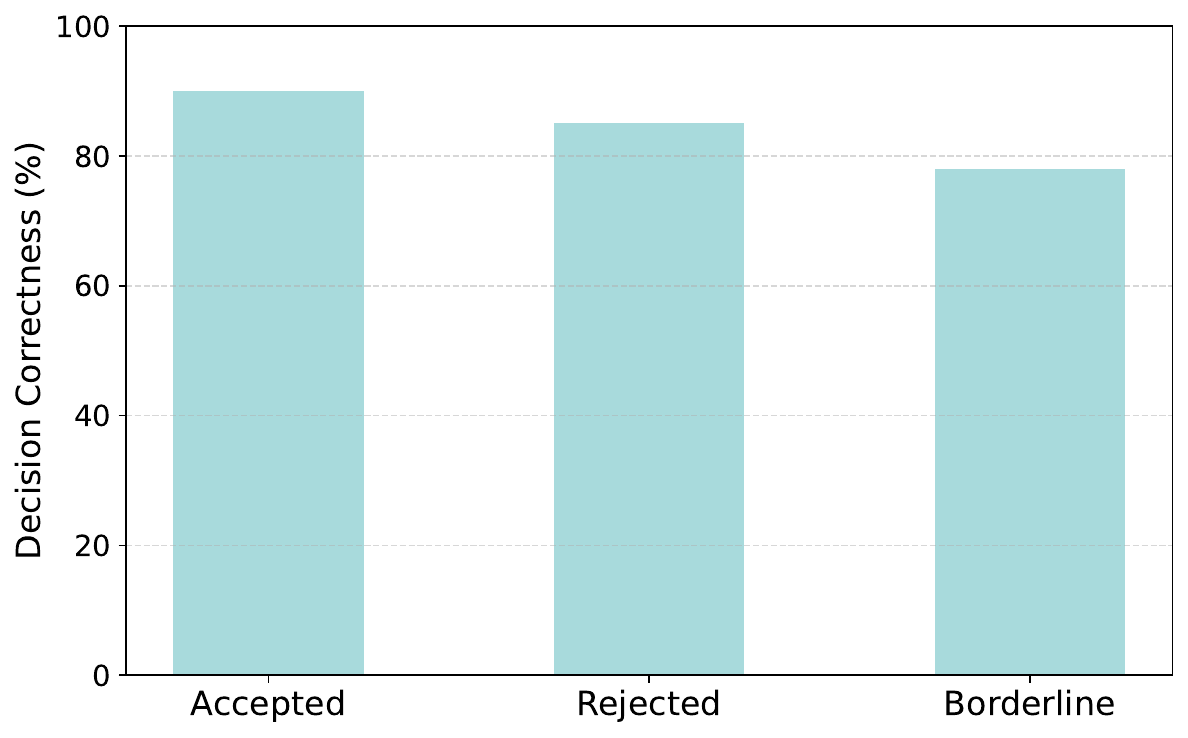}
    \caption{Distribution of decision correctness across paper types.}
    \label{fig:dec_correct}
\end{figure}

\begin{figure}[!t]
    \centering
    \includegraphics[width=0.3\textwidth]{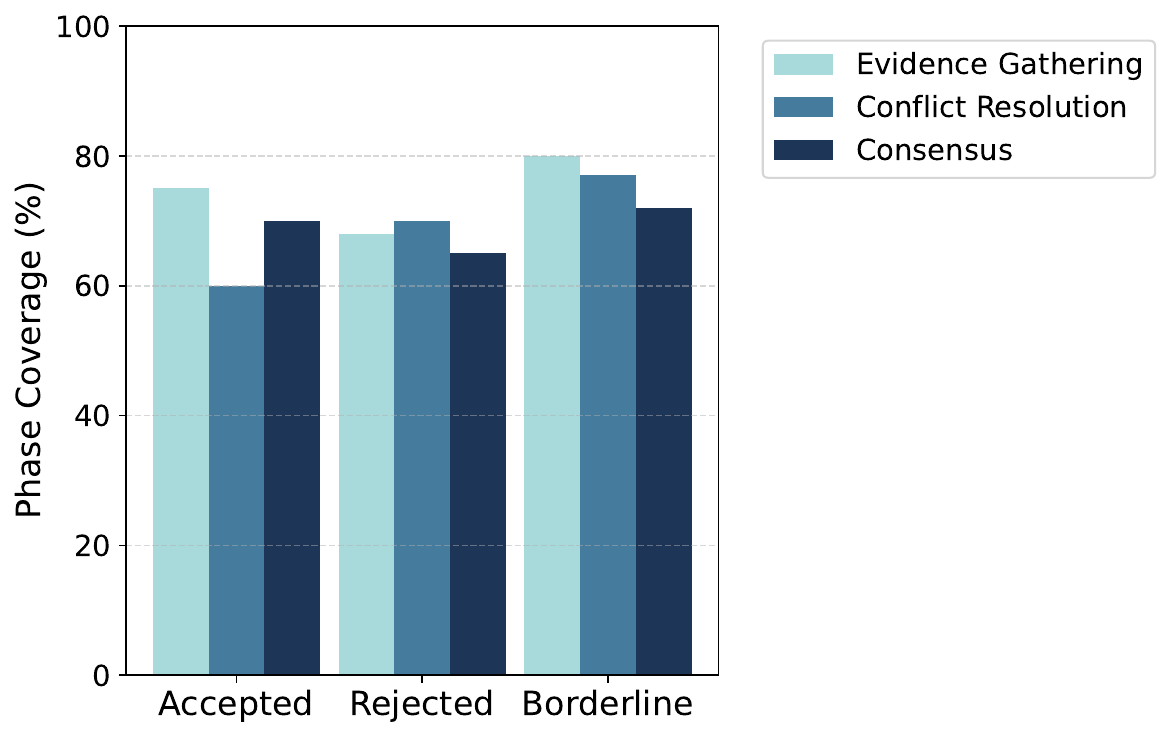}
    \caption{Distribution of phase coverage of dialogues across paper types.}
    \label{fig:phase_coverage}
\end{figure}

We show in Figure~\ref{fig:dial_length} that dialogue length varies across paper types, with borderline papers exhibiting the longest dialogues (12.2 turns), reflecting the greater deliberation required for conflicting or ambiguous reviews. Accepted papers have moderate lengths (8.3 turns), while rejected papers are shortest (6.1 turns). Figure~\ref{fig:dec_correct} shows decision correctness, which is highest for accepted papers (90\%) and slightly lower for rejected (85\%), with borderline papers lowest (78\%), indicating that the agent performs best when reviews converge and faces greater challenges in ambiguous cases. Finally, Figure~\ref{fig:phase_coverage} presents phase coverage for evidence gathering, conflict resolution, and consensus formation. Borderline papers have the highest coverage across all phases, especially conflict resolution (77\%), emphasizing structured deliberation, whereas accepted and rejected papers show slightly lower coverage. These results highlight that the agent adapts its behavior according to the complexity of the paper, supporting longer and more structured dialogues where necessary.

\subsection{Statistical Measures} \label{sec:appendix_stats}
We find that paired t-tests and Wilcoxon signed-rank tests, with Cohen's d, show robust improvements~\cite{cleophas2016paired}. For 15 PhD students, content relevance using the dialogue agent increased by 0.42 (t(14)=10.84, p$\leq$.001, d=3.43) and coverage by 0.52 (t(14)=10.56, p$\leq$.001, d=3.34), with task completion time decreasing by 15 minutes (t(14)=-19.45, p$\leq$.001, d=-6.15). In comparison, Auto Meta-Rev yielded smaller gains (Content Relevance +0.13, Coverage +0.19). For 10 non-PhD participants, content relevance rose by 0.54 (t(9)=-13.50, p$\leq$.001, d=6.04), coverage by 0.54 (t(9)=-10.59, p$\leq$.001, d=4.74), and time decreased by 45 minutes (t(9)=8.51, p$\leq$.001, d=-3.81), while Auto Meta-Rev showed more modest improvements (Content Relevance +0.22, Coverage +0.25). We do observe that the very large effect sizes and \textbf{highly significant p-values} indicate these improvements are likely to generalize to a larger population, and that the dialogue agent consistently outperforms both zero-shot and automated meta-review baselines (see Table~\ref{tab:dial_eval}).

\subsection{Performance of other open-source LLMs} \label{sec:other_open}
We also evaluate additional open-source LLMs, including Qwen 3-8B Instruct~\cite{yang2025qwen3technicalreport} and Gemma 3-12B-IT~\cite{gemmateam2025gemma3technicalreport}, with results reported in Table~\ref{tab:methods_metrics_rows} in comparsion to baselines discussed in \S\ref{sec:data} . Notably, Mixtral’s performance with ReMuSE still surpasses the other models, demonstrating the effectiveness of the dialogues generated by our approach.

\subsection{Operational guidelines for deploying the dialogue agent} \label{sec:operational}
The following operational guidelines outline measures to prevent misuse of dialogue agents in systems that implement our approach presented in Table~\ref{tab:misuse_prevention}.
\begin{table*}[!t]
\centering
\small
\begin{tabularx}{\linewidth}{X X X}
\toprule
\textbf{Misuse Prevention Mechanism (Design/UI)} & \textbf{Failure Mode Mitigated} & \textbf{Description} \\
\midrule
Citation Traceability & ``Ghostwriting'' / Unverified Claims & UI must display source snippets (reviewer ID, line/context) for all agent-generated claims, enabling instant verification. \\
\addlinespace
Human Final Decision & Automation Bias / Replacement of Human Judgment & The final decision (e.g., ``recommend reject'') must be a human-entered field, not an agent-generated output. \\
\addlinespace
Interleaved Presentation & Automation Bias / Reduced Critical Reading & Agent summaries and original source reviews should be presented side-by-side or be easily accessible to encourage comparison. \\
\addlinespace
Balanced Summary View & Anchoring Bias & Agent should present arguments for and against the final recommendation with equal prominence (e.g., randomized order). \\
\addlinespace
Decision Logging & Lack of Accountability / Misuse & Submission process logs whether the agent was used and which agent-generated arguments were directly adopted, creating an auditable trail. \\
\addlinespace
Strict Grounding & Confabulation / Hallucination & Enforce a hard constraint on generation, penalizing statements not mapped to input text; include a UI warning for any unsourced claims. \\
\bottomrule
\end{tabularx}
\caption{Design/UI mechanisms to mitigate misuse of agent-generated review feedback.}
\label{tab:misuse_prevention}
\end{table*}

\begin{table}[!t]
\centering
\resizebox{0.4\textwidth}{!}{\begin{tabular}{lccc}
\toprule
\textbf{Method} & \textbf{K-Prec} & \textbf{Q2-F1} & \textbf{Spec} \\
\midrule
\multicolumn{4}{c}{\textbf{Qwen 3}} \\
Zero-Shot         & 0.36 & 0.19 & 0.42 \\
Self-Ref.    & 0.45 & 0.20 & 0.61 \\
Self-Cons. + Ref.   & 0.74 & 0.22 & 0.65 \\
ReMuSE             & 0.75 & 0.30 & 0.72 \\
\midrule
\multicolumn{4}{c}{\textbf{Gemma 3}} \\
Zero-Shot         & 0.32 & 0.17 & 0.41 \\
Self-Ref.     & 0.47 & 0.20 & 0.62 \\
Self-Cons. + Ref.   & 0.75 & 0.21 & 0.63 \\
ReMuSE             & 0.75 & 0.30 & 0.71 \\
\midrule
\multicolumn{4}{c}{\textbf{Mixtral}} \\
Zero-Shot.    & 0.38 & 0.17 & 0.42 \\
Self-Ref.    & 0.50 & 0.20 & 0.51 \\
Self-Cons. + Ref.   & 0.56 & 0.24 & 0.59 \\
ReMuSE             & \textbf{0.78} & \textbf{0.32} & \textbf{0.72} \\
\bottomrule
\end{tabular}}
\caption{Performance of different LLMs across various methods. Metrics reported are K-Precision, Q2-F1, and Specificity.}
\label{tab:methods_metrics_rows}
\end{table}

\subsection{End-to-end Workflow} \label{sec:end2end}
\begin{figure*}
    \centering
    \includegraphics[width=\textwidth]{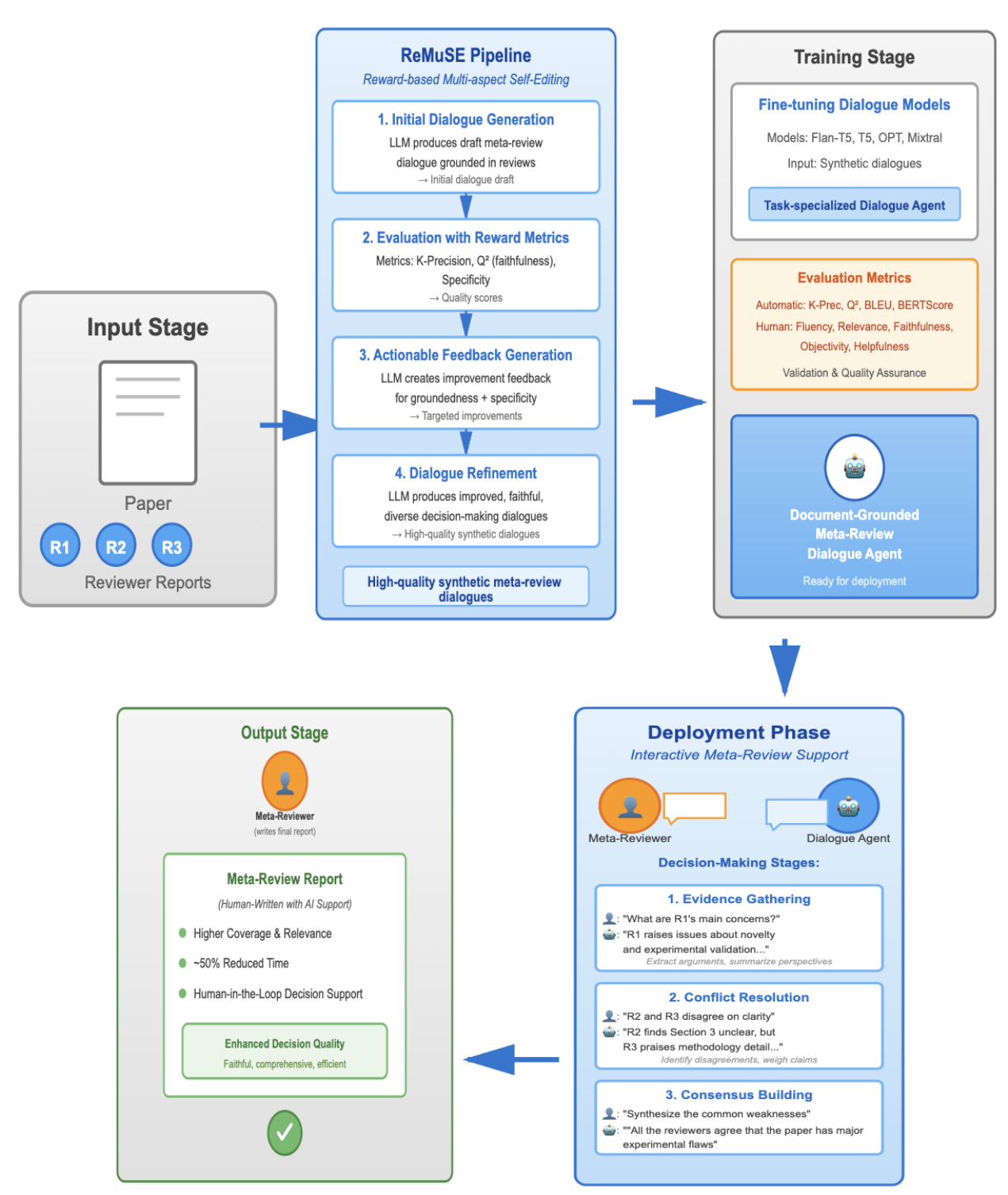}
    \caption{End-to-end diagram outlining the various stages in this work}
    \label{fig:end_to_end}
\end{figure*}

We present the end-to-end workflow in our paper, summarizing all stages in Fig.~\ref{fig:end_to_end}. First, we generate synthetic dialogues for meta-reviewing using our proposed method, \textbf{ReMuSE}. Next, we fine-tune dialogue agents to adapt smaller models for the task of meta-reviewing assistance. Finally, we deploy the dialogue agent in real-world settings, demonstrating that this deployment reduces meta-reviewing time by 50\%.

\subsection{Dialogue excerpts showing various stages of dialogue agent assistance} \label{sec:dial_assist}
In the following table, we show the various stages of dialogue agent assistance.
\begin{table*}[ht]
\centering
\resizebox{0.8\textwidth}{!}{\begin{tabular}{>{\raggedright\arraybackslash}p{3cm} >{\raggedright\arraybackslash}p{8cm} >{\raggedright\arraybackslash}p{4cm}}
\toprule
\textbf{Role} & \textbf{Dialogue Excerpt} & \textbf{Focus} \\
\midrule
Meta-Reviewer (MR) & What are the main opposing viewpoints regarding the paper’s novelty and empirical support? & Evidence Collection \\

Dialogue Agent (DA) & Reviewer 1 (R1) rates the paper as Accept (Score: 5) due to 
"its novel and efficient approach to real-time data streaming that outperforms all listed baselines" [R1, §3]. 
In sharp contrast, Reviewer 3 (R3) recommends Reject (Score: 2), arguing that the 
"empirical gains are only marginal on standard benchmarks and the proposed method introduces too much complexity" [R3, §2]. & Conflict Identification \\

MR & R3’s point about "marginal gains" and complexity is concerning. Does R1 or R2 offer any perspective on why the complexity is justified? & Resolution Inquiry \\

DA & Reviewer 2 (R2), who rated the paper 4/Weak Accept, directly supports the added complexity. R2 states, 
"The conceptual elegance of the attention mechanism alone justifies the extra parameters, as it opens up new avenues for interpretability" [R2, §4]. 
This suggests the complexity is a trade-off for a valuable architectural feature. & Evidence Retrieval / Conflict Mitigation \\

\bottomrule
\end{tabular}}
\caption{Dialogue excerpts illustrating meta-reviewing process.}
\label{tab:dialogue_excerpts}
\end{table*}
\clearpage
\subsection{Example Dialogues from our Dataset} \label{sec:examples}
%\subsubsection{Meta-Reviews}
We show a few reviews from our dataset along with the dialogue generated using ReMuSE below.
\\
\subsubsection{Meta-Reviewing} \label{sec:meta-reviews}
The title of the paper is not revealed to retain anonymization. The type of the paper is ``long".\\

\noindent\fbox{%
    \parbox{\textwidth}{%
        \textbf{Review 1: } Paper is on modeling the prediction of ancestor relation between names of science institutions.  This is on the GRID dataset which already has some hierarchical information.  The proposed approach is set-based models (with neural encodings) where the overlap between two names is measured by set overlap at the unigram level.  In extended experiments additional metadata like address and type of institution are also incorporated into the model (which contribute a lot to the improvements).  A set of simple to intermediate baseline along with different thresholds of token overlap has been tested and the proposed model shows strong improvement in the MAP metric. Paper has a decent writing and structure.  Problem and the approach has been explained and motivated well with descriptive examples. [...]

    }% 

}

\vspace{50\fboxrule}

\noindent\fbox{%
    \parbox{\textwidth}{%
        \textbf{Review 2: } The paper shows how to infer the organisational structure of an institution. That is, it presents a model for predicting the is-ancestor relationships of institutions based on their string names. To this end, it makes use of Set-Transformers to model the token overlap between the institution names. This use is nice but also not highly original. The experimental evaluation is on a single dataset only. While the authors do present some examples, and overall hierarchy or something that provides some more insights into the learned model should be provided in order to show potential issues with transitivity and connected components. The evaluation only considers known pairs. But an organisational structure should also be consistent.  That is, the interesting motivation provided in the intro is not met in the experimental evaluation. Furthermore, the experimental protocol is  unclear [...]
    }% 

}

\
\begin{table*}[]
\begin{small}
\begin{tabularx}{\textwidth}{@{}lX@{}}
\toprule
\multicolumn{2}{l}{\textbf{Meta-reviewing Dialogue}}                                                                                           \\ \midrule
\textit{Meta-Reviewer} & Hello, I'm reviewing a paper on predicting institution hierarchies with set-based models. Can you tell me a little bit about the paper?\\ \\ \textit{Dialogue Agent} &  Sure! The paper presents a new approach to predicting the hierarchical structure of institutions using set-based models with neural encodings. The approach is based on the GRID dataset, which already has some hierarchical information.\\ \\ \textit{Meta-Reviewer} &  That sounds interesting. Can you tell me more about the dataset and the experimental setup?\\ \\ \textit{Dialogue Agent} & Sure! The dataset is the GRID dataset, which is a global research identifier database. The authors have constructed a new dataset by pairing institution names with their corresponding hierarchy information. They have also compared their approach against a few baseline approaches and presented some examples.\\  \\ \textit{Meta-Reviewer} &  Okay, that sounds good. But I have a few doubts regarding the experimental setup. For example, I would have expected a comparison with well-known baseline models from the knowledge base completion literature, such as TransE. And I'm not sure if the manually extracted test set is complete. Can you tell me more about this?\\ \\ \textit{Dialogue Agent} &  Yes, the authors have mentioned some of these concerns in their paper. They have developed a new dataset, and the task is well-defined. However, they have not compared their approach with well-known baseline models from the knowledge base completion literature, such as TransE. And the manually extracted test set may not be complete, which may affect the evaluation.\\ \\ \textit{Meta-Reviewer} & I see. That's a good point. What about the token-based models? Are they a promising choice for the task?\\ \\ \textit{Dialogue Agent} & The authors have presented some examples in Table 3 to show the effectiveness of their approach. However, the task is presented as a knowledge-base completion task, and the token-based models may not be the best choice for this task.\\ \\ \textit{Meta-Reviewer} & Okay, I understand. What about the related work on set-based models? Is there any extension of the related work in the paper?\\ \\ \textit{Dialogue Agent} & The authors have extended the related work on set-based models in their paper. They have presented some new approaches and compared them with the baseline models. However, the related work on set-based models should be extended further.\\ \\ \textit{Meta-Reviewer} & I see. Thank you for the information. Based on your provided information, I would rate this paper as a 6 out of 10, with a confidence level of 4. The paper has some interesting ideas and the authors have presented some examples to show the effectiveness of their approach. However, there are some concerns regarding the experimental setup and the token-based models. The related work on set-based models should be extended further.\\ \\ \textit{Dialogue Agent} & Thank you for your feedback. \\ \bottomrule
 \end{tabularx}
\caption{Generated Meta-Reviewing Dialogue from our dataset based on the provided reviews.}
\label{tab:meta_rev_eg}
\end{small}
\end{table*}
\clearpage
% Please add the following required packages to your document preamble:
% \usepackage{multirow}

% Please add the following required packages to your document preamble:
% \usepackage{multirow}
% Please add the following required packages to your document preamble:
% \usepackage{multirow}
% Please add the following required packages to your document preamble:
% \usepackage{multirow}
\subsubsection{Debates}
The topic of the paper is: `Global Warming is Not a Crisis'. The arguments for for and against sides of the debate are provided below:\\

\noindent\fbox{%
    \parbox{\textwidth}{% 
    \textbf{Against side}: ``Global warming is not a crisis. Please keep in mind what the proposition is. It is not a debate over whether the earth has been warming over the past century. Uh, the earth is always warming or cooling, at least a few tenths of a degree. And we’re talking about, so far, something on the order of six tenths of a degree centigrade. We’re not even arguing about whether greenhouse gas emissions are contributing at some level to warming. And they most certainly should or I would suggest it would be very little. Indeed, as far as I can tell, even our opponents do not claim that global warming is a crisis at present. Rather, we are primarily addressing the future. Now, much of the current alarm, I would suggest, is based on ignorance of what is normal for weather and climate. [...]
    }%
    }
   
    \vspace{100\fboxrule}

\noindent\fbox{%
    \parbox{\textwidth}{% 
    \textbf{For side}: ``The motion before us, global warming is not a crisis, means we ought to know what crisis means. The word does not mean catastrophe or alarmism. It means a crucial or decisive moment, a turning point, a state of affairs in which a decisive change for better or worse is imminent. We are talking about the future here. The entire world now really does have a critical choice to make. It is whether to continue on the present path of adding more and more carbon dioxide and other greenhouse gases to the atmosphere or whether to find another path. We’re speaking of the future. And science tells us that the path we choose will largely determine what kind of earth our children and grandchildren will inherit. Our task tonight is to persuade you that global warming is indeed a crisis in exactly that precise sense so you should vote against the motion. The science community today has impeccable settled science, despite what you have just heard, that demonstrates the reality of global warming and its primary origin in human activities. [...]
    }%
    }

\begin{table*}[]
\begin{small}
\begin{tabularx}{\textwidth}{@{}lX@{}}
\hline
\multicolumn{2}{l}{\textbf{Debate Decision-Making Dialogue}}                                                                                                                                                                          \\ \hline
\textit{Decision Maker} & Hello, I am here to learn more about the debate topic, ``Global warming is not a crisis." Can you provide me with information about both sides of the argument? \\ \\
\textit{Dialogue Agent} & Of course, for the ``For" side, they argue that global warming is not a crisis at present and much of the current alarm is based on ignorance of what is normal for weather and climate. They suggest that extreme weather events always occur without evidence of systematic increases. Additionally, the sea level has been increasing since the end of the last Ice Age glaciation. The impact of warming on agriculture and disease seems dubious at best. On the other side, the ``Against" arguments state that global warming is indeed a crisis in the sense that it is a turning point, a state of affairs in which a decisive change for better or worse is imminent. They argue that science has proven the reality of global warming and that mankind's fingerprints have now clearly emerged above the noise of natural variability. They also mention the increasing frequency of heat waves, higher sea levels, and vanishing glaciers as evidence of the severity of the situation. \\ \\
\textit{Decision Maker} & Thank you for providing that information. Can you tell me about any projections or forecasts related to global warming?\\ \\
\textit{Dialogue Agent} & Yes, the "Against" side states that they project further warming of half a degree Fahrenheit for the next twenty-five years. Beyond that, it largely depends on how much more CO2 and other greenhouse gases humanity dumps into the atmosphere. They also warn that continuing to fuel the world using present technology will bring dangerous and possibly surprising climate changes by the end of this century, if not sooner. \\ \\
\textit{Decision Maker} & What is the scientific community's stance on global warming? \\ \\
\textit{Dialogue Agent} & The "Against" side argues that today's science community has impeccably settled science that demonstrates the reality of global warming and its primary origin in human activities. They also mention the Intergovernmental Panel on Climate Change's report and the extensive process of experts who represent mainstream science. On the other hand, the ``For" side acknowledges that even their opponents do not claim that global warming is a crisis at present and suggest that the impact of CO2 per unit of CO2 actually goes down with increasing CO2. They also argue that it is crucial to distinguish between the claim that models can display past behavior from the actual situation, which is that models can be adjusted to display past behavior once that behavior is known. \\ \\
\textit{Decision Maker} & Thank you for the information. Based on both sides of the argument, what do you suggest is the conclusion? \\ \\
\textit{Dialogue Agent} & As a dialogue agent, I cannot express opinions or make decisions. However, I provided the arguments from both sides that may help you reach your own conclusion. \\ \bottomrule
\end{tabularx}
\caption{Debate Dialogue on the topic `Global warming is not a crisis' generated from the provided arguments.}
\end{small}
\end{table*}
\clearpage
\subsubsection{Product Reviews}
The product in discussion is ``Flexion KS-902 Kinetic Series Wireless Bluetooth Noise Cancelling Headphones with Microphone, Black". The reviews are below: \\\\
\noindent\fbox{%
    \parbox{\textwidth}{% 
    \textbf{Reviews}: ``The battery life is sufficient where I only need to charge once every 2 days These headsets are equipped with the new Bluetooth 9 edition too advance for current music, so don’t even bother to enable because they don’t work. \textbackslash n \\
    
    These gave very good sound. Good sound quality, fits pretty well in your ear And it is also bluetooth, so I can connect it to my phone easily!\textbackslash n\\
    
    The user interface is really bad, when I try to turn it off, it usually calls someone on my phone, so I have had several butt dials thanks to these earphones.\textbackslash n \\
    
    There well rounded in all areas something that others don't have. And they totally out perform the 2 hour battery life of the Flexion. The case is badly designed and too small.\textbackslash n \\
    
    These work really great when you want to listen to something without bothering other people. sleep.- Bass is amazing and really good- Loud and really clear For the price I paid they work excellent!\textbackslash n \\
    
    my background noise was amplified. He told me that he loves the headphones, but it was the amazing C.S that will have him as a returning customer. I recommend it to everyone! They actually turn on with a quick click, not a 3 second hold as in the instructions. The sound is just fantastic, the best sound I ever hear from an ear bud type speaker. Then my replacement pair broke, the cord just separated from the ear bud during normal use (see picture attached). \textbackslash n \\
    
    Sound quality is good and has buttons on the ear pieces for volume control, answering phone calls, and changing songs! The pause button and the advance track button doesn't work and the previous track button is awkward and works only some of the time. As a commuter using public transportation, I was looking for something that would eliminate wires going to to my device and these do a great job of that. build quality is great compared to the price they do look expensive* I love that they were really easy to sync, which is a huge bonus for me! The headphones connected to my iPhone with no problems. \textbackslash n \\ 
    
    Sound quality was great but even after a full night of charging the batteries died after 2 hours. they give you a low battery warning a few seconds before they shut off which is kinda pointless. \textbackslash n [...]

    }%
    }
% Please add the following required packages to your document preamble:
% \usepackage{multirow}
\begin{table*}[]
\begin{small}
\begin{tabularx}{\textwidth}{@{}lX@{}}
\hline
\multicolumn{2}{l}{\textbf{Product Buying Dialogue}}                                                                                                                                                                                                                                     \\ \hline
\textit{Buyer} & Hi, I'm interested in purchasing the Flexion KS-902 Kinetic Series Wireless Bluetooth Noise
Cancelling Headphones with Microphone, Black. Can you provide me with some information about the product?\\ \\ 
\textit{Dialogue Agent} & Sure, what would you like to know?\\ \\ \textit{Buyer} & Can you tell me about the battery life of the headphones?\\ \\ 
\textit{Dialogue Agent} & According to the reviews, the battery life is sufficient where you only need to charge once every 2 days. However, there have been a few reviewers who experienced a shorter battery life.\\ \\ 
\textit{Buyer} & What about the sound quality?\\ \\ 
\textit{Dialogue Agent} & The reviews state that the sound quality is good, and they worked really great when you want to listen to something without bothering other people. However, there were some reviews that reported the sound quality wasn't that great, and some experienced full of static if you get more than 7 feet from your phone.\\ \\ 
\textit{Buyer} & Is it easy to connect to my phone?\\ \\ 
\textit{Dialogue Agent} & Yes, the headphones are equipped with Bluetooth and are easy to sync with your iPhone or Android device.\\ \\ 
\textit{Buyer} & What about the user interface? Is it easy to use?\\ \\ \textit{Dialogue Agent} & According to some reviews, the user interface is really bad. They experienced several butt dials when trying to turn it off. However, others found the headphones to be easy to connect and use.\\ \\ 
\textit{Buyer} & With all that information, do you recommend these headphones?\\ \\ 
\textit{Dialogue Agent} & I'm sorry, I cannot provide recommendations or opinions. Please use the reviews provided to make an informed decision.\\ \\ 
\textit{Buyer} & Okay, thanks for the information. I think I'll go ahead and purchase them! \\ \bottomrule
\end{tabularx}
\caption{Dialogue generated for the product `Flexion KS-902 Kinetic Series Wireless Bluetooth Noise Cancelling Headphones with Microphone, Black' with the reviews provided before.}
\end{small}
\end{table*}

\end{document}